\documentclass[lettersize,journal]{IEEEtran}
\usepackage[dvipsnames]{xcolor}
\usepackage{amsmath,amssymb,amsfonts}
\usepackage{mathtools}
\usepackage[colorlinks,bookmarksopen,bookmarksnumbered,citecolor=black,urlcolor=black]{hyperref}

\usepackage{amsthm}
\theoremstyle{plain}

\usepackage{siunitx}
\ifdefined\unit\else
\ifdefined\NewCommandCopy
\NewCommandCopy\unit\si
\else
\NewDocumentCommand\unit{O{}m}{\si[#1]{#2}}
\fi
\fi

\newtheorem{assume}{Assumption}

\usepackage{cite}

\usepackage{graphicx}
\usepackage{textcomp}
\usepackage{color}
\usepackage{physics}
\usepackage{booktabs}
\usepackage{pifont}
\usepackage{enumitem}
\usepackage{algorithm}
\usepackage{algpseudocode}
\usepackage{glossaries}
\newcommand{\rev}[1]{{\color{black} #1}} 
\newcommand{\new}[1]{{\color{black}{#1}}}
\usepackage{subfigure}
\usepackage{multicol}
\newcommand{\change}[1]{{\color{black}{#1}}}

\begin{document}
	\hypersetup{ pdftitle = {Optimal Control for Articulated Soft Robots}, pdfauthor = {Saroj Prasad Chhatoi, Michele Pierallini, Franco Angelini, Carlos Mastalli, Manolo Garabini}, pdfsubject = {Submitted to Transaction on Robotics (T-RO)}, pdfkeywords = {articulated soft robots, \change{underactuated compliant} robots, optimal and state-feedback control, feasibility-driven differential dynamic programming}, 
		pdftoolbar = true, 
		colorlinks = true, 
		linkcolor = black, 
		citecolor = black, 
		urlcolor = black}

\title{Optimal Control for Articulated Soft Robots}

\author{Saroj Prasad Chhatoi$^{1,2}$\quad Michele Pierallini$^{1,2}$\quad Franco Angelini$^{1,2}$\\ Carlos Mastalli$^{3,4}$\quad Manolo Garabini$^{1,2}$%
\thanks{This research is partially supported by the European Union’s Horizon 2020 Research and Innovation Programme under Grant Agreement No. 101016970 (Natural Intelligence), No. 871237 (SOPHIA) and No. 780684 (Memory of Motion), and in part by the Ministry of University and Research (MUR) as part of the PON 2014-2021 “Research and Innovation" resources – Green Action - DM MUR 1062/2021.}%
	\thanks{$^{1}$Centro di Ricerca ``En\-ri\-co Pi\-ag\-gio'', U\-ni\-ver\-si\-t\`{a} di Pisa, 
		Largo Lucio Lazzarino 1, 56126 Pisa, Italy} 
	\thanks{$^{2}$Dipartimento di Ingegneria dell'Informazione,  U\-ni\-ver\-si\-t\`{a} di Pisa, 
		Largo Lucio Lazzarino 1, 56126 Pisa, Italy}%
	\thanks{$^{3}$Institute of Sensors, Signals and Systems, School of Engineering and Physical Sciences, Heriot-Watt University, Edinburgh EH14 4AS, UK}%
	\thanks{$^{4}$National Robotarium, Edinburgh, UK}%
	\thanks{Corresponding Author: Saroj Prasad Chhatoi {\tt\footnotesize email: s.chhatoi@studenti.unipi.it}}%
}

\maketitle

\begin{abstract}
	Soft robots can execute tasks with safer interactions. However, control techniques that can effectively exploit the systems' capabilities are still missing. Differential dynamic programming (DDP) \change{has} emerged as \change{a} promising tool for achieving highly dynamic tasks. But most of the literature \change{deals with applying} DDP to articulated soft robots \change{by} using numerical differentiation, \change{in addition to using} pure feed-forward control \change{to perform explosive tasks}. Further, \change{\change{underactuated compliant}} robots are known to be difficult to control and the use of DDP-based algorithms to control them is not yet addressed. We propose an efficient DDP-based algorithm for trajectory optimization of articulated soft robots that can optimize the state trajectory, input torques, and  stiffness profile. We provide an efficient method to compute the forward dynamics and the analytical derivatives of series elastic actuators (SEA)/variable stiffness actuators (VSA) and \change{\change{underactuated compliant}} robots. We present a state-feedback controller \change{that} uses locally optimal feedback policies obtained from DDP. We show through simulations and experiments that the use of feedback is crucial  in improving the performance and stabilization properties of various tasks. We also show that the proposed method can be used to plan and control \change{\change{underactuated compliant}} robots, \rev{with varying degrees of underactuation} effectively.

\end{abstract}

\begin{IEEEkeywords}
articulated soft robots, \change{\change{underactuated compliant}} robots, optimal and state-feedback control, feasibility-driven differential dynamic programming
\end{IEEEkeywords}

\section{Introduction}

\begin{figure}[t]
  \centering
 \subfigure[variable stiffness actuation.]{\includegraphics[width=.49\columnwidth]{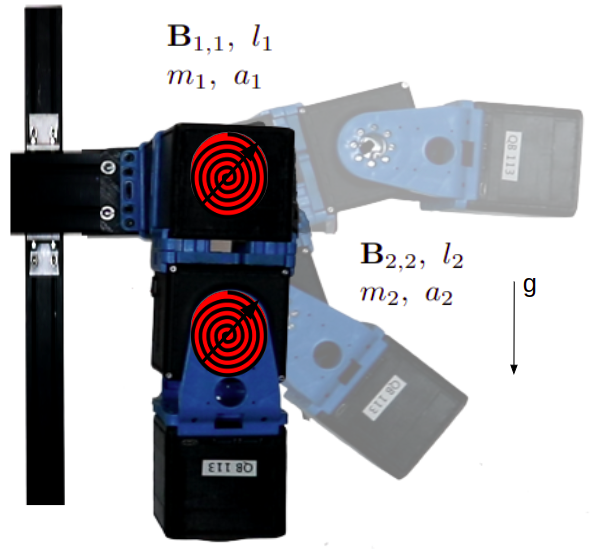}}
 \subfigure[series elastic actuation.]{\includegraphics[width=.435\columnwidth]{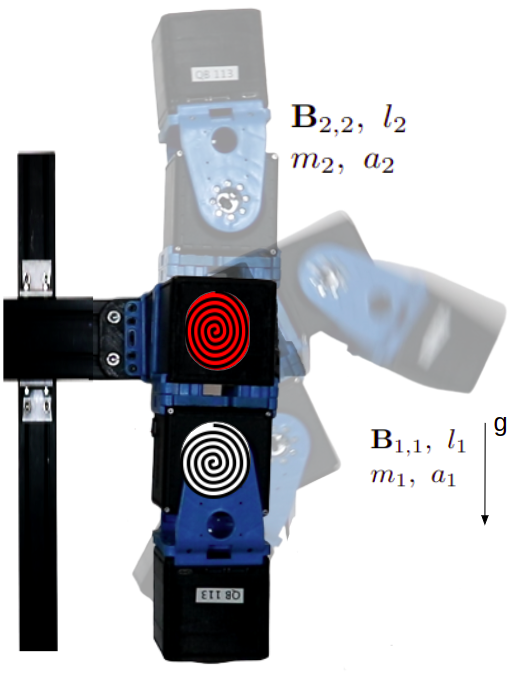}}
 \caption{{Examples of articulated soft robots with joints that present torsional springs. The red joints are actuated, while the white one is unactuated. (a) Two degrees of freedom robot actuated by VSA performing a regulation task. (b) Two degrees of freedom robot with an actuated SEA joint, and an unactuated elastic joint.}
}
  \label{fig:cover}
\end{figure}

Across many sectors such as the healthcare industry, we require robots that can actively interact with humans in unstructured environments.
To enable safe interactions and increase energy efficiency, %
we often include soft elements in the robot structure \cite{della2020soft}\cite{energy_efficieny}. For instance, in an articulated soft robot (ASR) the rigid actuators connect to the joints through passive elements with or without variable stiffness (Fig. \ref{fig:cover}).
These \new{types} of robots aim to mimic the musculoskeletal structure in vertebrated animals \cite{animals, chang2019feedforward}, which enables them to perform highly dynamic tasks efficiently \cite{gasparri2018efficient,werner_ott}.
\rev{A series elastic actuator (SEA) has a linear spring between the actuator and the load \cite{prattsea}. Instead, a variable stiffness actuator (VSA) integrates an elastic element that can be adjusted mechanically. These actuators provide many potential advantages but also increase the control complexity  \cite{vanderborght2013variable}.} \new{ Similarly, compliant robots, a subclass of soft robots, are systems with rigid links and elastic joints (e.g., flexible joint robot SEA/VSA) in which a generic number of unactuated joints can also be present \cite{pierallini2023iterative}. This mechanism further increases the modeling and control complexity. In addition, this class resembles other modeling formulations used in the soft robotics literature \cite{della2020soft,flexible_link_review}: Pseudo Rigid Body(PRB) model \cite{prb_1, prb_2}, Cosserat Model \cite{cosserat_1}, Constant Curvature models \cite{const_curv_review} and thus are an important class of models.}

Applying controllers derived for rigid robots tends to provide an undesired performance in soft robots (see Sec. \ref{sec:motivation}). It may even have a detrimental effect, as it provides dynamically infeasible motions and controls.
Therefore, we need to design control techniques that fully exploit the dynamic potential of soft robots. In this regard, optimal control solutions promise to be an effective tool.

Differential dynamic programming (DDP) is an optimal control method that offers fast computation and can be employed in systems with high degrees of freedom and multi-contact \change{setups} \cite{crocoddyl,multiphase}. 
 
In the context of soft robots, iterative LQR (iLQR) has been used to perform explosive tasks such as throwing a ball with maximum velocity by optimizing the stiffness and controls \cite{ilqr_explosive}. Similarly, DDP has enabled \change{us} plan time-energy optimal trajectories for systems with VSAs as well \cite{time_energy_optimal}. These works employ numerical differentiation to compute the derivatives of the dynamics and cost functions. But such \change{an} approach is computationally expensive, and it is often prone to numerical errors. These works also completely rely on feed-forward control. Further, to the best of the authors' knowledge, the control of \change{underactuated compliant} systems has not been addressed using DDP algorithms. \change{Furthermore,} devising control laws for such systems is known to be difficult \cite{flexible_link_mich}. \rev{We propose an optimal control method for articulated soft robots in this work.}

\subsection{Contribution}
\rev{ In this paper, we propose an efficient optimal control method for articulated soft robots based on the feasibility-driven differential dynamic programming (FDDP)/Box-FDDP algorithm that can accomplish different tasks. It boils down to three technical contributions:
\begin{enumerate}[label=(\roman*)]
    \item an efficient approach to compute the forward dynamics and its analytical derivatives for robots with SEAs, VSAs and under-actuated compliant arms,
    \item empirical evidence of the benefits of analytical derivatives in terms of convergence rate and computation time, and
    \item a state-feedback controller that improves tracking performance in soft robots.
\end{enumerate}

Our approach boosts computational performance and improves numerical accuracy compared to numerical differentiation. The state-feedback controller is validated in experimental trials on systems with varying degrees of freedom.  We provide the code to be publicly accessible.\footnote{\href{https://github.com/spykspeigel/aslr_to}{github.com/spykspeigel/aslr\_to }}
}

The article is organized as \change{follows}: after discussing state of the art in optimal control for soft robots (Section \ref{sec:soa}), we describe their dynamics and formulate their optimal control problem in Section \ref{sec:method}. Section \ref{sec:solution} begins by summarizing the DDP formalism and ends with the state-feedback controller. In Section \ref{sec:validation}, we introduce various systems that we use for validating the proposed method. Finally, Section \ref{sec:results} shows and discusses the efficiency of our method through a set of simulations and experimental trials.

\section{Related Work}\label{sec:soa}
Compliant elements introduce redundancies in the system that increases the complexity of the control problem. Optimal control is a promising tool to solve such \new{kinds} of problems.
It can be classified into two major categories: 1) Direct, 
2) Indirect methods.
Indirect methods first optimize the controls using Pontryagin's Maximum Principle (PMP) and then discretize the problem.
This approach has been used to compute optimal stiffness profiles while maximizing the terminal velocity
as shown in \cite{vsa_kick},\cite{dlr_vsa}. 
In \cite{luca_optimal_flexible}, the authors use linear quadratic control of an Euler beam  model and show its effectiveness w.r.t. PD/ state regulation method.  But such methods have \change{poor convergence under bad initialization} and cannot handle systems with many degrees.

\rev{Instead, direct methods transcribe the differential equations into algebraic ones that are solved using general-purpose nonlinear optimizers. In \cite{8957077}, authors propose a time-optimal control problem for soft robots, and it is solved using \new{the} direct method where the non-convexity of the problem is converted into bilinear constraints.} Similarly, in \cite{min_time_flexible_joint, Altay_time} direct methods are used to solve minimum time problems, and in \cite{Altay_energy,time_energy_optimal} direct methods are used to solve energy-optimal problems for soft robots. However, these methods often cannot be used in model predictive control settings as they are computationally \change{slow.}

Dynamic programming uses the Bellman principle of optimality to solve a sequence of \change{small}er problems recursively. But this approach suffers from \change{the} curse of dimensionality \change{and depends on input complexity.} Rather than searching for global solutions, DDP finds a local solution \cite{mayne}. These methods are computationally efficient but are highly sensitive to initialization, which limits \change{their} application to simple tasks.
However, recent work proposes a feasibility-driven DDP (FDDP) algorithm improves \rev{the convergence under poor initialization \cite{crocoddyl} enabling us to compute motions subject to contact constraints.} DDP-based approaches provide both feed-forward actions and feedback gains within the optimization horizon. Both elements enable our system to track the optimal policy, which increases performance as shown in \cite{mastalli22mpc}. The FDDP algorithm is efficiently implemented in \change{the} \textsc{Crocoddyl} library.
Similarly, as described in \cite{boxfddp}, the Box-FDDP algorithm handles box constraints on the control variables and uses a feasibility-driven search. \rev{Both FDDP/Box-FDDP increases the basin of attraction to local minima and the convergence rate when compared to the DDP algorithm.}

DDP and its variants have been used in \new{the }planning and control of robots with soft actuators. For instance, we can execute explosive tasks with \change{VSAs} using the iLQR algorithm \cite{ilqr_explosive}. Similarly, we can \change{apply} DDP to describe a hybrid formulation \change{for} robots with soft actuators \cite{hybrid_ilqr}. Both works \change{demonstrate} the benefits of \new{modeling} their VSAs in highly dynamic tasks like jumping hopper and brachiation. 
But two major \change{drawbacks} of these approaches are \change{their} dependence on numerical differentiation, which increases computational time, and the lack of feedback terms, which \change{decreases} performance.
To \change{analytically compute} the derivatives of rigid systems, \cite{analyticalD} exploits the induced kinematic sparsity pattern in the kinematic tree. This method reduces the computation time obtained using \change{other} common techniques: automatic or numerical differentiation. It is possible to use the tools developed for the rigid body case and tailor \change{them} for applications related to \rev{systems with soft actuators}. This will be beneficial for both \change{the} online deployment of the algorithms and \change{the} control \change{of} systems with high degrees of freedom. \change{Secondly, in \cite{mastalli22mpc}, the feedback policy obtained from DDP, is employed instead in place of a user- tuned tracking controller. The results show that the local feedback policy obtained from DDP could be a promising solution for state feedback.}

\section{Problem Definition}\label{sec:method}
In this section, we formulate the optimal control problem to plan a desired task with an articulated soft robot with a fixed-base and \change{without any contacts.}  %
\subsection{Motivational example}\label{sec:motivation}
\begin{figure*}
    \subfigure[rigid robot: joint motion.]{\includegraphics[width=.245\linewidth]{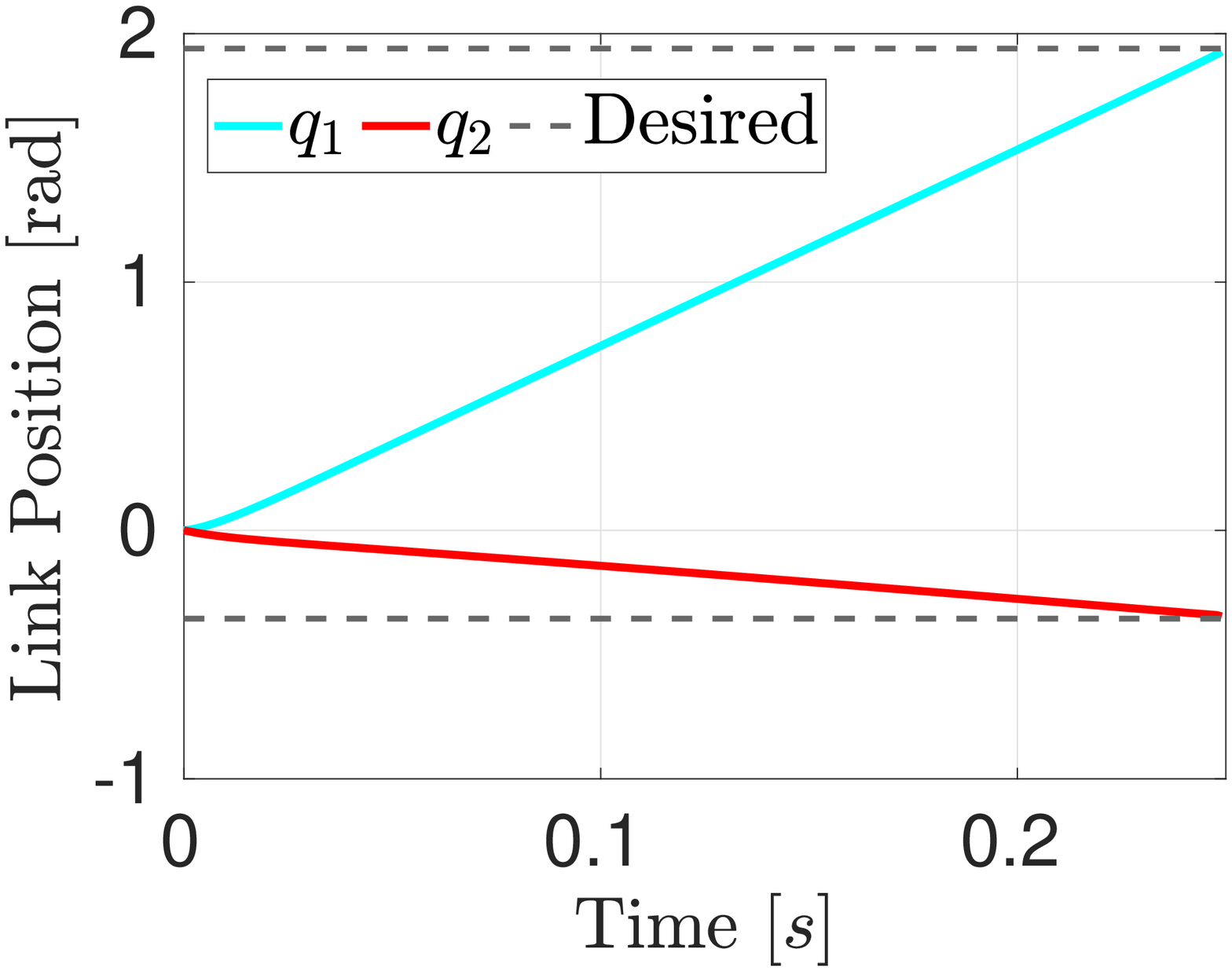}\label{fig:ra}}
    \subfigure[ASR (10\unit{\newton\meter/\radian}): joint motion.]{\includegraphics[width=.245\linewidth]{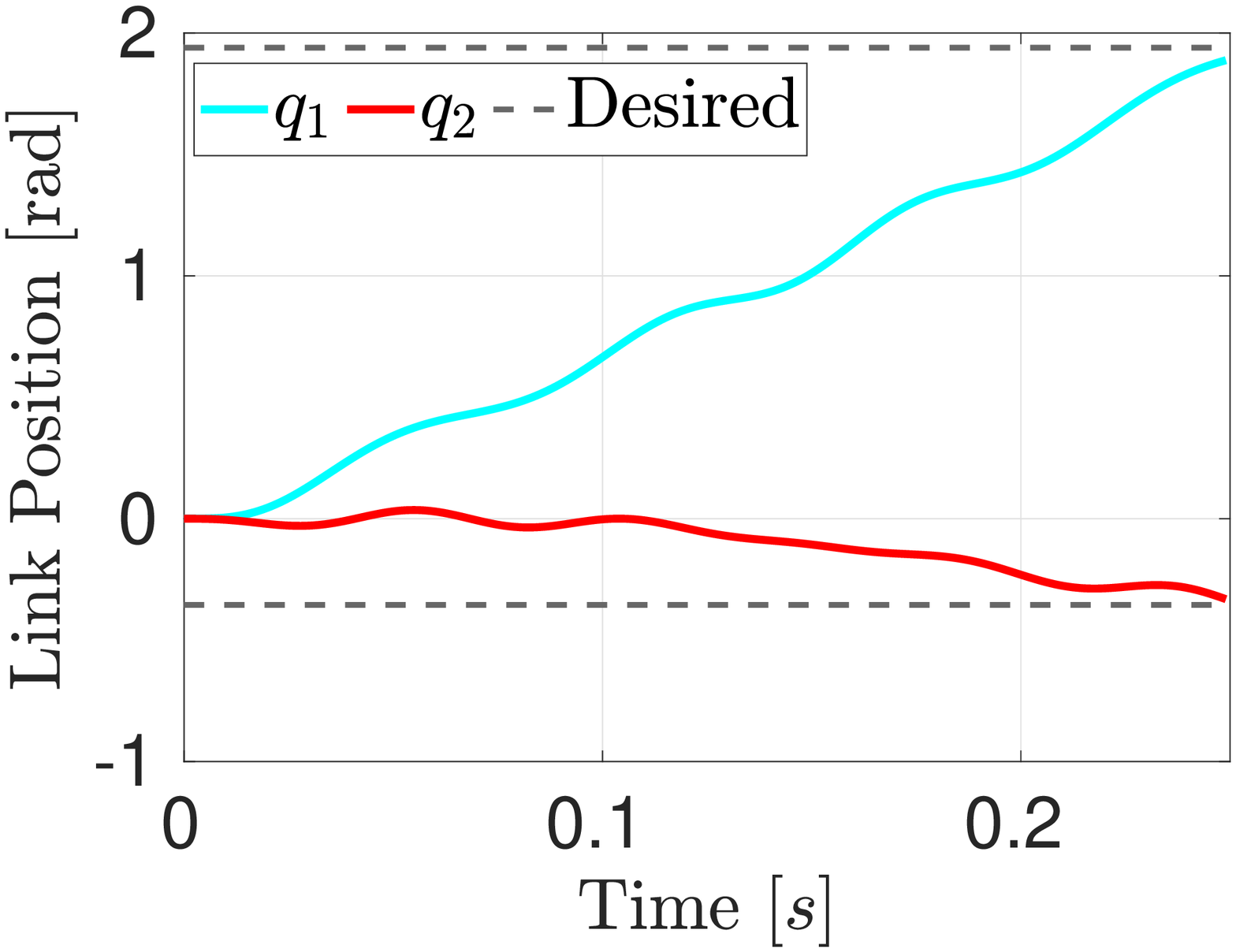}\label{fig:sa_1}}
    \subfigure[ASR (3\unit{\newton\meter/\radian}): joint motion.]{\includegraphics[width=.245\linewidth]{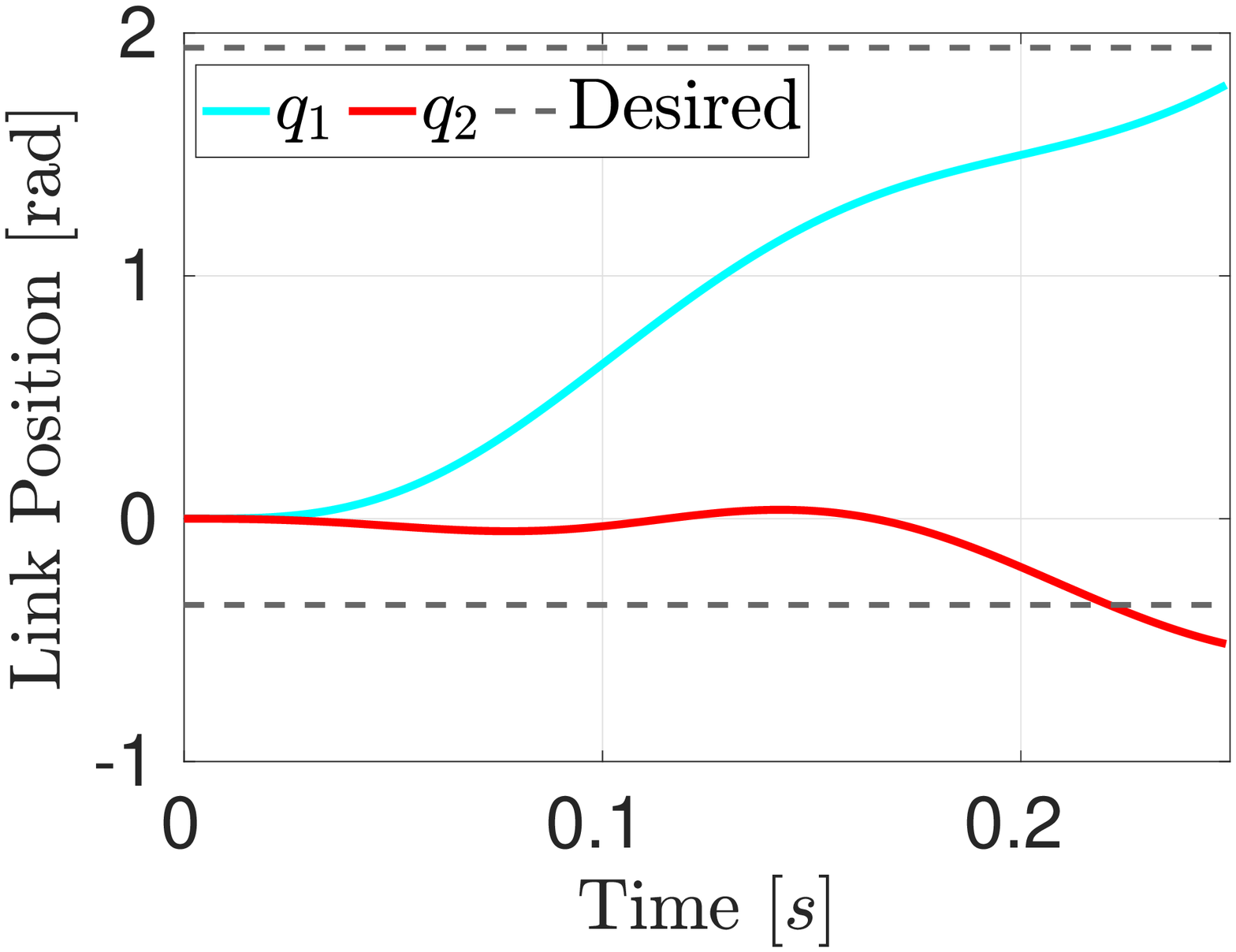}\label{fig:sa_2}}
    \subfigure[ASR (0.01\unit{\newton\meter/\radian}): joint motion.]{\includegraphics[width=.245\linewidth]{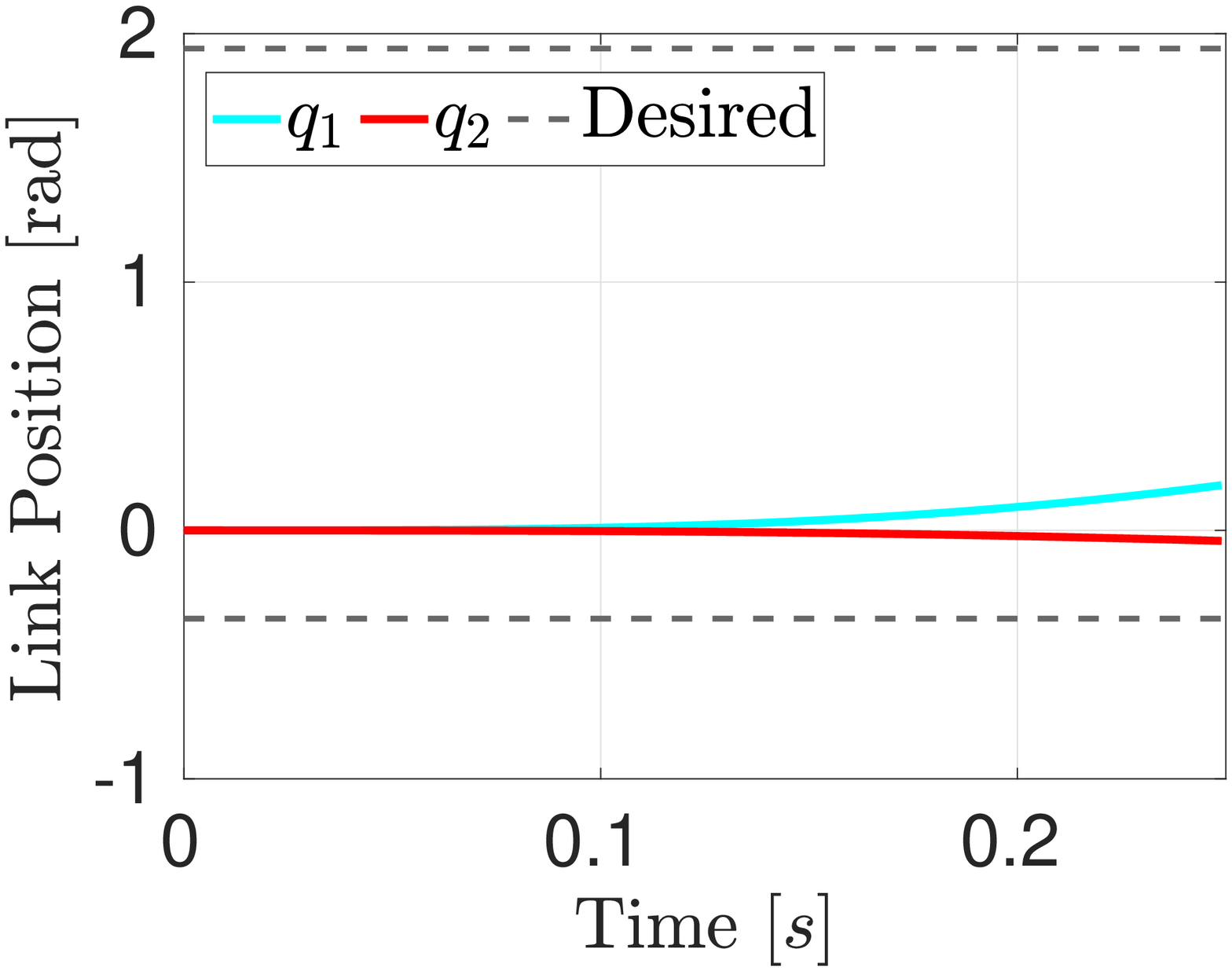}\label{fig:sa_3}} \\
    \vspace{-0.3cm}
    
    \subfigure[rigid robot: Cart. motion.]{\includegraphics[width=.245\linewidth]{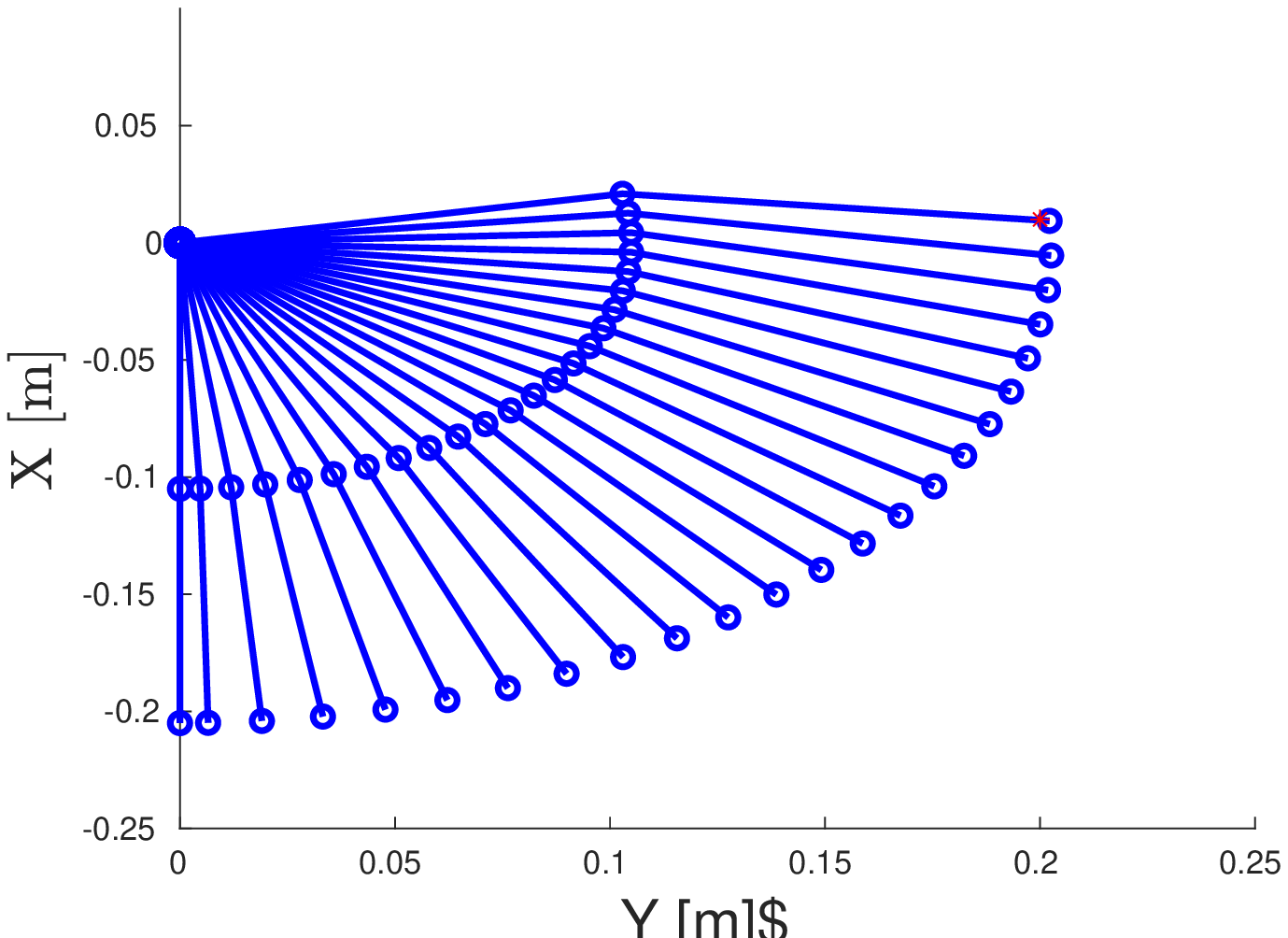}\label{fig:ra_r}}
    \subfigure[ASR (10\unit{\newton\meter/\radian}): Cart. motion.]{\includegraphics[width=.245\linewidth]{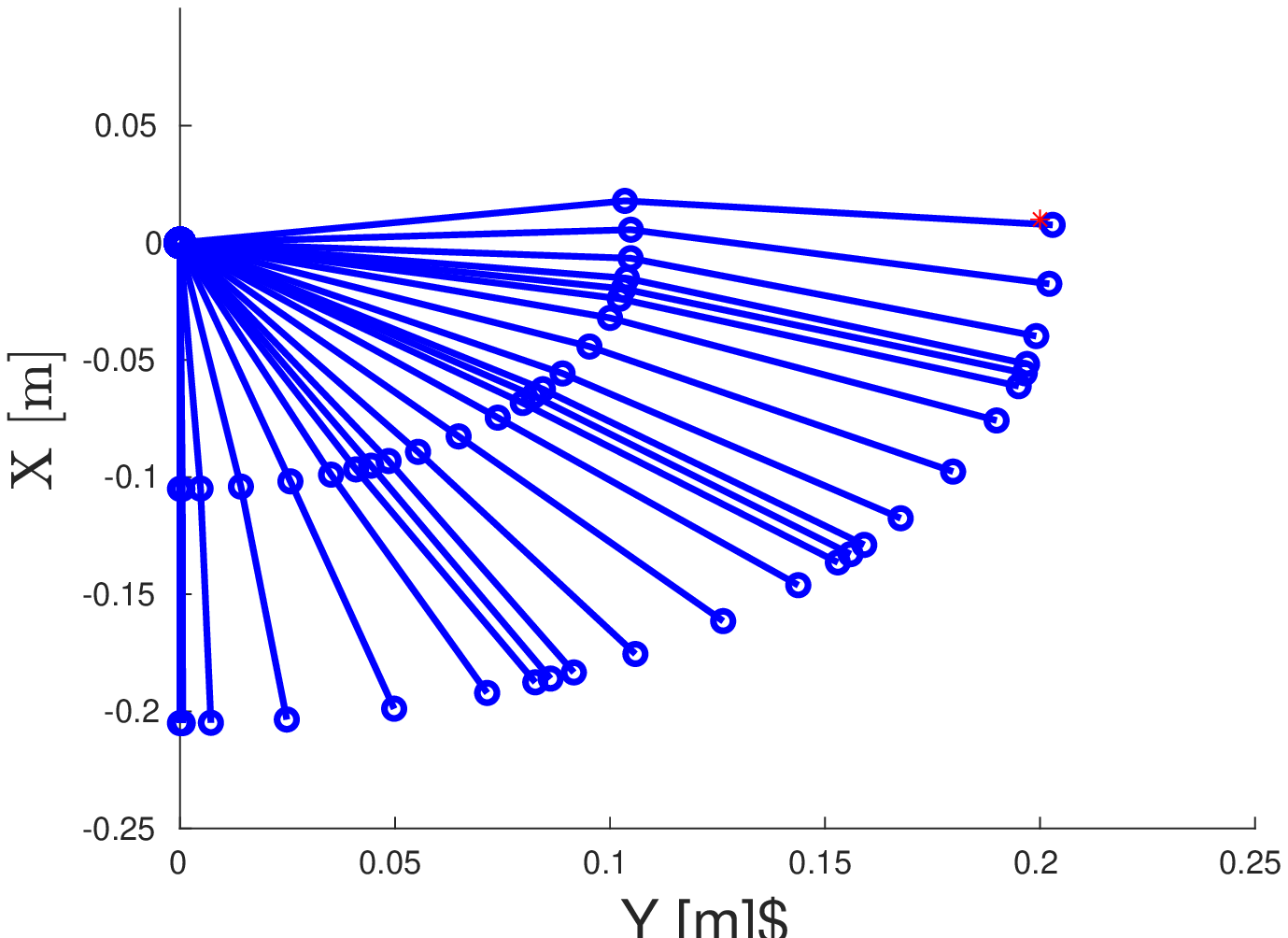}\label{fig:sa_1_r}}
    \subfigure[ASR (3\unit{\newton\meter/\radian}): Cart. motion.]{\includegraphics[width=.245\linewidth]{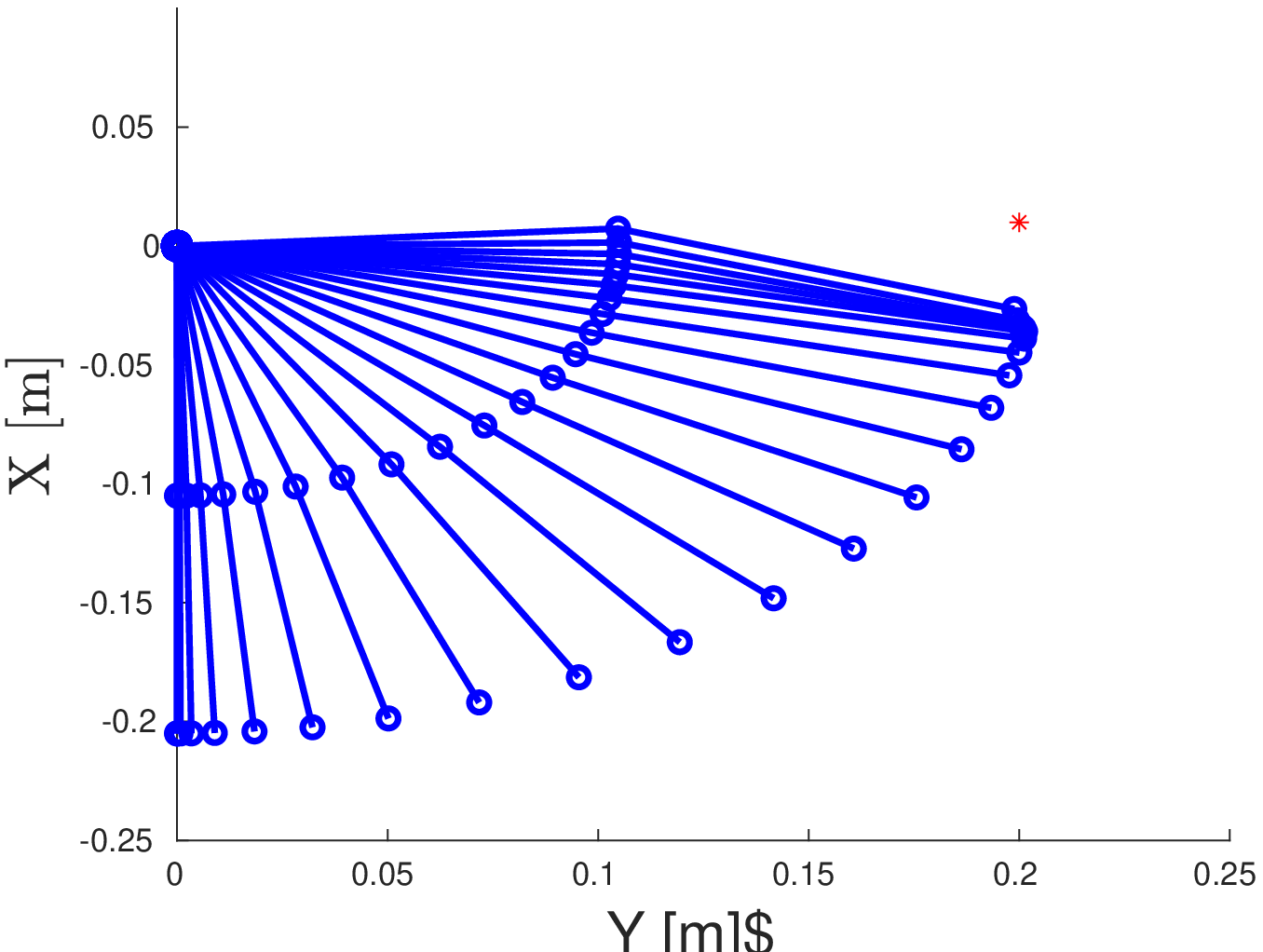}}
    \subfigure[ASR (0.01\unit{\newton\meter/\radian}) Cart. motion.]{\includegraphics[width=.245\linewidth]{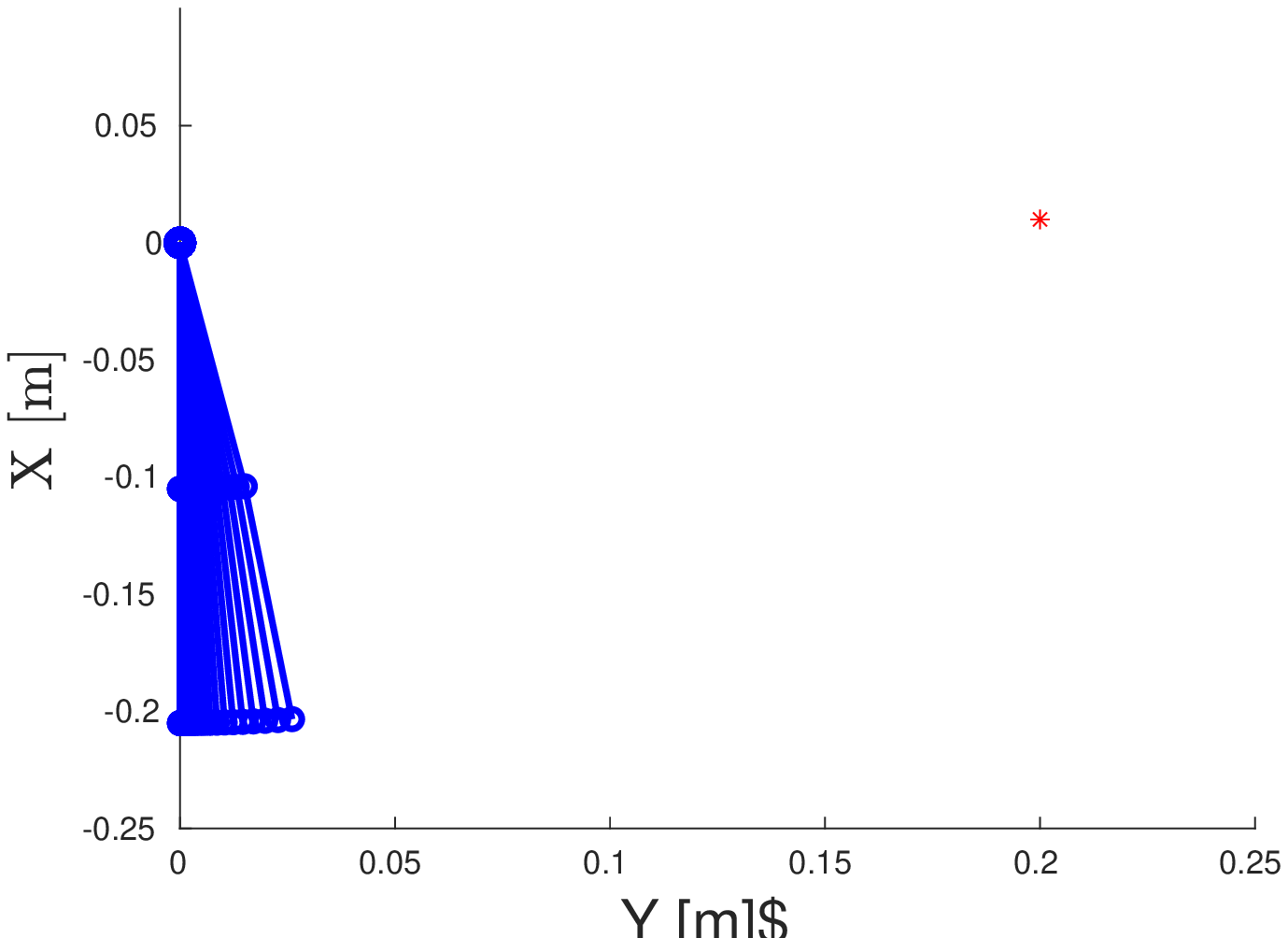}} 
    \caption{\new{Motivational example: a 2DoF robot affected by gravity performing a regulation task with the desired final end-effector position equal to $[0.01,~0.2]$ m. The top row shows the joint evolution, the bottom row shows the Cartesian evolution. The dashed lines in the plots of top row indicate the desired final joint angles and the red dot in the plots of bottom row indicates the desired position.
    The control input is obtained considering the robot as rigid (a),(e); in this case, the robot reaches the final position $[0.009,~ 0.2034]$ \unit{\meter}. Then, the same control input is applied to three 2DoF articulated soft robots (ASR) with different joint stiffness values. In (b),(f) the joint stiffness is $10$ \unit{\newton\meter/\radian}, and the the final end-effector position is $[0.012, 0.204]$ \unit{\meter}. In (c),(g) the joint stiffness is $3$ \unit{\newton\meter/\radian}, and the the final end-effector position is $[0.04, 0.21]$ \unit{\meter}. In (d),(h) the joint stiffness is $0.01$ \unit{\newton\meter/\radian}, and the the final end-effector position is $[-0.0156,~ 0.2023 ]$ \unit{\meter}. These results highlight the limit of modeling soft robot links as rigid models.}
}
\label{fig:ra_vs_sa}
\end{figure*}

Soft robots present a model with larger state space dimension compared to rigid robots with the same number of degrees of freedom (DoF). Therefore, including the soft robot model into the optimal control problem inevitably increases the computational load, which is caused by operations like dynamics computation and other such operations part of the optimal control routine.
Thus, it is natural to question if we really need to use the \textit{soft models}. To answer this, we consider an end-effector regulation task. In this task, we command a 2DoF soft actuated system (the physical parameters of this 2DoF system are introduced in Section \ref{sec:simulation}) using an optimal control sequence that ignores the actuation dynamics (i.e., a \textit{rigid model}). The desired final position is $[0.01,~0.2]$ \unit{\meter}.
Fig. \ref{fig:ra}, \ref{fig:ra_r} show the optimal trajectory, and robot motion, respectively. When the same control sequence is applied to an articulated soft robot with a low stiffness value, we observe an inconsistent behavior, and the end-effector position at the end of the task is far from the desired point \ref{fig:sa_1}, \ref{fig:sa_2}, \ref{fig:sa_3}. We also observe that the same control sequence \new{shows} different performance when applied to systems with varying stiffness values. Thus the use of control solutions devised for a rigid actuated model may not work well for soft actuated systems and may prove to be inconsistent. 
\vspace{-.4cm}

\subsection{Model}
Consider a robot with an open kinematic chain with $n+1$ rigid links, and $n$ compliant joints. Let the link-side coordinates be $\textbf{q} \in \mathbb{R}^{n}$, link-side velocity be $\mathbf{\dot{q}}\in \mathbb{R}^{n}$, motor-side coordinates $\boldsymbol \theta \in \mathbb{R}^{m}$,  and motor-side velocity be $\boldsymbol{\dot{\theta}} \in \mathbb{R}^{m}$.

\rev{These \new{kinds} of systems usually present large reduction ratios, and the angular velocity of the rotor is due only to  their own spinning.
Therefore, the energy contributions due to inertial couplings between the motors and link can be neglected. Given this observation, we assume the following:} 
\begin{assume}\label{assume_1}
\rev{We assume that the inertial coupling between the rigid body and the motor is negligible.}
\end{assume}
Under Assumption \ref{assume_1}, using the Lagrangian formulation for the coupled system, one can  derive the equations of motion as~\cite{albu2016actuators},
\begin{align}\label{eq:init_dyn1} \nonumber
&\mathbf{M(q}(t))\mathbf{\ddot{q}}(t) + \mathbf{C(q}(t),\mathbf{\dot{q}}(t))\mathbf{\dot{q}}(t) \\ 
&\hspace{5.5em} + \mathbf{G(q}(t)) +\frac{\partial \mathbf{U}(\mathbf{q}(t),\boldsymbol\theta(t))}{\partial \mathbf{q}(t)}^\top = \mathbf{0}\\
&\hspace{3em}\mathbf{B} \boldsymbol{\ddot{\theta}}(t) +  \frac{\partial \mathbf{U}(\mathbf{q}(t),\boldsymbol\theta(t))}{\partial \boldsymbol\theta(t)}^\top -  \boldsymbol{\tau}(t)=\mathbf{0},
\label{eq:init_dyn2}
\end{align}
where, $\mathbf{M}(\mathbf{q}(t)) \in \mathbb{R}^{n\times n}$ is the robot inertia matrix, $\mathbf{C}(\mathbf{q}(t),\mathbf{\dot{q}}(t)) \in \mathbb{R}^{n\times n}$ contains the centripetal and Coriolis terms,  and $\mathbf{G}(\mathbf{q}(t)) \in \mathbb{R}^{n}$ is the gravity term, $\mathbf{B} \in \mathbb{R}^{m\times m}$ is the motor inertia, $\mathbf{U}(\mathbf{q}(t),\boldsymbol\theta(t))$ is the elastic potential, and $\boldsymbol \tau(t) \in  \mathbb{R}^{m}$ is the  torque. 
\rev{The general nonlinear characterization of the motor-side can be considered but we operate in the linear region of the deflection. Thus we assume that:}
\begin{assume} \label{assume_2}
\rev{The elastic coupling is linear in $\mathbf{q}$ and $\boldsymbol\theta$.}
\end{assume}

\textcolor{black}{Using Assumption \ref{assume_2}, the torque due to elastic potential is linear i.e. $\frac{\partial \mathbf{U}(\mathbf{q}(t),\boldsymbol\theta(t))}{\partial \mathbf{q}(t)}^\top = \mathbf{K}(t) (\mathbf{q}(t) - \mathbf{S}\boldsymbol\theta(t))$ and 
	$\frac{\partial\mathbf{U}(\mathbf{q}(t),\boldsymbol\theta(t))}{\partial \boldsymbol\theta(t)}^\top = \mathbf{S}^\top  \mathbf{K}(t){(\mathbf{S}\boldsymbol\theta(t) - \mathbf{q}(t))}$.} \rev{Here $\mathbf{K}(t)$ is a stiffness matrix and $\mathbf{S} \in \mathbb{R}^{n\times m}$ is the selection matrix. The selection matrix $\mathbf{S}$ is of rank $m$. 
	The stiffness matrix $\mathbf{K}$ can be either constant or time-varying  corresponding to SEA and VSA, respectively. In the case of SEA, the stiffness of each actuated joint is fixed to some $\sigma$. In the case of VSA,  the stiffness of each actuated joint can vary between $\sigma_{\textnormal{min}}$ and $\sigma_{\textnormal{max}}$ and to maintain positivity of the spring stiffness we impose $\sigma_\textnormal{min}>0$. \new{Similarly},  under-acutated compliant arm refers to the systems with the rank of selection matrix (rank($\mathbf{S}$)) being less than $m$ and the joints can be either actuated by SEA/VSA.
	
	Now, using the linearity of elastic coupling,   \eqref{eq:init_dyn1}-\eqref{eq:init_dyn2}
	reduce to,}
	\begin{align}\label{eq:flex_dyn1}
	\begin{split}
	&\mathbf{M(q(}t))\mathbf{\ddot{q}}(t) + \mathbf{C(q}(t), \mathbf{\dot{q}}(t))\mathbf{\dot{q}}(t) +\\ &\hspace{5.5em}\mathbf{G(q}(t)) +\mathbf{K}(t)(\mathbf{q}(t)-\mathbf{S}\boldsymbol\theta(t)) = \mathbf{0},
	\end{split}\\
	\begin{split}\label{eq:flex_dyn2}
	&\hspace{2em}\mathbf{B \ddot{\boldsymbol{\theta}}}(t) +  \mathbf{S}^\top \mathbf{K}(t){(\mathbf{S}\boldsymbol\theta(t) - \mathbf{q}(t)})  -  \boldsymbol{\tau}(t)=\mathbf{0}.
	\end{split}
	\end{align}

\rev{ It is worth mentioning that the model class in  \eqref{eq:flex_dyn1}-\eqref{eq:flex_dyn2} \new{is} also used to model flexible link robots in some state of the art papers \cite{flexible_link_mich, flex_deluca, prb_1, prb_2} and in some soft robot simulators \cite{pysomo}.
	
In the case where all the joints are actuated, \textbf{S} is \new{the }identity matrix, and $n=m$, then  \eqref{eq:flex_dyn1}-\eqref{eq:flex_dyn2} can be written as}

	\begin{align}
	\label{eq:full_dyn1}
	\begin{split}
	&\mathbf{M(q}(t))\mathbf{\ddot{q}}(t) + \mathbf{C(q}(t), \mathbf{\dot{q}}(t))\mathbf{\dot{q}}(t) +\\ &\hspace{5.5em}\mathbf{G(q}(t)) +\mathbf{K}(t)\mathbf{q}(t)-\boldsymbol\theta(t)) = \mathbf{0},
	\end{split}\\
	\begin{split}\label{eq:full_dyn2}
	&\hspace{2em}\mathbf{B} \boldsymbol{\ddot{\theta}}(t) +  \mathbf{K}(t){(\boldsymbol\theta(t) - \mathbf{q}(t))}  -  \boldsymbol{\tau}=\mathbf{0}.
	\end{split}
	\end{align}

\rev{The rotors of the actuators are designed with their COM on the rotor axis to extend the life of the electrical drives. The motor inertia matrix is diagonal as a result of this. 
Further, the stiffness matrix should be invertible to ensure consistent solutions to \eqref{eq:init_dyn1}-\eqref{eq:full_dyn2}. }

\begin{assume}\label{assume_3}
\rev{The motor inertia matrix is  diagonal.}
\end{assume}
\rev{Using Assumption \ref{assume_3}, $\mathbf{B}(t)$ can be written as:
 \begin{align}
\mathrm{\textbf{B}(t)_{i,j}} = \begin{cases} 
B_i & \text{if $i = j$} \\  
0 & \text{if $i\neq j$}  
\end{cases} 
\end{align}}

\rev{Further to simplify computation, $\mathbf{K}(t)$ is diagonal and this resembles the case where one spring is coupled between a rotor and a link.}
Thus, $\mathbf{K}(t)$ can be written as,
 \begin{align}
 \mathrm{\textbf{K}(t)_{i,j}} = \begin{cases} 
 \sigma_i & \text{if $i = j$} \\  
 0 & \text{if $i\neq j$} \\  
 \end{cases} 
\end{align}
In the following, for the sake of simplicity, we will omit the explicit time dependence. 
\vspace{-.2cm}
\subsection{Goals}
\rev{Using an optimal control approach,} we aim to solve dynamic tasks for robots actuated by SEA, VSA and \change{underactuated compliant} robots. \rev{In this case}, the forward dynamics will be determined by \eqref{eq:flex_dyn1}-\eqref{eq:flex_dyn2} or
 \eqref{eq:full_dyn1}-\eqref{eq:full_dyn2}. Additionally, we aim to exploit feedback gains to increase performance and stabilization properties.
 The tasks presented in this paper are end effector regulation tasks for SEA/VSA and swing-up for the \change{underactuated compliant} systems case. 
\subsection{Optimal control formulation}\label{sec:optimal_control_problem}
We formulate a discrete-time optimal control problem for soft robots as \rev{follows}:
\begin{align}\label{eq:fwddyn_oc}\nonumber
\begin{split}
&{\underset{(\mathbf{q}_s,\mathbf{\dot{q}}_s,\boldsymbol{\theta}_s, \boldsymbol{\dot{\theta}}_s),(\boldsymbol{\tau}_s)}{\min}} ~ \ell_N(\mathbf{q}_N,\mathbf{\dot{q}}_N,\boldsymbol{\theta}_N,\boldsymbol{\dot{\theta}}_N) \\
&\hspace{4.75em} + \sum_{k=0} ^{N-1}\int_{t_k}^{t_{k+1}} \ell_k(\mathbf{q}_k,\mathbf{\dot{q}}_k, \boldsymbol{\theta}_k,\boldsymbol{\dot{\theta}}_k,\boldsymbol{\tau}_k) dt \\\nonumber
&\textrm{s.t.} ~~ [\mathbf{q}_{k+1},\mathbf{\dot{q}}_{k+1},\boldsymbol{\theta}_{k+1},\boldsymbol{\dot{\theta}}_{k+1}] = \boldsymbol{\psi}(\mathbf{\dot{q}}_k,\mathbf{\ddot{q}}_k,\boldsymbol{\dot{\theta}}_k,\boldsymbol{\ddot{\theta}}_k),\\\nonumber
&\hspace{6.75em}[\mathbf{\ddot{q}}_k,\boldsymbol{\ddot{\theta}}_k]= \mathrm{FD}(\mathbf{q}_k,\mathbf{\dot{q}}_k,\boldsymbol{\theta}_k,\boldsymbol{\dot{\theta}}_k,\boldsymbol{\tau}_k),\\\nonumber
&\hspace{6.75em}[\mathbf{q}_k,\boldsymbol\theta_k]\in\mathcal{Q}, [\mathbf{\dot{q}}_k,\boldsymbol{\dot{\theta}}_k]\in\mathcal{V}, \boldsymbol{\tau}_k\in\mathcal{U},
\end{split},
\end{align}
where, $\mathbf{q}_k$, $\mathbf{\dot{q}}_k$, $\boldsymbol\theta_k$, $\boldsymbol{\dot{\theta}}_k$ and $\boldsymbol{\tau}_k$ describe the configuration point, generalized velocity, motor-side angle, motor-side velocity, joint torque commands of the  system at \change{time-step (node) $k$};  $\ell_N$ is the terminal cost function; $\ell_k$ is the \rev{running} cost function; $\boldsymbol{\psi}(\cdot)$ defines the integrator function; $\mathrm{FD}(\cdot)$ represents the forward dynamics of the soft robot; $\mathcal{Q}$ represents the admissible state space; $\mathcal{V}$ \rev{describes} the admissible velocity space and $\mathcal{U}$ defines the allowed control.

\section{Solution}\label{sec:solution}
We solve the optimal control problem described in Section  \ref{sec:optimal_control_problem} using the Box-FDDP algorithm. This section first summarizes the Box-FDDP algorithm, which is a variant of \rev{the} DDP algorithm, and then \rev{analyzes} the dynamics and  analytical derivatives of robots with SEAs, VSAs, and the \change{underactuated compliant} robots. To account for the cost incurred by the mechanism implementing the variable stiffness in VSAs, we also introduce a cost function used in systems actuated by VSA. Finally, we describe the state-feedback controller derived from  Box-FDDP.
We would like to \rev{emphasize} that Box-FDDP is developed in \cite{boxfddp} and is not a novel contribution of this work. 

\rev{iLQR/DDP methods are known to be prone to numerical instabilities as these are single-shooting methods. Whereas, FDDP is a multiple shooting method and thus provides numerical benefits like better numerical stability. The feasibility-driven search and the nonlinear roll-out features of the algorithm ensure better convergence under poor initialization, enabling better performance for highly nonlinear problems compared to iLQR/DDP \cite{crocoddyl}. Box-FDDP is a variant of the FDDP algorithm which can handle box constraints on control variables. Box-FDDP is a more general algorithm that is based on projected Newton updates to account for the box constraints on control variables. Box-FDDP reduces to Newton updates of FDDP in the case without box constraints on the control variables. The method also provides a locally optimal feedback policy which is expected to improve performance in various tasks. 
This ability to handle optimal control problems for highly nonlinear systems with the option of unfeasible guess trajectory and the synthesis of feedback policies makes FDDP/Box-FDDP a suitable candidate for articulated soft robots. 
}

\subsection{Background on Box Feasibility-Driven DDP} \label{sec:ddp}
DDP solves optimal control problems by breaking down the original problem into smaller sub-problems. So instead of finding the entire trajectory at once, it recursively solves the Bellman optimal equation backwards in time. To handle control bounds and improve globalization properties, the Box-FDDP algorithm modifies the backward and forward passes of DDP.

The Bellman relation is stated as
\begin{equation}\label{eq:bellman}
    V(\mathbf{x}_k) = \underset{\mathbf{u}_k}{\min}~ \ell_k(\mathbf{x}_k,\mathbf{u}_k) + V_{k+1}(\mathbf{f}(\mathbf{x}_k,\mathbf{u}_k)),
\end{equation}
where, $V(\mathbf{x}_k)$ is the value function at the node $k$, $V_{k+1}(\mathbf{f}(\mathbf{x}_k,\mathbf{u}_k))$ is the Value function at the node $k+1$, $\ell$ is the one step cost, $\mathbf{x}$ is the state vector ($\mathbf{x} \triangleq [\mathbf{q}^\top,\dot{\mathbf{q}}^\top,\boldsymbol{\theta}^\top,\dot{\boldsymbol{\theta}}^\top]^\top$), $\mathbf{u}$ is the control vector and $\mathbf{f(x,u)}$ represents the dynamics of the system.

FDDP uses a quadratic approximation of \rev{the} differential change in  \eqref{eq:bellman}
\begin{align}\label{eq:eq_qp}
\Delta V=\underset{\delta\mathbf{u}_k}{\min} ~ \frac{1}{2}
\begin{bmatrix}
\delta\mathbf{x}_k \\ \delta\mathbf{u}_k
\end{bmatrix}^\top 
&\begin{bmatrix}
\mathbf{Q}_{\mathbf{xx}_k} & \mathbf{Q}_{\mathbf{xu}_k} \\
\mathbf{Q}_{\mathbf{ux}_k} & \mathbf{Q}_{\mathbf{uu}_k}
\end{bmatrix}
\begin{bmatrix}
\delta\mathbf{x}_k \\ \delta\mathbf{u}_k
\end{bmatrix} \\ \nonumber
&+\begin{bmatrix}
\delta\mathbf{x}_k \\ \delta\mathbf{u}_k
\end{bmatrix}^\top
\begin{bmatrix}
\mathbf{Q}_{\mathbf{x}_k} \\ \mathbf{Q}_{\mathbf{u}_k}
\end{bmatrix}.
\end{align}
$\mathbf{Q}$ is the local approximation of the action-value function  and its derivatives are
\rev{
	\begin{eqnarray}\label{eq:hamiltonian_computation}\nonumber
	\mathbf{Q}_{\mathbf{xx}_k} = \ell_{\mathbf{xx}_k} + \mathbf{f}^\top_{\mathbf{x}_k} V_{\mathbf{xx}_{k+1}} \mathbf{f}_{\mathbf{x}_k}, & & \nonumber
	\mathbf{Q}_{\mathbf{x}_k} = \ell_{\mathbf{x}_k} + \mathbf{f}^\top_{\mathbf{x}_k} V^{+}_{\mathbf{x}_{k+1}}, \\ \nonumber
	\mathbf{Q}_{\mathbf{uu}_k} = \ell_{\mathbf{uu}_k} + \mathbf{f}^\top_{\mathbf{u}_k} V_{\mathbf{xx}_{k+1}} \mathbf{f}_{\mathbf{u}_k}, & & \nonumber
	\mathbf{Q}_{\mathbf{u}_k} = \ell_{\mathbf{u}_k} + \mathbf{f}^\top_{\mathbf{u}_k} V^{+}_{\mathbf{x}_{k+1}}, \\ \nonumber
	\mathbf{Q}_{\mathbf{xu}_k} = \ell_{\mathbf{xu}_k} + \mathbf{f}^\top_{\mathbf{x}_k} V_{\mathbf{xx}_{k+1}}\mathbf{f}_{\mathbf{u}_k}, \\ \nonumber
	\end{eqnarray}
	
\rev{where, $V^{+}_{\mathbf{x}_{k+1}} = V_{\mathbf{x}_{k+1}} + V_{\mathbf{xx}_{k+1} }\mathbf{\bar{f}}_{k+1}$} is the Jacobian of the value function, $\ell_{\mathbf{x}_k}$ is the Jacobian of the one step cost, $\ell_{\mathbf{xx}_k}$ is the Hessian  of the one step cost and $\mathbf{\bar{f}}_{k+1}$ is the deflection in the dynamics at the node $k+1$:}
\begin{equation*}
    \mathbf{\bar{f}}_{k+1} = \mathbf{f}(\mathbf{x}_k,\mathbf{u}_k) - \mathbf{x}_{k+1}.
\end{equation*}

\subsubsection{Backward Pass}
In the backward pass, the search direction is computed by recursively solving

\begin{align} \label{eq:delta_u}
\begin{split}
         \mathbf{\delta u}_k = &\underset{\delta \mathbf{u}_k}{\arg \min}~\mathbf{Q}(\delta \mathbf{x}_k, \delta \mathbf{u}_k) = \mathbf{\hat{k}} + \mathbf{\hat{K}\delta x}_k, \\
          &\textrm{s.t.} ~~~\underline{\mathbf{u}} \leq \mathbf{u}_k + \delta \mathbf{u}_k \leq \overline{\mathbf{u}},
\end{split}
\end{align}
where, $\mathbf{\hat{k}} = -\mathbf{\hat{Q}}^{-1}_{\mathbf{uu}_k} \mathbf{Q}_{\mathbf{u}_k}$ is the feed-forward term and $\mathbf{\hat{K}} = - \mathbf{\hat{Q}}^{-1}_{\mathbf{uu}_k} \mathbf{Q}_{\mathbf{ux}_k}$ is the feedback term at the node $k$, and $\mathbf{\hat{Q}}_{\mathbf{uu}_k}$ is the control Hessian of the free subspace.
Using the optimal $\mathbf{\delta u}_k$, the gradient and Hessian of the Value function are updated.

\subsubsection{Forward Pass}
Once the search direction is obtained in \eqref{eq:delta_u}, then the step size $\alpha$ is chosen based on an Armijo-based line search routine. The control and state trajectory are updated using this step size
\begin{align}
    &\mathbf{\hat{u}}_k = \mathbf{u}_k + \alpha \mathbf{\hat{k}} + \mathbf{\hat{K}}(\mathbf{\hat{x}}_k - \mathbf{x}_k),\label{eq:feedback}\\
    &\mathbf{\hat{x}}_{k+1} = \mathbf{f}_k (\mathbf{\hat{x}}_k,\mathbf{\hat{u}}_k)- (1-\alpha) \bar{\mathbf{f}}_{k-1},
\end{align}
where, $\{\mathbf{\hat{x}}_k,\mathbf{\hat{u}}_k\}$ are the state and control vectors. In problems without control bounds, the algorithm reduces to FDDP \cite{crocoddyl}.
The interested reader is referred to \cite{boxfddp,crocoddyl} for more details about the algorithm.

\subsection{Dynamics for soft robots}
The forward dynamics in \eqref{eq:full_dyn1}-\eqref{eq:full_dyn2} can be written in compact form as \rev{follows:}
\begin{equation}\label{eq:free_fwddyn}
\left[\begin{matrix}\mathbf{\ddot{q}} \\ \boldsymbol{\ddot{\theta}} \end{matrix}\right] =
\left[\begin{matrix}\mathbf{M} &  \mathbf{0} \\ \mathbf{0} & \boldsymbol{B}\\ \end{matrix}\right]^{-1}
\left[\begin{matrix}\boldsymbol{\tau}_l  \\ \boldsymbol{\tau}_m\end{matrix}\right],
\end{equation}
where, 
\begin{align}\label{eq:tau_def}
&\boldsymbol{\tau}_l \triangleq {-\mathbf{C(q,\dot{q}}) -\mathbf{G}(\mathbf{q}) -\mathbf{K}(\mathbf{q}-\boldsymbol\theta) },\\
&\boldsymbol{\tau}_m \triangleq \mathbf{ K}(\boldsymbol\theta-\mathbf{q}) + \boldsymbol\tau. 
\end{align}
We compute link-side dynamics efficiently \rev{via} the use articulated body algorithm (ABA) for the first block of effective inertia matrix (which corresponds to the rigid body algorithm). We then use the analytical inversion of $\mathbf{B}$ to efficiently compute the motor-side dynamics in  \eqref{eq:free_fwddyn}.
\\

 The forward dynamics computation in  \eqref{eq:flex_dyn1}-\eqref{eq:flex_dyn2}  can be done similarly to the above process with a different definition of $\boldsymbol \tau_l$ and $\boldsymbol \tau_m$
\begin{align}\label{eq:tau_flex_def}
&\boldsymbol{\tau}_l \triangleq {-\mathbf{C(q,\dot{q}}) -\mathbf{G}(\mathbf{q}) -\mathbf{K}(\mathbf{q}-\mathbf{S}\boldsymbol\theta) },\\
&\boldsymbol{\tau}_m \triangleq \mathbf{-S^\top K}(\mathbf{S}\boldsymbol\theta-\mathbf{q}) + \boldsymbol\tau. 
\end{align}

\rev{The forward dynamics computation is summarized in Algorithm \ref{alg:fwd}: }

\begin{algorithm}
	\rev{	
	\caption{Forward dynamics}
	\label{alg:fwd}
	\begin{algorithmic}[1]
		\State \textbf{Input:} $\mathbf{\texttt{robotModel},q,\dot{q}, \boldsymbol\theta,\dot{\boldsymbol\theta}}$\newline
		
		\State \textbf{Output:} $\mathbf{\ddot{q}},\boldsymbol{\ddot{\theta}}$ \newline
		
		\State $\mathbf{\ddot{q}} \gets$ ~ Articulated Body Algorithm\texttt{(ABA)} ( $\mathbf{q},\mathbf{v}, \mathbf{0}) ~+ ~\mathbf{M}^{-1}(-\mathbf{K}(\boldsymbol{\theta} - \mathbf{q})) $
		\newline
		\State $\boldsymbol{\ddot{\theta}} \gets$ $\mathbf{B}^{-1}(\boldsymbol{\tau} + \mathbf{K}(\boldsymbol{\theta} - \mathbf{q}))$
	\end{algorithmic}}
\end{algorithm}

\rev{ $\mathbf{M}^{-1}$ is computed as part of the forward dynamics algorithm. Additionally, the computation of $\ddot{\mathbf{\theta}}$ involves inversion of $\mathbf{B}$ which in itself is diagonal in our cases.}

\subsection{Analytical derivatives}
The block diagonal structure of the inertia matrix in \eqref{eq:free_fwddyn} allows  us to independently evaluate the partial derivatives related to the link-side and motor-side.
Among several methods to compute the partial derivatives of the dynamics, the finite difference method is popular. This is because the difference between the input dynamics is computed  $n+1$  times while perturbing the input variables. The successful implementation of numerical differentiation requires fine parallelization techniques,  thus the finite difference could result in \change{computational complexity}  of  $\mathcal{O}(n^2)$ \cite{featherstone2014rigid}. Another way is to derive the Lagrangian equation of motion, which requires only one function call. We use \textsc{Pinocchio} \cite{pinocchio}, an efficient library for rigid body algorithms, which exploits sparsity induced by kinematic patterns to compute the analytical derivatives with $\mathcal{O}(n)$ cost.
Now we illustrate the analytical derivatives for SEA/VSA and the \change{underactuated compliant} model.

\subsubsection{Series elastic actuation}
To solve the full dynamics model in the Box-FDDP/FDDP formalism we define the state vector as
$\mathbf{x} \triangleq [\mathbf{q}^\top \; \mathbf{\dot{q}}^\top \; \boldsymbol{\theta}^\top \; {\boldsymbol{\dot{\theta}}}^\top]^\top$, and the input vector is $ \mathbf{u} = [\boldsymbol\tau]$.

An explicit inversion of the KKT matrix is avoided in the forward pass by inverting the matrix analytically: 
\begin{equation}\label{eq:jac_contact_fwddyn}
\left[\begin{matrix}\delta\mathbf{\ddot{q}} \\ \delta\boldsymbol{ \ddot{\theta}} \end{matrix}\right] =-\left[\begin{matrix}\mathbf{M} &  \mathbf{0} \\ \mathbf{0} & \boldsymbol{B}\\ \end{matrix}\right]^{-1}
\Big(\left[\begin{matrix}
\frac{\partial\boldsymbol{\tau}_l}{\partial\mathbf{x}} \\ \frac{\partial\boldsymbol{\tau}_m}{\partial \mathbf{x}}\end{matrix}\right] \delta \mathbf{x}+
\left[\begin{matrix}\frac{\partial\boldsymbol{\tau}_l}{\partial\mathbf{u}}  \\ \frac{\partial\boldsymbol{\tau}_m}{\partial \mathbf{u}}\end{matrix}\right] \delta \mathbf{u} \Big)\;.
\end{equation}
The matrix has a diagonal block structure and can be switched separately. The motor inertia matrix is diagonal for all practical purposes and thus can be analytically inverted.
Using the definition of $\boldsymbol{\tau}_l$ in $\eqref{eq:tau_def}$ one can analytically compute the Jacobians. Here we only list the non-zero components, i.e.,
\begin{align}\label{eq:dtau}
&\frac{\partial\boldsymbol{\tau}_l}{\partial \mathbf{q}} = -\frac{\partial\mathbf{C(q,\dot{q})}}{\partial \mathbf{q}} -\frac{\partial \mathbf{G(q)}}{\partial \mathbf{q}} -\mathbf{K}, \\
&\frac{\partial\boldsymbol{\tau}_l}{\partial\mathbf{\dot{q}}} =-\frac{\partial\mathbf{C}(\mathbf{q},\mathbf{\dot{q}})}{\partial \mathbf{\dot{q}}},  \hspace{2em} \frac{\partial \boldsymbol{\tau}_l}{\partial \boldsymbol \theta } =  - \mathbf{K},\\
&\frac{\partial \boldsymbol{\tau}_m}{\partial \mathbf{q}} = \mathbf{K} + \frac{\partial \boldsymbol\tau}{\partial \mathbf{q}}, \hspace{2em} \frac{\partial \boldsymbol{\tau}_m}{\partial \boldsymbol \theta} = \boldsymbol{K}. \label{eq:dtau_last} 
\end{align}

Similarly the Jacobian w.r.t. $\mathbf{u}$ is 
$\frac{\partial \boldsymbol{\tau_m}}{\partial \mathbf{u}} = \mathbf{I}$.
Using \rev{the} same principles, the analytical derivatives of \rev{the} cost function w.r.t. state and control can be derived.

\subsubsection{Variable stiffness actuation}
In the case of variable stiffness actuators, we model the system using similar equations, but stiffness at each joint is treated as a decision variable.
Thus the state vector is still $\mathbf{x} \triangleq [\mathbf{q}^\top \; \mathbf{\dot{q}}^\top \; \boldsymbol{\theta}^\top \; {\boldsymbol{\dot{\theta}}}^\top]^\top$,
but the decision vector is $\mathbf{u} \triangleq [\boldsymbol{\tau}^{\top} \boldsymbol\sigma^{\top}]^{\top}$ where, $\boldsymbol\sigma$ is the vector of diagonal entries from $\mathbf{K}$.
So the Jacobians w.r.t. the state variables \rev{remain} the same as \eqref{eq:dtau}-\eqref{eq:dtau_last}.
Now the derivatives w.r.t. decision vector are

\begin{align}\label{eq:vsadtau}
    \frac{\partial \boldsymbol{\tau}_m}{\partial \boldsymbol{\tau}} = \mathbf{I},  \hspace{2em} \frac{\partial \boldsymbol{\tau}_m}{\partial \boldsymbol\sigma} = \boldsymbol \theta - \mathbf{q}, \hspace{2em}
    \frac{\partial \boldsymbol{\tau}_l}{\partial \boldsymbol\sigma} =  \mathbf{q} - \boldsymbol \theta \;.
\end{align}

To effectively incorporate the constraint on $\boldsymbol\sigma$, we impose a box constraint on the stiffness variables
$    \sigma_{\textnormal{min}_i} < \sigma_i < \sigma_{\textnormal{max}_i}$, where $\sigma_i$ is the  \rev{$i$-th} component of the $\boldsymbol\sigma$ vector.
We use \rev{the} Box-FDDP algorithm \rev{in} \cite{box_ddp,boxfddp} to solve constrained optimal control problems with box constraints on control variables (Sec. \ref{sec:optimal_control_problem}).

\subsubsection{Under-actuated Compliant Arm} 
For \change{underactuated compliant systems}, $\mathbf{q} \in \mathbb{R}^{n}$ and $\boldsymbol\theta \in \mathbb{R}^m$ are of different dimensions.   
\rev{\new{Moreover}, their analytical derivatives are different from the fully actuated flexible joint case (i.e., \eqref{eq:dtau}-\eqref{eq:vsadtau}) as shown below:}
\begin{align}
    \frac{\partial \boldsymbol{\tau}_m}{\partial \boldsymbol{\theta}} = -\mathbf{S}^\top& \mathbf{K}\mathbf{S}  , \hspace{1em} 
    \frac{\partial \boldsymbol{\tau}_m}{\partial \mathbf{q}} = \mathbf{S}^\top \mathbf{K}, \hspace{1em}
    \frac{\partial \boldsymbol{\tau}_m}{\partial \boldsymbol\sigma} =  \mathbf{S}^\top(\mathbf{S}\boldsymbol\theta -\mathbf{q}), \\
    &\frac{\partial\boldsymbol\tau_l}{\partial\boldsymbol\theta} = \mathbf{K}\mathbf{S},
    \hspace{2em} \frac{\partial\boldsymbol\tau_l}{\partial\boldsymbol\sigma} = -(\mathbf{q}-\mathbf{S}\boldsymbol\theta)\;.
\end{align}

 \rev{The analytical derivatives of the dynamics \new{are} summarized in Algorithm \ref{alg:analytical_derivatives_1}:}

\begin{algorithm}
	\rev{
	\caption{Analytical Derivatives}
	\label{alg:analytical_derivatives_1}
	\begin{algorithmic}[2]
		\State \textbf{Input:} $\mathbf{\texttt{robotModel},q,\dot{q},\ddot{q},  \boldsymbol\theta,\dot{\boldsymbol\theta},\ddot{\boldsymbol\theta}}$\newline
		\State \textbf{Output:} $\frac{\partial \mathbf{\ddot{q}}}{\partial \mathbf{q}},\frac{\partial \mathbf{\ddot{q}}}{\partial \mathbf{\dot{q}}},\frac{\partial \mathbf{\ddot{q}}}{\partial \boldsymbol\theta},\frac{\partial \boldsymbol{\ddot{\theta}}}{\partial \mathbf{q}},\frac{\partial \boldsymbol{\ddot{\theta}}}{\partial \boldsymbol\theta},\frac{\partial \boldsymbol{\ddot{\theta}}}{\partial \boldsymbol{\dot{\theta}}}$ \newline
		\State $\frac{\partial \boldsymbol\tau_{rb}}{\partial \mathbf{q}},\frac{\partial \boldsymbol\tau_{rb}}{\partial \dot{\mathbf{q}}} \gets$ Compute Recursive Newton Euler algorithm(RNEA) derivatives$(\mathbf{q},\dot{\mathbf{q}},\ddot{\mathbf{q}})$ \newline
		\State $\frac{\partial \mathbf{\ddot{q}}}{\partial \mathbf{q}} = \mathbf{M}^{-1} (\frac{\partial\boldsymbol\tau_{rb}}{\partial \mathbf{q}} -\mathbf{K})$; \hspace{.3cm}  $\frac{\partial \mathbf{\ddot{q}}}{\partial \mathbf{\dot{q}}} = \mathbf{M}^{-1} (\frac{\partial \boldsymbol\tau_{rb}}{\partial \mathbf{\dot{q}}} ) $;\hspace{.5cm} $\frac{\partial \mathbf{\ddot{q}}}{\partial \boldsymbol\theta} = \mathbf{M}^{-1} (\mathbf{K})$;\newline 
		
		\State $\frac{\partial \boldsymbol{\ddot{\theta}}}{\partial \mathbf{q}} = \mathbf{B}^{-1} (\mathbf{K});$ \hspace{.5cm} $\frac{\partial \boldsymbol{ \ddot{\theta}}}{\partial \boldsymbol\theta} = -\mathbf{B}^{-1} \mathbf{(K)}$; \hspace{.5cm} $\frac{\partial \boldsymbol{\ddot{\theta}}}{\partial \boldsymbol{\dot{{\theta}}}} = \mathbf{0} $
	\end{algorithmic}}
\end{algorithm}

\subsection{VSA cost function}
The physical mechanism that implements variable stiffness consumes energy. To include this cost of changing stiffness \rev{in} the optimal control problem, we define a linear cost in stiffness\cite{variable_stiffness_cost}. 

\begin{align} \label{eq:stiffness_cost}
    \ell_{\textnormal{vsa}} = \sum_{j=1}^{m} \int_{0}^{T} \lambda (\sigma_j - \sigma_r) \textnormal{d}t\;,
\end{align}
where, $m$ is the number of actuated joints, $\sigma_r$ is the stiffness value under no-load conditions, $\sigma_j$ is the stiffness value of the joint $j$, and $\ell_{\textnormal{vsa}}$ is the one step VSA mechanism cost. Using this \rev{term}, we ensure that the cost is zero at $\sigma_r$ and \new{the} cost is imposed only when stiffness is varied. \rev{The overall cost value is dependent on the motor mechanics and it is modulated by $\lambda$.} The value of  $\lambda$ is related to the torque value required to maintain a particular stiffness value. To ensure that 
$\|\boldsymbol\tau\|_2^2 < \ell_{\textnormal{vsa}}$, the $\lambda$ is assumed to be a linear interpolation between the $\sigma_{\textnormal{min}}$ and $\sigma_{\textnormal{max}}$
\begin{align}
\lambda = \frac{g^2(\sigma_\textnormal{max})- g^2(\sigma_\textnormal{min})}{\sigma_\textnormal{max} -\sigma_\textnormal{min}} \nonumber\;,
\end{align}
where, $g^2(\cdot)$ is a function defined to represent the stiffness change for a specific actuator used. 

\rev{
	For variable stiffness actuator with an antagonastic mechanism such as \cite{qbmove}, $g^2(\cdot) = \tau_1^2 + \tau_2^2$.}
\new{The actual cost incurred in change of stiffness is related to the square of the torque curve and the design of the cost function \eqref{eq:stiffness_cost},  ensures  that it overestimates the cost incurred due to a change in stiffness \cite{variable_stiffness_cost}.}
\vspace{-.2cm}
\subsection{State feedback controller}
Feedback controllers based on PD gains are user tuned and sub-optimal. Instead of using sub-optimal policies to track \rev{the} optimal trajectory, we propose to both plan the optimal trajectory and use the local policy obtained from the backward pass of Box-FDDP/FDDP. In Section \ref{sec:ddp}, the feedback gain matrix is computed in the backward pass to ensure strict adherence to the optimal policy as described in~\cite{mastalli22mpc}. Thus the controller can be written as
\begin{equation}
    \mathbf{u} = \mathbf{\hat{k}} +\mathbf{\hat{K}} (\mathbf{x}^*-\mathbf{x}).
\end{equation}
The optimal control formulation, which considers the complete dynamics and costs, is used to calculate this local and optimal control policy. The feedback matrix for a VSA-actuated system is $\hat{\mathbf{K}} \in \mathbb{R}^{2m \times (n+m)}$, which also produces state feedback to the stiffness control alongwith the input torques. Furthermore, it is computationally less expensive than using a separate feedback controller.

\change{To calculate this local and optimal control policy, we formulate an} optimal control \change{problem that} considers the complete dynamics and costs. The feedback matrix for a VSA-actuated system is $\hat{\mathbf{K}} \in \mathbb{R}^{2m \times (n+m)}$, which also produces \change{state-feedback gains for} stiffness control along with the input torques. Furthermore, it is computationally less expensive than using a separate feedback controller.

\section{Validation Setup}
\label{sec:validation}
In this section, we introduce the simulation and experimental setup. Results and discussion will be presented in Section \ref{sec:results}.

\subsection{\change{System} setup}
\label{sec:simulation}

We employ six different compliant systems: (a) a 2DoF robot with SEAs at each joint; (b) a 2DoF robot with VSA at each joint; \rev{(c) a 4DoF robot with SEAs at each joint; (d) a 4DoF robot with VSAs at each joint;} (e) a 7DoF system with SEA at each joint; (f) a 7DoF system with VSA at each joint; (g) an \rev{underactuated compliant} robot modeled as a 2DoF robot where the first joint is actuated by a SEA and the second elastic joint is unactuated; and (h) an \change{underactuated compliant} robot modeled as a 2DoF robot where the first joint is actuated by a VSA and the second elastic joint is unactuated \rev{(i) an underactuated serial manipulator with 21 elastic joints and only \rev{the} first one is actuated by SEA.
We perform simulations and experiments for (a), (b), (c), (d), (g), (h) and only simulations for (e), (f), (i). We   
 now introduce the various systems }

The physical parameters for 2DoF, \rev{4DoF} robots with  SEA and VSA are: the mass of the links $m_i=0.55$ \unit{\kilo\gram}, the inertia of the motors $\mathbf{B}_{i,i} = 10^{-3}$ \unit{\kilo\gram\meter^2}, the center of mass distance $a_i = 0.085$ \unit{\meter}, and the link lengths $l_i=0.089$ \unit{\meter}. 
Similarly, we consider the following physical parameters for the 7DoF system actuated by SEA and VSA at each joint. 
\begin{table}
\caption{Physical parameters of 7DoF arm}
\begin{center}
\begin{tabular}{ c|c|c|| c|c|c } \label{tab: phys_params} 
& $m_{i}$[kg] & $l_i$[m]&  & $m_{i}$[kg] & $l_i$[m]\\
\hline
Link 1 & 1.42 & 0.1 &  
Link 2 & 1.67 & 0.5 \\ 
Link 3 & 1.47 & 0.06 & 
Link 4 & 1.10 & 0.06 \\ 
Link 5 & 1.77 & 0.06 & 
Link 6 & 0.3 & 0.06 \\
Link 7 & 0.3 & 0.06 \\
\hline
\end{tabular}
\end{center}
\end{table}
The physical parameters resemble those of a Talos arm \cite{talos} (Table \ref{tab: phys_params}), but we add a compliant actuation at each joint. The motor inertia is set to $\mathbf{B}_{i,i} = 10^{-3}$ \unit{\kilo\gram\meter^2} for all joints. 

We classify our results as \rev{follows:}

\begin{enumerate}
    \item In Section \ref{sec:subsec1}, we report the difference between the Jacobians of the dynamics computed by numerical differentiation and analytical differentiation. The simulation is conducted for the 2DoF and 7DoF cases actuated by both SEA and VSA. We show the average and standard deviation of the computed difference at \rev{ 20 random configurations and with velocities  and control inputs set as zero. The number of iterations for convergence and \new{the }time taken for convergence are also reported. }
 
    \item In Section \ref{sec:subsec2}, we present the simulation and \change{experimental} results of the end-effector regulation task with SEAs and VSAs for a 2DoF, \change{4DoF} and 7DoF arms.
    \begin{itemize}
        \item \textit{2DoF robot}: the Cartesian target point is $[0.01,~0.2]$ \unit{\meter}, the time horizon is $T = 3$ \unit{\second}, and the weights corresponding to control regularization is $10^{-2}$, state regularization is $1$, goal-tracking cost is $10^{-1}$, and terminal cost contains a goal-tracking cost with $10^{4}$ weighting factor. In the case of SEAs, we define the stiffness of each joint as $\sigma_i = 3$ \unit{\newton\meter/\radian}. Instead, in the case of VSAs, the value of the stiffness lie between $\sigma_{\textnormal{min}} =0.05$ \unit{\newton\meter/\radian} and $\sigma_{\textnormal{max}}=15$ \unit{\newton\meter/\radian}. We also include an additional cost term \eqref{eq:stiffness_cost} which accounts for the power consumption due to change in the stiffness and the weight is set to $1$ and the $\lambda$ is set to $10$.
        \item \rev{4DoF robot: the Cartesian target point is $[.1, .3,.15]$ \unit{\meter}, the time horizon is $T = 4$ \unit{\second}, and the weights corresponding to control regularization is $10^{-1}$, state regularization is $10^{-3}$, goal-tracking cost is $10^{0}$, and terminal cost contains a goal-tracking cost with $10^{4}$ weighting factor. In the case of SEAs, we define the stiffness of the first joint as $\sigma_i = 10$ \unit{\newton\meter/\radian} each joint as $\sigma_i = 5$ \unit{\newton\meter/\radian}. Instead, in the case of VSAs, the Cartesian target point is $[.15, .3,.15]$ m, the value of the stiffness lies between $\sigma_{\textnormal{min}} =2$ \unit{\newton\meter/\radian} and $\sigma_{\textnormal{max}}=15$ \unit{\newton\meter/\radian}. We also include an additional cost term \eqref{eq:stiffness_cost} which accounts for the power consumption due to change in the stiffness and the weight is set to $1$ and the $\lambda$ is set to $10$.}
        		
        \item \textit{7DoF robot}: the Cartesian target point is set as $[0, 0, 0.4]$ \unit{\meter}, and the time horizon is $T = 1.5$ \unit{\second}. The weights corresponding to control regularization is $10^{-2}$, state regularization is $1$ and goal-tracking cost is $10^{-1}$. In the case of SEAs, we define the stiffness value of each joint is set to $\sigma = 10$ \unit{\newton\meter/\radian}. Instead, in the case of VSAs, the stiffness value of each joint varies between $\sigma_\textnormal{min} = 0.7$ \unit{\newton\meter/\radian} and $\sigma_\textnormal{max} = 10 $ \unit{\newton\meter/\radian}.
    \end{itemize}
    \item In Section \ref{sec:subsec3}, we illustrate the results obtained for \change{underactuated compliant robots} with varying DoFs and varying degrees of underactuation. 
    \begin{itemize}
    	\item \rev{\textit{Underactuated 2DoF compliant arm:}  In the case where the first joint is actuated by a SEA, the stiffness constant $\sigma =5$ \unit{\newton\meter/\radian} and the time horizon is $T = 3$ \unit{\second}. For the case where the first joint is actuated by VSA, the value of $\sigma_{\textnormal{min}} =0.05$ \unit{\newton\meter/\radian} and $\sigma_{\textnormal{max}} = 15$ \unit{\newton\meter/\radian} and the time horizon is $T = 2$ \unit{\second}. The weights corresponding to control regularization is $10^{-1}$, state regularization is $10^{-2}$, $\ell_{\textnormal{vsa}}$ is $10^{-2}$ and the goal-tracking cost is $10^{-1}$. The stiffness of the second link is 2 \unit{\newton\meter/\radian}.}
    	
		\item	\rev{\textit{Underactuated serial manipulator}: The stiffness matrix of a flexible link can be written as $\mathbf{K} = \textnormal{diag}(\mathbf{K}_{11}, ... , \mathbf{K}_{mm})$. Further, the motor inertias are considered to be negligible. We model an underactuated serial manipulator as a 21 joint under-actuated compliant arm with the total length being 3.15 \unit{\meter} and only the first joint is actuated and has 20 passive elastic joints. }
    \end{itemize}

    \item In the final Section \ref{sec:subsec4}, we compare the energy consumption in an end-effector regulation task, for rigid, SEA, and VSA cases. We report the results with three different systems: \rev{a} fully actuated 2DoF system, \rev{a} fully actuated 7DoF system and, \rev{a} 2DoF \change{underactuated compliant arm}. For comparison \rev{purposes}, the weights of the various cost terms in the objective function for the task for rigid, SEA, and VSA are the same.
    
\end{enumerate}

For cases involving SEAs, we use \rev{the} FDDP solver, and for cases with VSAs, we impose these box constraints on the stiffness values using the Box-FDDP solver. Both solvers used in the simulations and experiments are available in \rev{the} \textsc{Crocoddyl} library \cite{crocoddyl}.

\subsection{Experimental setup}
We use the experimental setup illustrated in Fig. \ref{fig:cover} for all experiments. At each joint, we utilize \textsc{qbMove advanced} \cite{qbmove} as the elastic actuator. It features two motors attached to the output shaft and is based on an \rev{antagonistic} mechanism. The motors and the links of \new{the }actuator both include AS5045 12 bit magnetic encoders. The actuator's elastic torque $\tau$ and nonlinear stiffness function $\sigma$ satisfy the following equation
\begin{align}
    \tau &= 2\beta \cosh{\alpha \theta_{\textnormal{s}}} \sinh{\alpha(q-\theta_{\textnormal{e}})} \;,\\
    \sigma &= 2 \alpha \beta \cosh(\alpha \theta_{\textnormal{s}}) \cosh{\alpha(q-\theta_{\textnormal{e}})} \;,
    \label{eq:actuator_dyn}
\end{align}
where, $\alpha=6.7328$ \unit{\radian^{-2}}, $\beta=0.0222$ \unit{\newton\meter}, $\theta_{\textnormal{e}}$ is the motor equilibrium position, $\theta_{\textnormal{s}}$ tunes the desired motor stiffness and $q$ is the link-side position.

For the experiments related to systems with an un-actuated joint, we set the $\theta_{\textnormal{s}}$ of the passive joint as constant and $\theta_{\textnormal{e}}$ is set to be null. This pragmatic choice equips the  passive joint with a torsional spring and position encoder. This enables us to resemble an \change{underactuated compliant} robot. 

The optimal control with VSAs \rev{returns}   $\boldsymbol\tau$ and $\boldsymbol\sigma$ as the output of the optimal control problem.
We need to invert the equations \eqref{eq:actuator_dyn} to find $\boldsymbol\theta_{\textnormal{e}}$ and $\boldsymbol\theta_{\textnormal{s}}$ as these will be input to the actual motors. The parameters like $\alpha$ and $\beta$ can be found \new{in} the manufacturer's datasheet, and $\mathbf{q}$ is the link trajectory seen from the optimization routine. 
In the results obtained from the experiments, we compare the results of feedback control with pure feed-forward control. To  quantify  the  tracking  performance  of  these controllers,  we  use  as  metric  the  root  mean  square  (RMS) error.

\section{Results and discussions}\label{sec:results}
In this section, we present and discuss the simulation and experimental results. The code necessary to reproduce the results reported in this section are publicly available. %
\footnote{\href{https://github.com/spykspeigel/aslr_to}{github.com/spykspeigel/aslr\_to }}

\begin{table}[!tbh]
	\caption{\rev{Ratio between Difference of NumDiff and Analytical Diff and Max Analytical Diff}}\vspace{-1em}
	\label{tab:comparison_soft}
	\begin{center}
		\begin{tabular}{@{} l rcc r rcc @{}}
			\toprule
			&\multicolumn{2}{c}{SEA} & \multicolumn{2}{c}{VSA} \\
			\cmidrule{2-3}\cmidrule{4-5}
			\emph{Problems}              &Average &Std. Dev. &Average &Std. Dev.\\
			\midrule
			\texttt{2DoF}            &$2\cdot 10^{-7}$    &$1 \cdot 10^{-7}$ &$8\cdot 10^{-6}$    &$3\cdot 10^{-7}$  \\
			\texttt{7DoF}        &$5 \cdot 10^{-7}$ &$5 \cdot 10^{-6}$  &$9\cdot 10^{-6}$ &$6\cdot 10^{-6}$ \\
			\bottomrule
		\end{tabular}
	\end{center}
\end{table}
\vspace{-3mm}

\subsection{Analytical derivatives vs. numerical derivatives}\label{sec:subsec1}
\rev{Using analytical derivatives of the dynamics} is expected to improve the numerical accuracy. We present simulation results in this subsection to support this claim. In Table \ref{tab:comparison_soft}, we show the average and standard deviation of \rev{the} difference between numerical differentiation and analytical differentiation.  \rev{This is for} at 20 random configurations and with initial velocities  and controls set to zero. The values reported are the ratio with respect to the maximum value of the analytical derivative obtained among the randomized configurations for each case.

\rev{
In Table \ref{tab:iter_comparison_soft}, we show the average and standard deviation of the number of iterations for convergence between numerical differentiation and analytical differentiation-based optimal control method. The 2DoF and 7DoF systems are assigned end-effector regulation tasks with 20 random desired end-effector positions which include non-zero terminal velocity.
For the underactuated 2DoF system, the task is to swingup to the vertical position with 20 random initial configurations which include nonzero initial velocity. The standard deviation for numerical derivatives is significantly more than that of analytical derivative-based OC for underactuated 2DoF VSA cases, even though the average value is similar to analytical derivatives-based OC.

For the 7DoF system with each joint actuated by SEA case, the optimal control based on numerical derivatives converges more than 400 iterations in 8 out of 20 cases. Similarly, in the 7DoF system with each joint actuated by VSA, 12 out of 20 problems take more than 400 iterations to converge.
\new{One can obtain similar results for cases with higher degrees of freedom and generic under-actuation.}}
\begin{table}[!tbh]
	\caption{\rev{Iterations taken by Optimal Control based on NumDiff and Analytical Diff }}\vspace{-1em}
	\label{tab:iter_comparison_soft}
	\begin{center}
		\begin{tabular}{@{} l rcc r rcc @{}}
			\toprule
			&\multicolumn{2}{c}{AnalyticalDiff} & \multicolumn{2}{c}{NumDiff} \\
			\cmidrule{2-3}\cmidrule{4-5}
			\emph{Problems}              &Average &Std. &Average &Std.\\
			\midrule
			\texttt{2DoF SEA}            &$6$    &$2$ &$7.1$   &$1.6$  \\
			\texttt{2DoF VSA}        &$24$ &$2.1$  &$24.3$ &$2.3$ \\
			\texttt{under 2DoF SEA}   &$31.1$ &$2.1$  &$32.6$ &$3.3$ \\
			\texttt{under 2DoF VSA}        &$39.1$ &$6.7$  &$40.9$ &$38$ 
			\\
			\texttt{7DoF SEA}        &$45.6$ &$19.68$ & &8 cases take $>400$\\
			\texttt{7DoF VSA}        &$47.83$ &$15.3$  & &12 cases take $>400$  \\
			\bottomrule
		\end{tabular}
	\end{center}
\end{table}
\vspace{-3mm}

\rev{Most importantly, the computation of analytical derivatives is computationally cheaper. In Table \ref{tab:time_comparison_soft}, we provide a comparison of the time taken per iteration for the optimal problem with an end-effector regulation task for 2DoF and 7Dof case and swingup-task for the under-actuated 2DoF system. As can be noted, the analytical derivatives provide about 100 times increased performance. A C++ implementation is expected to show increased performance and enable MPC-based implementation as can be noted in our recent works \cite{mastalli22mpc, invdyn}.}
\begin{table}[!tbh]
	\caption{\rev{Average time taken per iterations for Optimal control based on NumDiff and Analytical Diff (in seconds)}}\vspace{-1em}
	\label{tab:time_comparison_soft}
	\begin{center}
		\begin{tabular}{@{} l rcc r rcc @{}}
			\toprule
			&\multicolumn{2}{c}{AnalyticalDiff} & \multicolumn{2}{c}{NumDiff} \\
			\cmidrule{2-3}\cmidrule{4-5}
			\emph{Problems}         &SEA &VSA     &SEA  &VSA \\
			 \midrule
			\texttt{2DoF}        &$0.098$ &$0.099$  &$3.9$ &$4.3$ \\
			\texttt{under 2DoF}  &$0.13$ &$0.14$  &$7.1$ &$7.9$ \\
			\texttt{7DoF}           &$0.17$ &$0.32$ &$23.01$    &$23.22$   \\
			\bottomrule
		\end{tabular}
	\end{center}
\end{table}
\vspace{-3mm}

\subsection{Optimal trajectory for regulation tasks of SEA and  VSA systems} \label{sec:subsec2}
\begin{figure}
	\centering
	\includegraphics[width=.32\columnwidth]{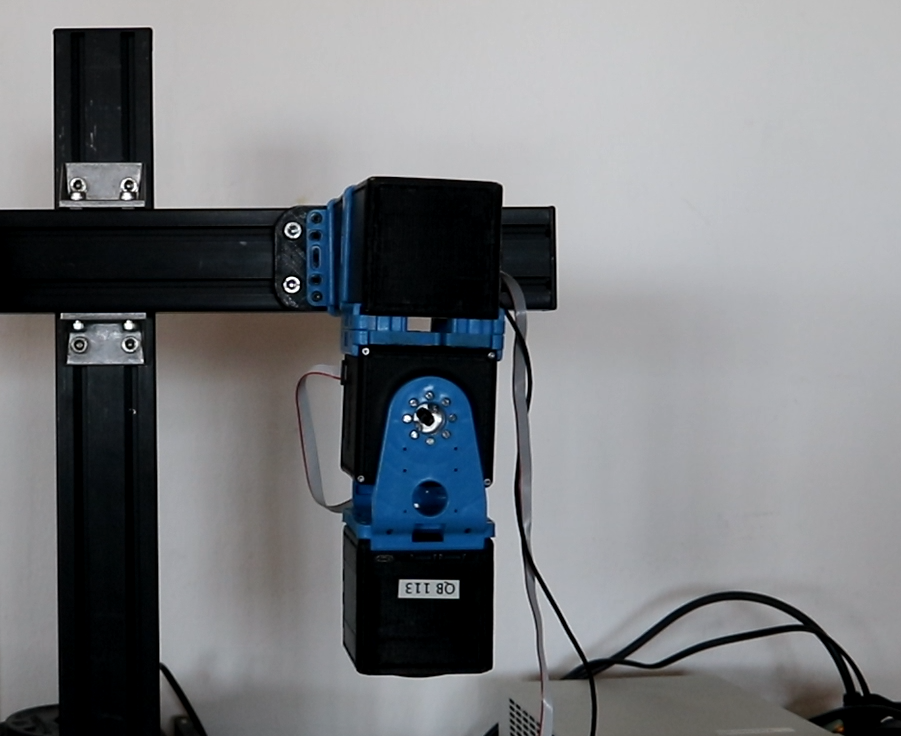}
	\includegraphics[width=.32\columnwidth]{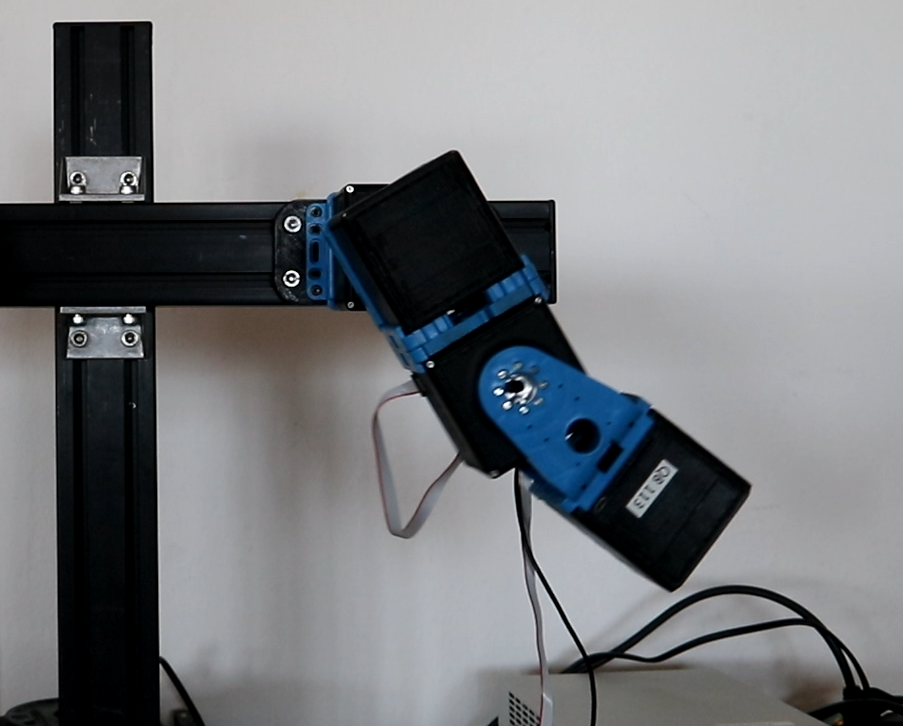}
	\includegraphics[width=.32\columnwidth]{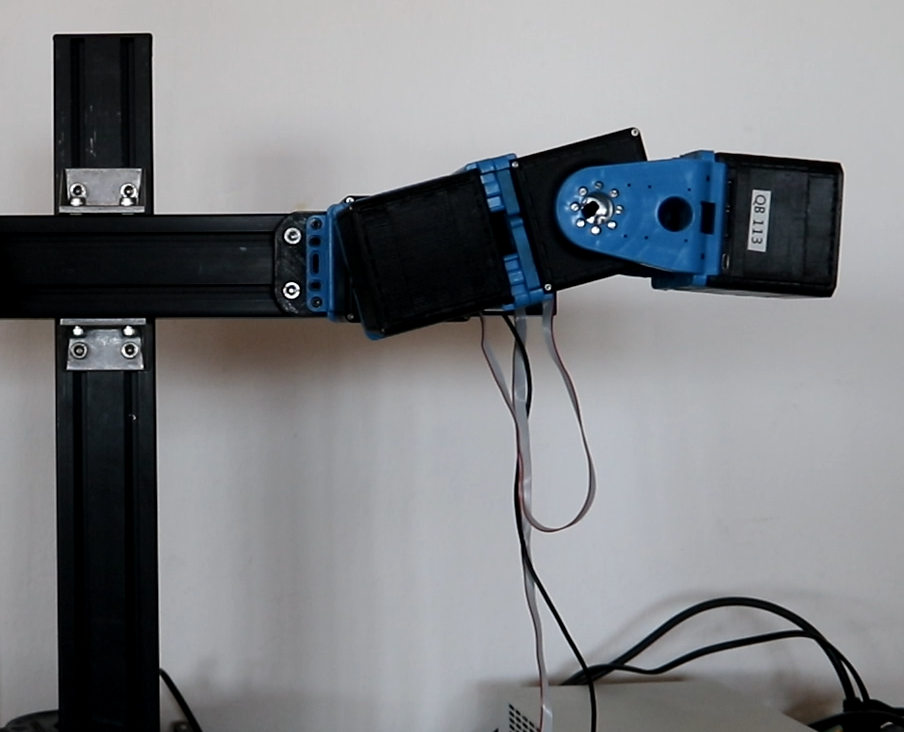}
	\caption{Snapshots of the end-effector regulation task with the 2DoF system actuated by SEA in both joints. Please refer to the video attachment for more details.\label{fig:2dof_sea_ps}}
\end{figure}

\begin{figure}
\centering
  \subfigure[Link positions]{\includegraphics[width=0.95\columnwidth]{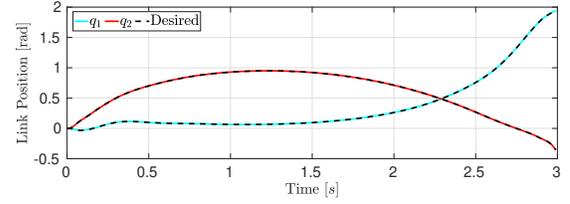} \label{fig:2dof_sea1}}\vspace{-0.3cm}
  \subfigure[Input torques]{\includegraphics[width=0.95\columnwidth]{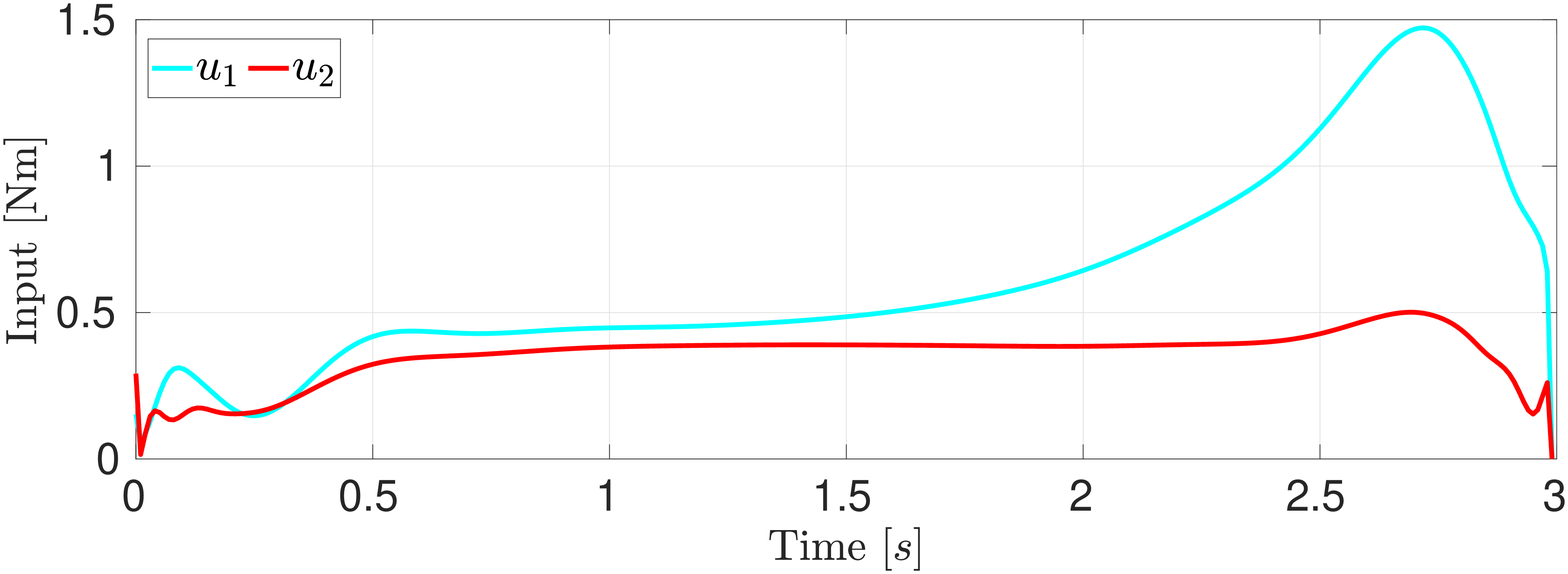}  \label{fig:2dof_sea2}}\vspace{-0.3cm}\\
  \subfigure[Link 1 position]{\includegraphics[width=0.95\columnwidth]{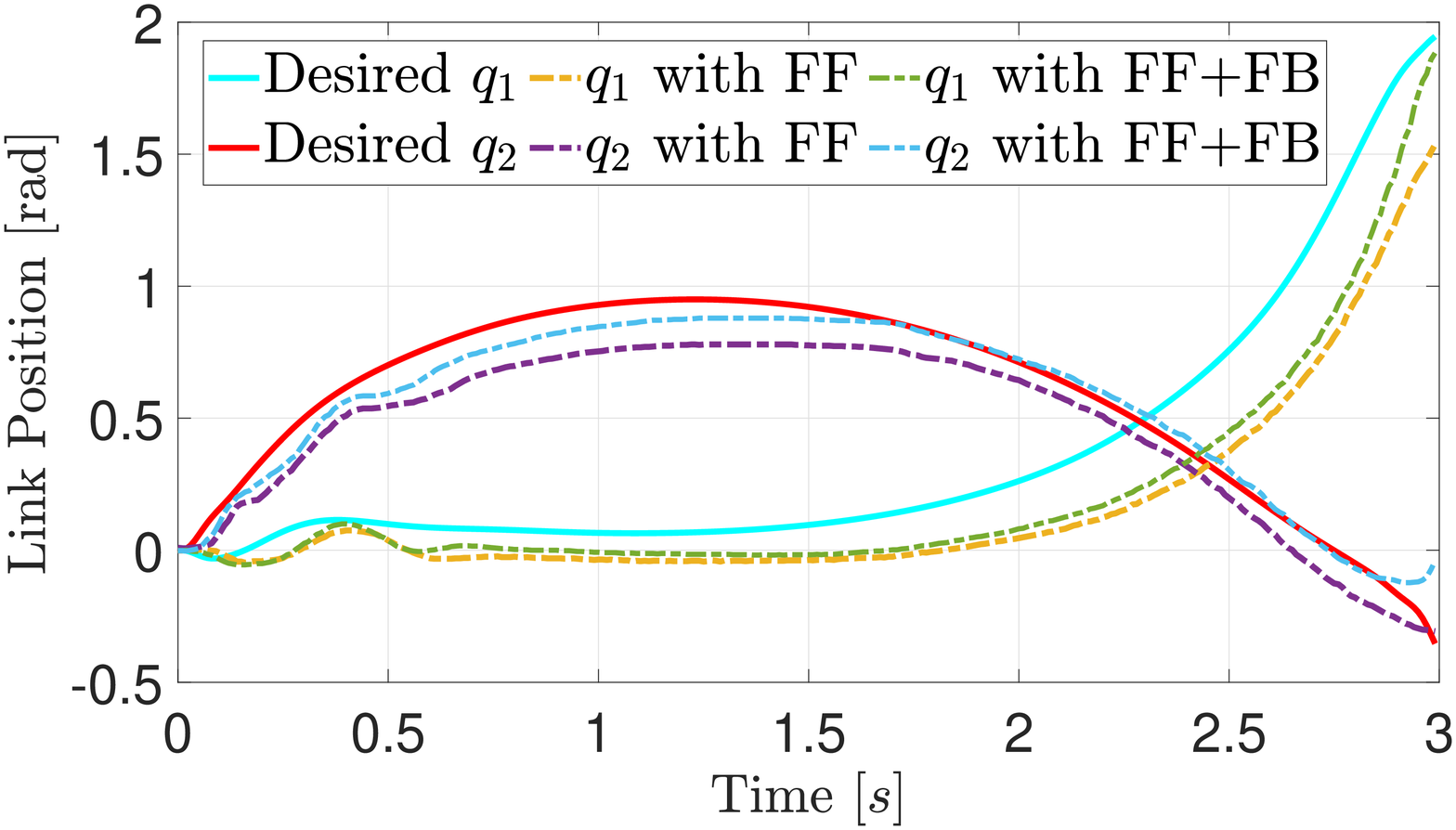}  \label{fig:2dof_sea3}}\vspace{-0.3cm}
  \caption{\new{End-effector regulation task of a 2DoF system with SEA in both joints. (a) Joint evolution in simulation. (b) Input torque evolution in simulation. 
  (c) Evolution of joint 1 and joint 2 in experiments. We compare the desired and the link positions using pure feed-foward (FF) and the feedback  and feed-forward (FF+FB) cases, \new{which shows better performance.} \label{fig:2dof_sea}}}
\end{figure}

Fig. \ref{fig:2dof_sea} shows the simulation and experimental results of the 2DoF SEA case. This includes the optimal trajectory (Fig. \ref{fig:2dof_sea1}) and the input sequence (Fig. \ref{fig:2dof_sea2}). At the end of the task, \rev{the} end -effector position was $[0.0098,~ 0.20]$ \unit{\meter}. \new{The link position in the experiment is presented in Fig. \ref{fig:2dof_sea3}}. A photo-sequence of the experiment is depicted in Fig. \ref{fig:2dof_sea_ps}; please \new{also refer} to the video attachment. The RMS error for the first joint in \rev{the} case of pure feed-forward control was $0.2503$ \unit{\radian} and in \rev{the} case of feedback control was $0.2296$ \unit{\radian}. Similarly, for the second joint, we observe that \rev{the} RMS error \rev{in the} pure feed-forward case was $0.1274$ \unit{\radian}, and with feedback control was $0.1076$ \unit{\radian}. 
So, this illustrates the advantage of using feedback gain along with the feed-forward action for the task.

\begin{figure}
\subfigure[Link Position]{\includegraphics[width=0.95\columnwidth]{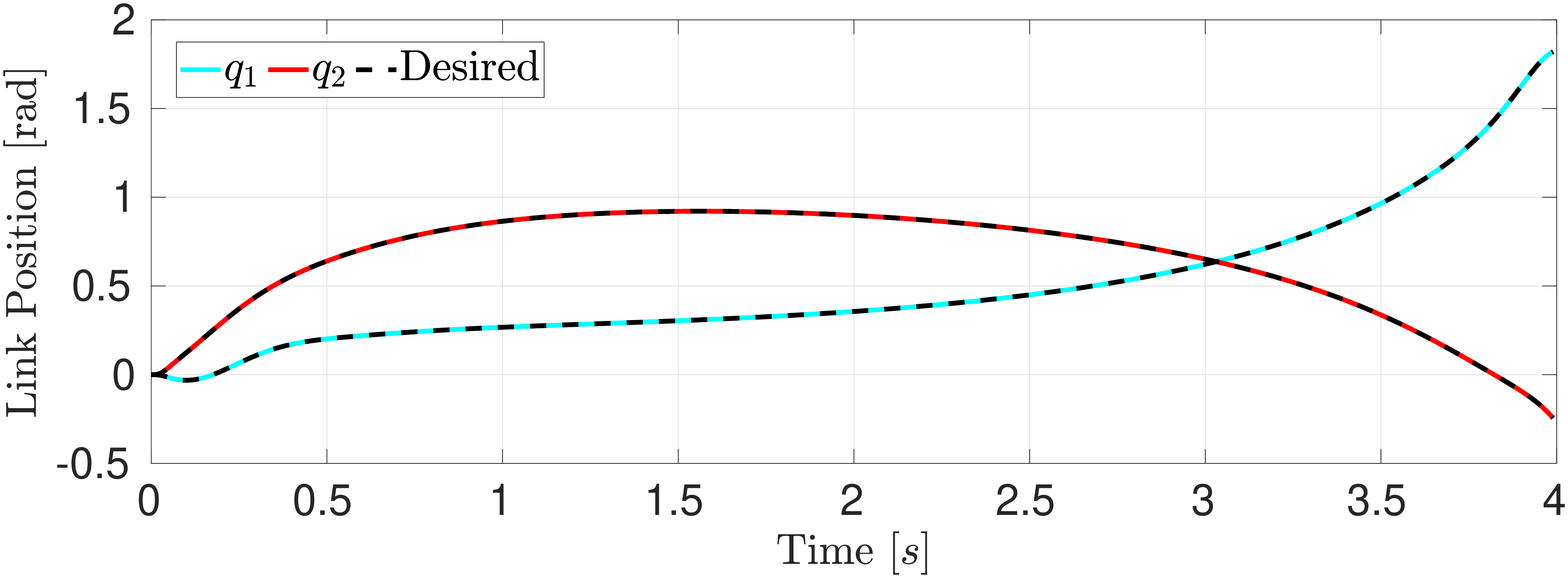}\label{fig:2dof_vsa1}}\vspace{-0.3cm}
\subfigure[Input Torques]{\includegraphics[width=0.95\columnwidth]{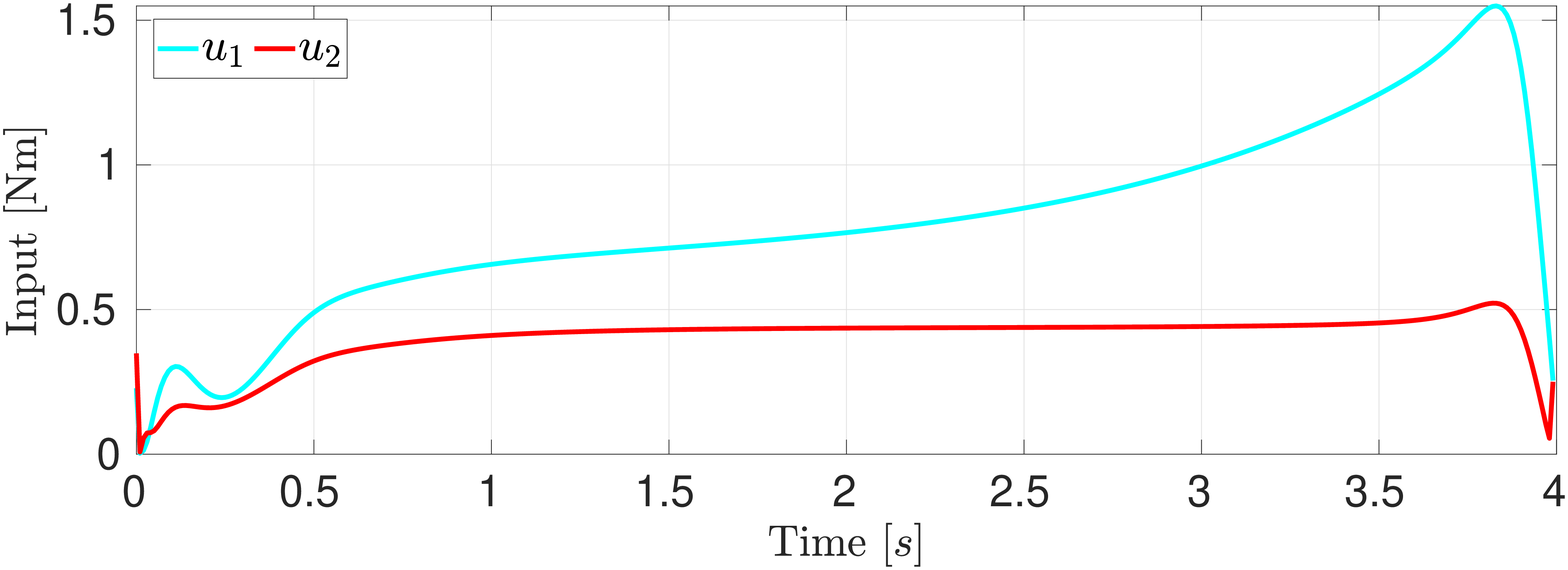}\label{fig:2dof_vsa1}}\vspace{-0.3cm}
\subfigure[Stiffness Profile]{\includegraphics[width=0.95\columnwidth]{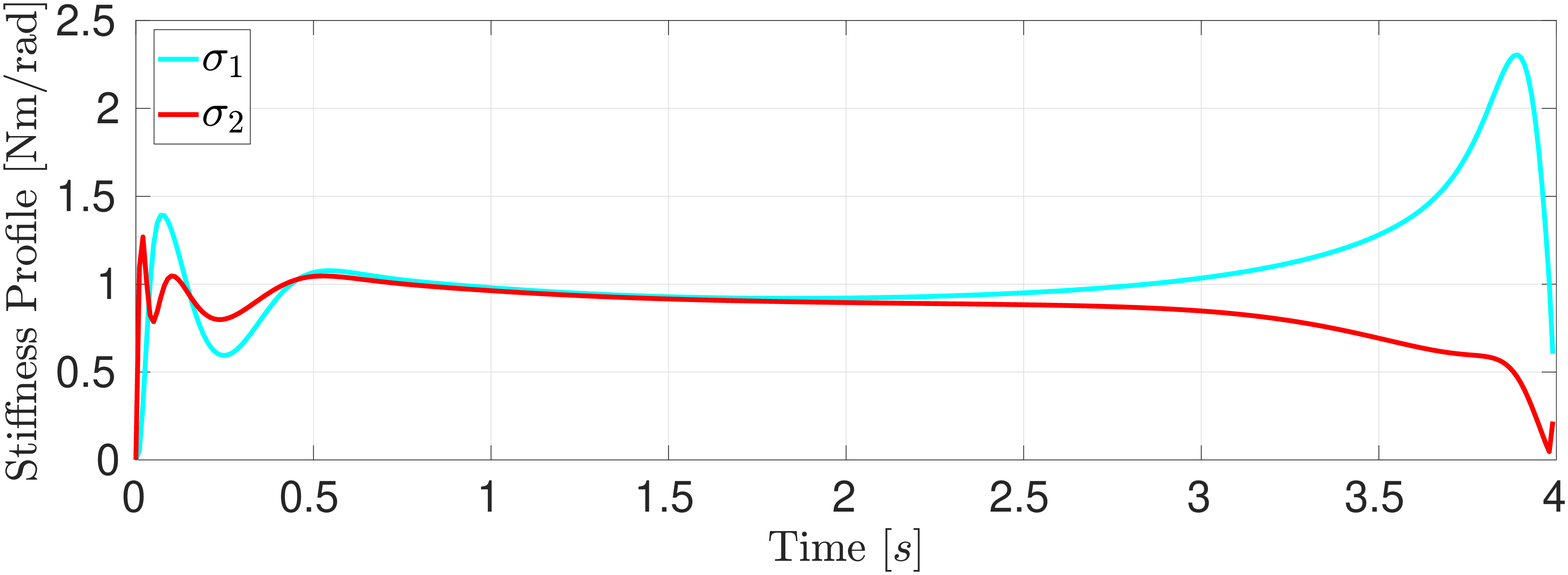}\label{fig:2dof_vsa2}}\vspace{-0.3cm}
\subfigure[Links position ]{\includegraphics[width=0.95\columnwidth]{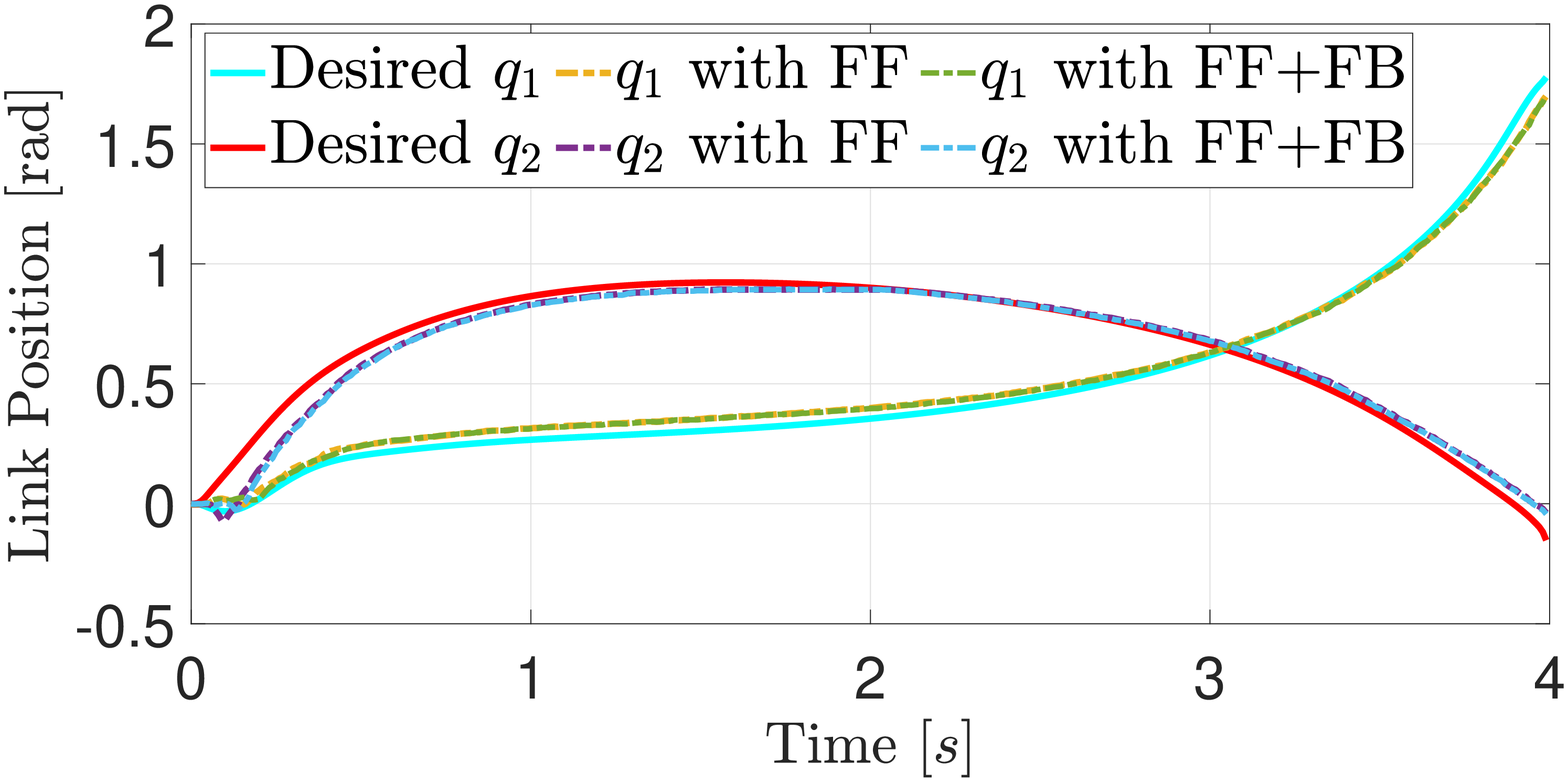}\label{fig:2dof_vsa3}}\vspace{-0.4cm}
  \caption{\new{End-effector regulation task of a 2DoF system with VSA in both joints. (a) Link position in simulations (b) Input torque evolution in simulation. (c) Input stiffness evolution in simulation.
  (d) Evolution of joint 1 and joint 2 in experiments. We compare the desired and the link positions using pure feed-foward (FF). }
  \label{fig:2dof_vsa}
 }
\end{figure}

Similarly, Fig. \ref{fig:2dof_vsa} illustrates the results of the same end-effector regulation task for the 2DoF system actuated by VSA as described in section \ref{sec:simulation}.  The \new{end-effector} position at the end of \new{the} task was  $[0.011,~ 0.202]$ \unit{\meter}. Fig. \ref{fig:2dof_vsa3} shows plot of the link positions obtained from the experiments.

\begin{figure}
\centering
\subfigure[\rev{Snapshots of the end-effector regulation for 4DoF with a SEA in each joint. Please refer to the video attachment for more details.}]{
\includegraphics[width=.31\columnwidth]{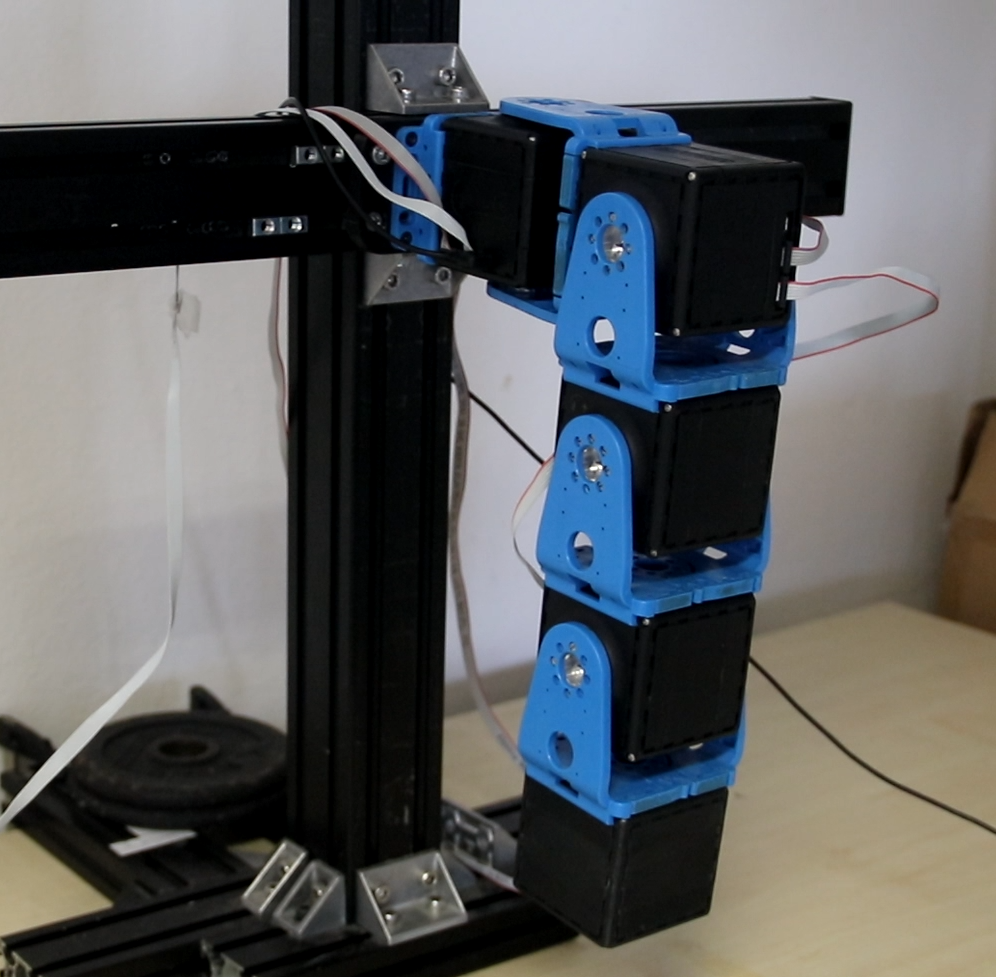}
\includegraphics[width=.3\columnwidth]{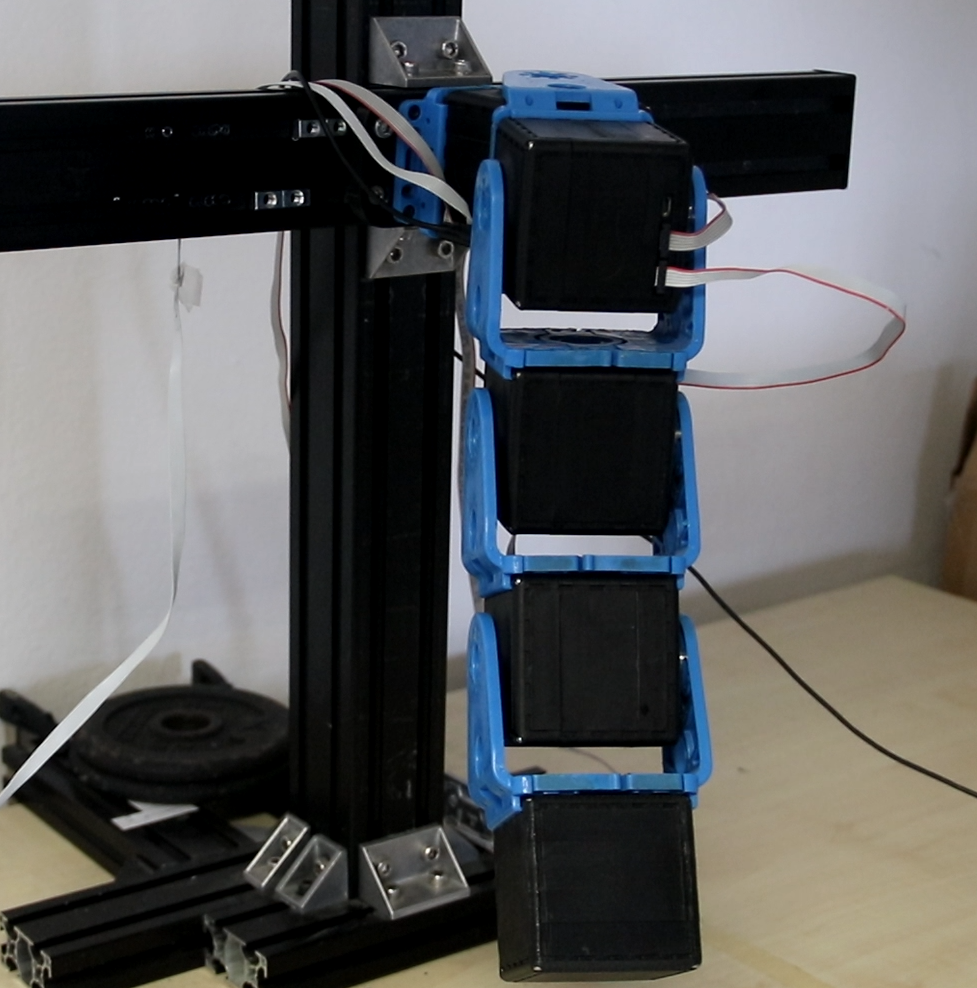}
\includegraphics[width=.305\columnwidth]{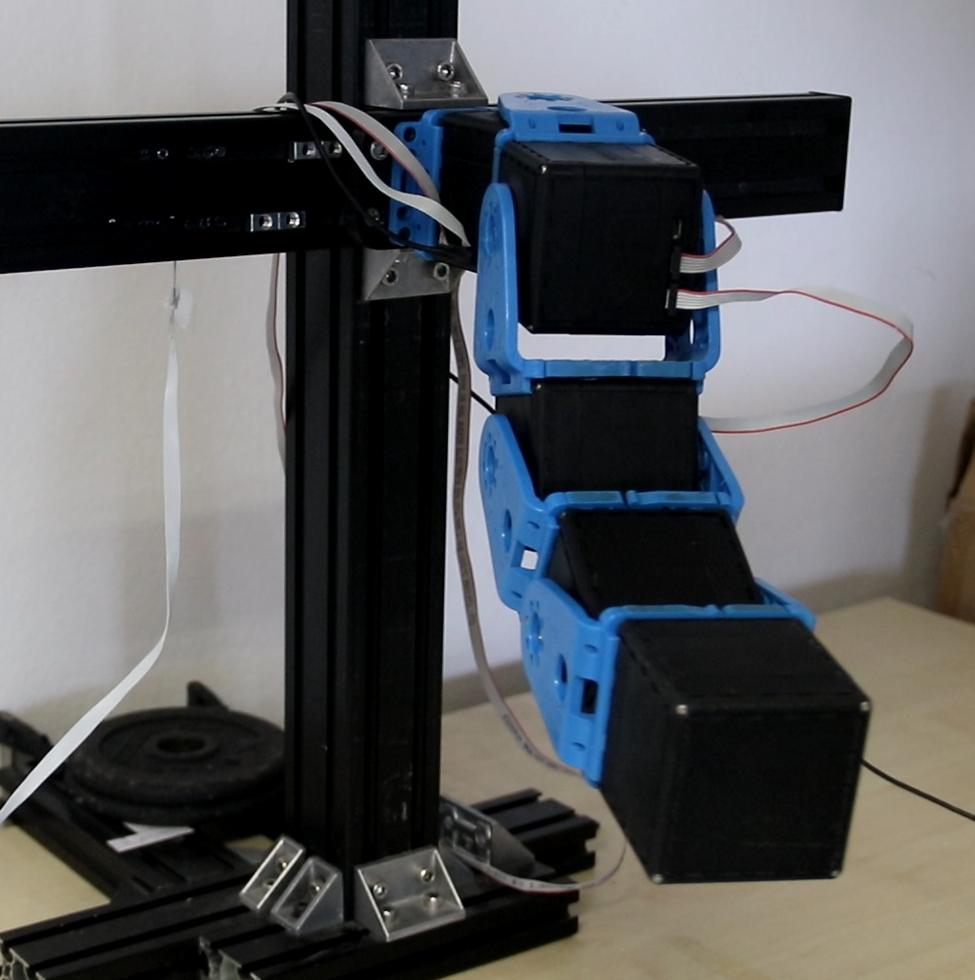}\label{fig:4dof_exp}}
\subfigure[Link position]
{\includegraphics[width=0.95\columnwidth]{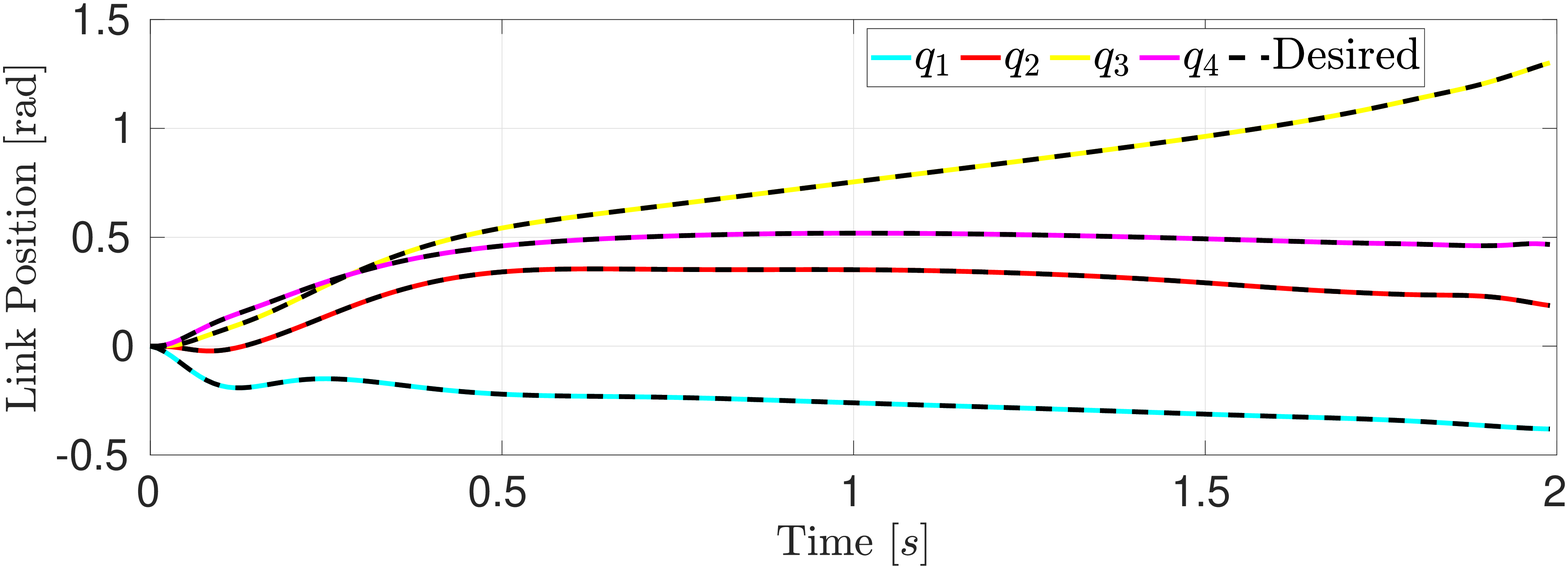}}\vspace{-0.3cm}
\subfigure[Input Controls]
{\includegraphics[width=0.95\columnwidth]{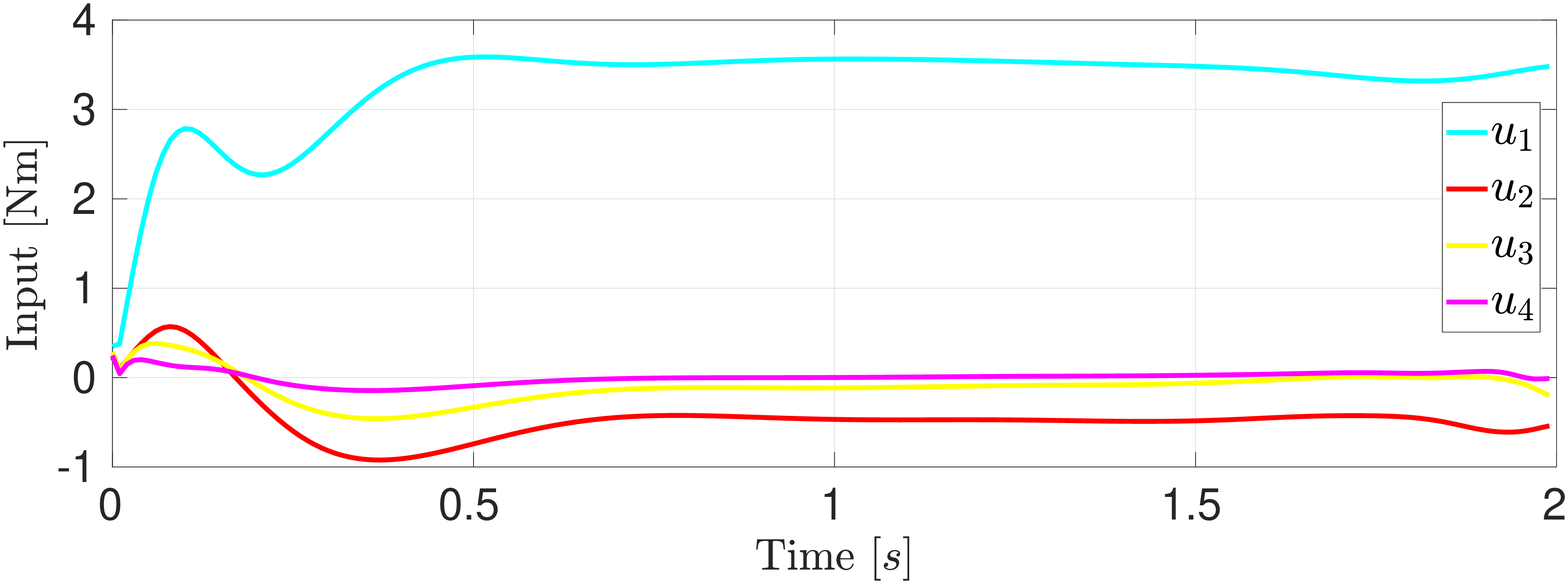}}\vspace{-0.3cm}
\subfigure[Link 1 position]
{\includegraphics[width=0.95\columnwidth]{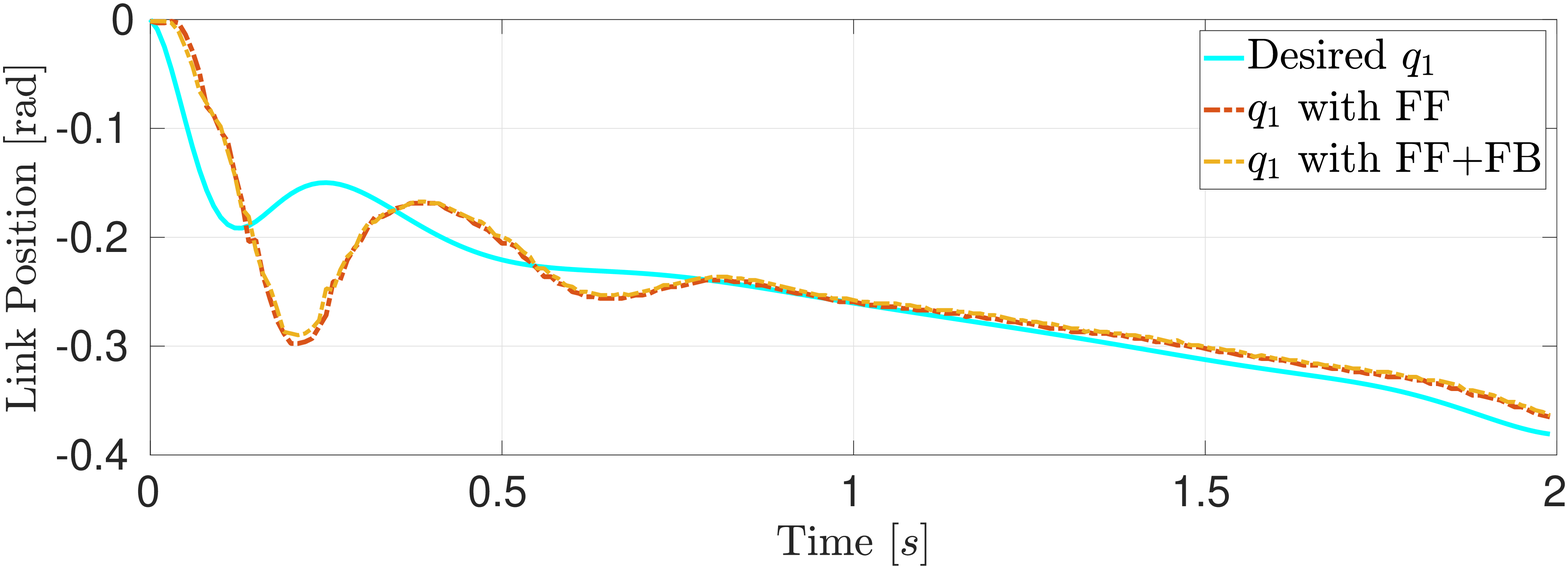}\label{fig:4dof_sea1}}\vspace{-0.3cm}
\subfigure[Link 2 position]
{\includegraphics[width=0.95\columnwidth]{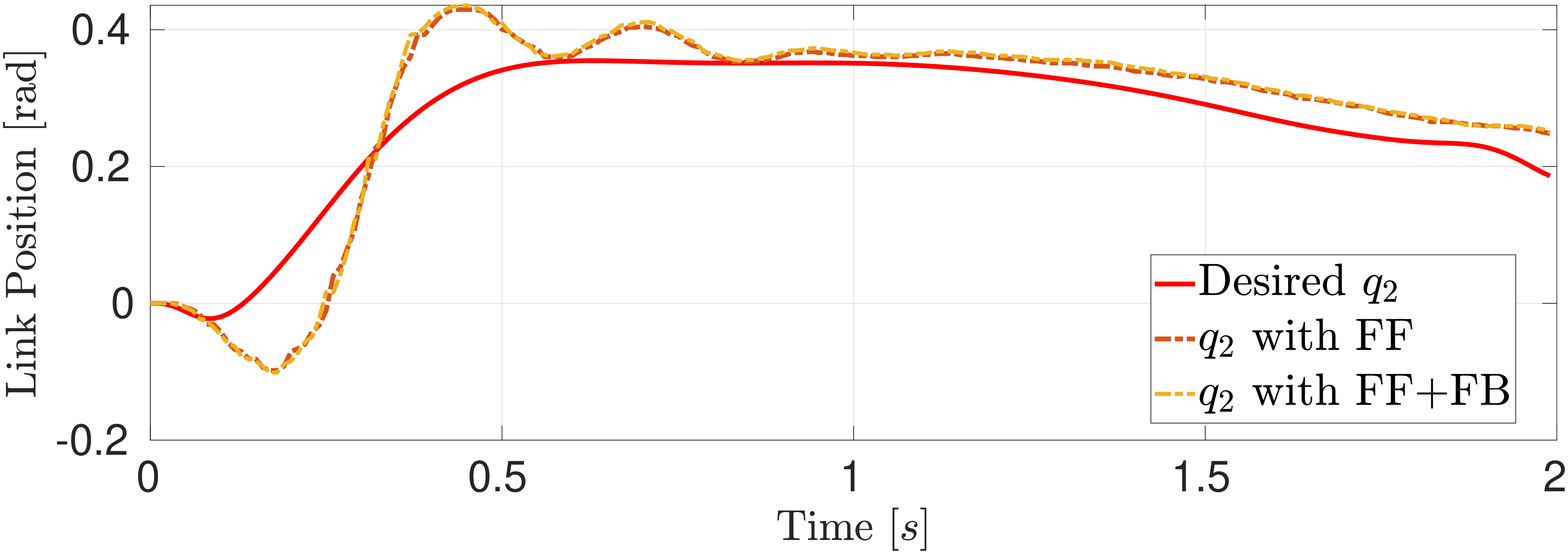}\label{fig:4dof_sea2}}\vspace{-0.3cm}
\subfigure[Link 3 position]
{\includegraphics[width=0.95\columnwidth]{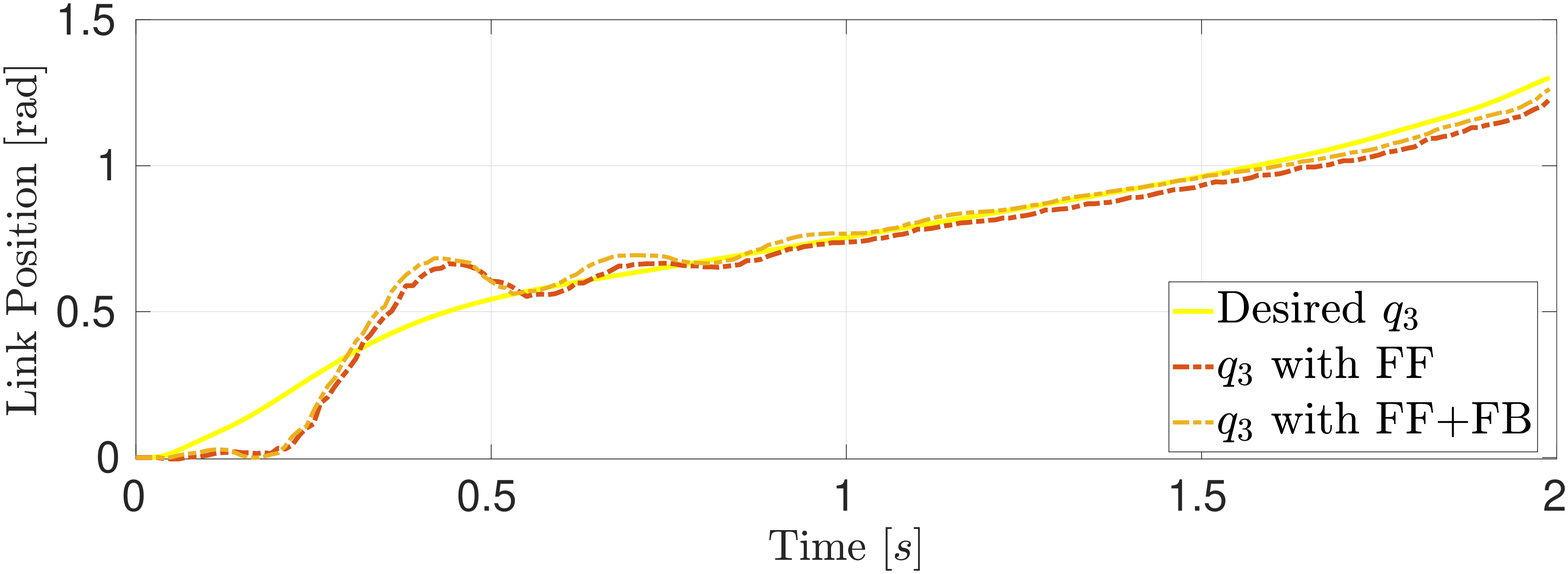}\label{fig:4dof_sea3}}\vspace{-0.3cm}
\subfigure[Link 4 position]
{\includegraphics[width=0.95\columnwidth]{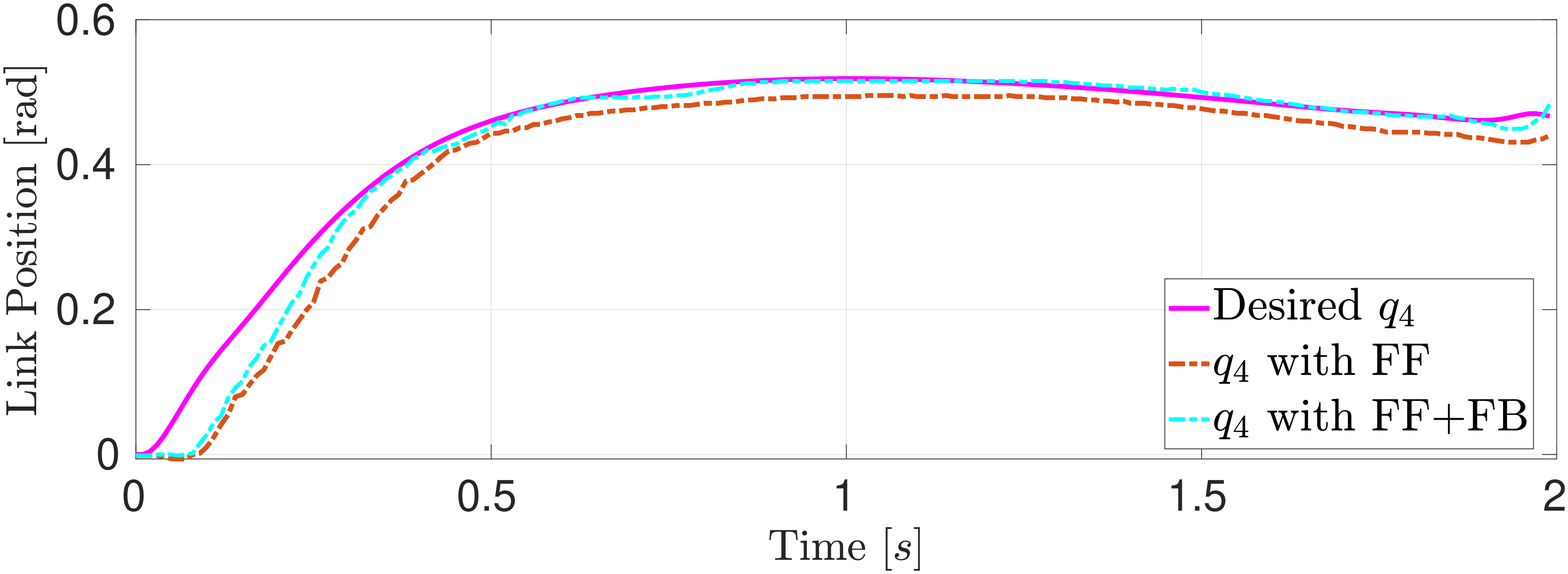}\label{fig:4dof_sea4}}\vspace{-0.3cm}
\caption{\new{End-effector regulation task of a 4DoF system with SEA in all joints. (a) Link position in simulation (b) Input Torque.   The desired end-effector position was [0.1, 0.3, 0.15] m and it reached [0.11, 0.33, 0.13] m.
(c), (d), (e), (f) Evolution of all joints in experiments. We compare the desired (simulation) and the link positions using pure feed-forward (FF) and feed-forward with feedback (FF+FB), which shows better performance}. 
	\label{fig:4dof_sea}}
\end{figure}

\begin{figure}
\subfigure[Link position]
{\includegraphics[width=0.95\columnwidth]{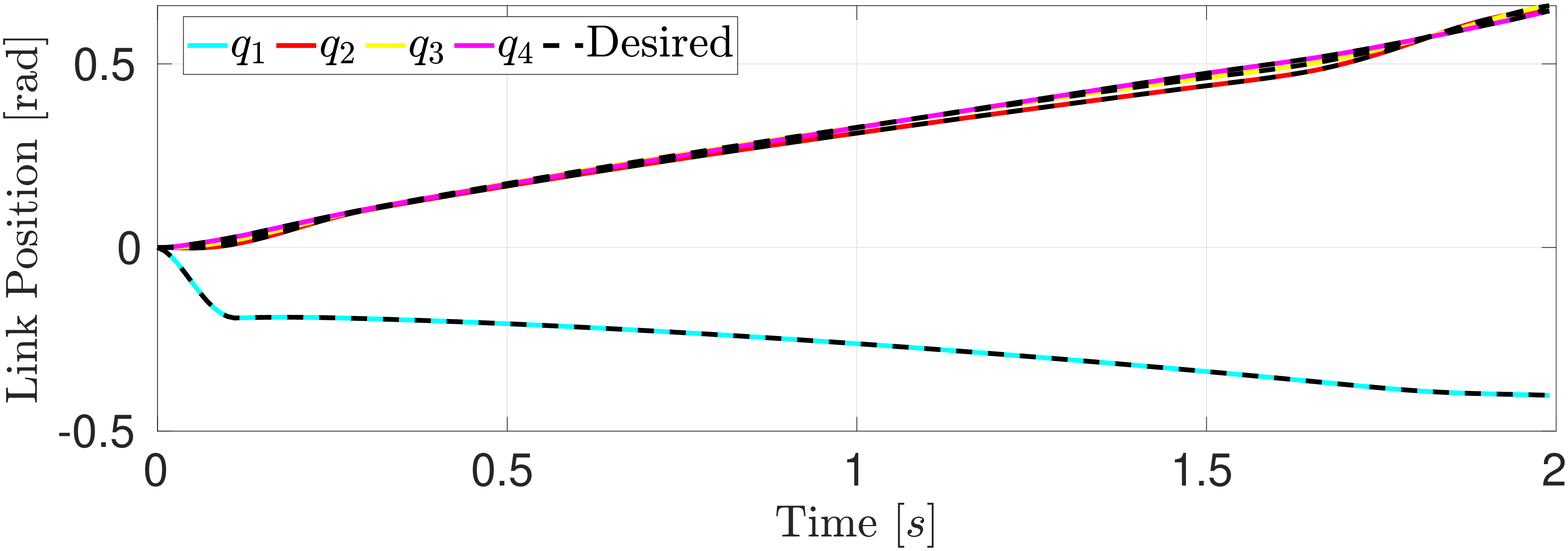}}\vspace{-0.3cm}
\subfigure[Input Control]
{\includegraphics[width=0.95\columnwidth]{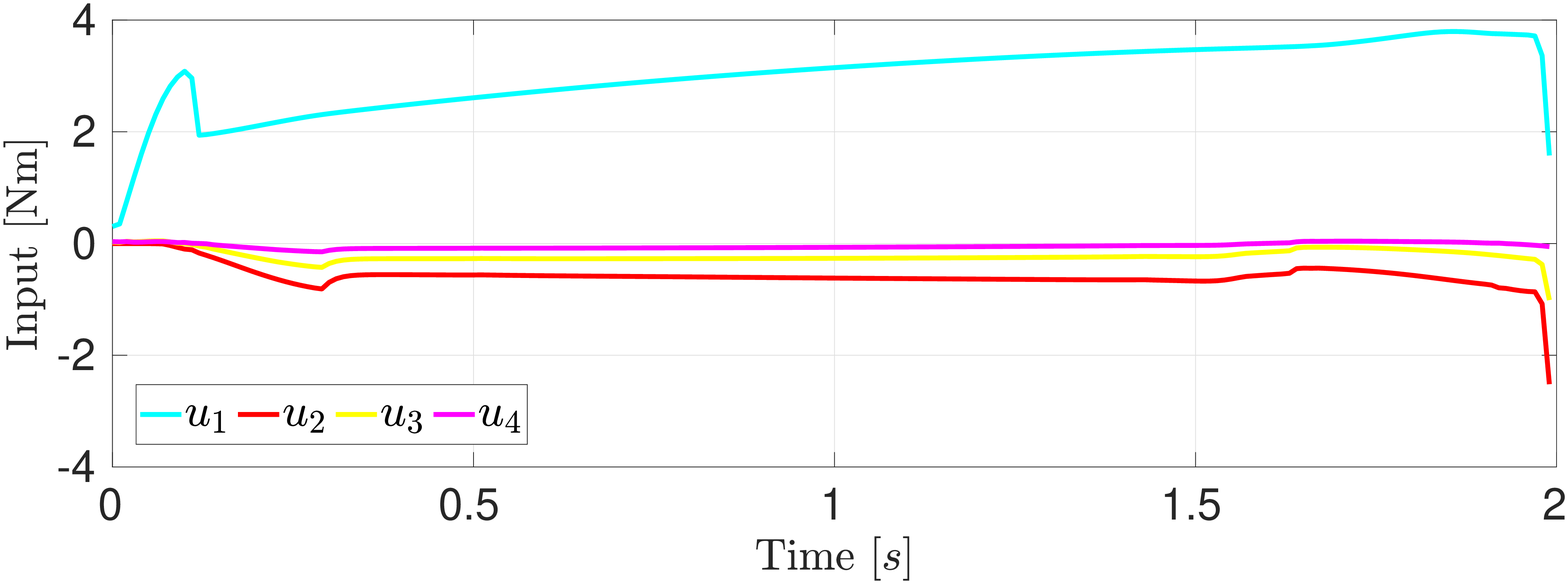}}\vspace{-0.3cm}
\subfigure[Stiffness Evolution]
{\includegraphics[width=0.95\columnwidth]{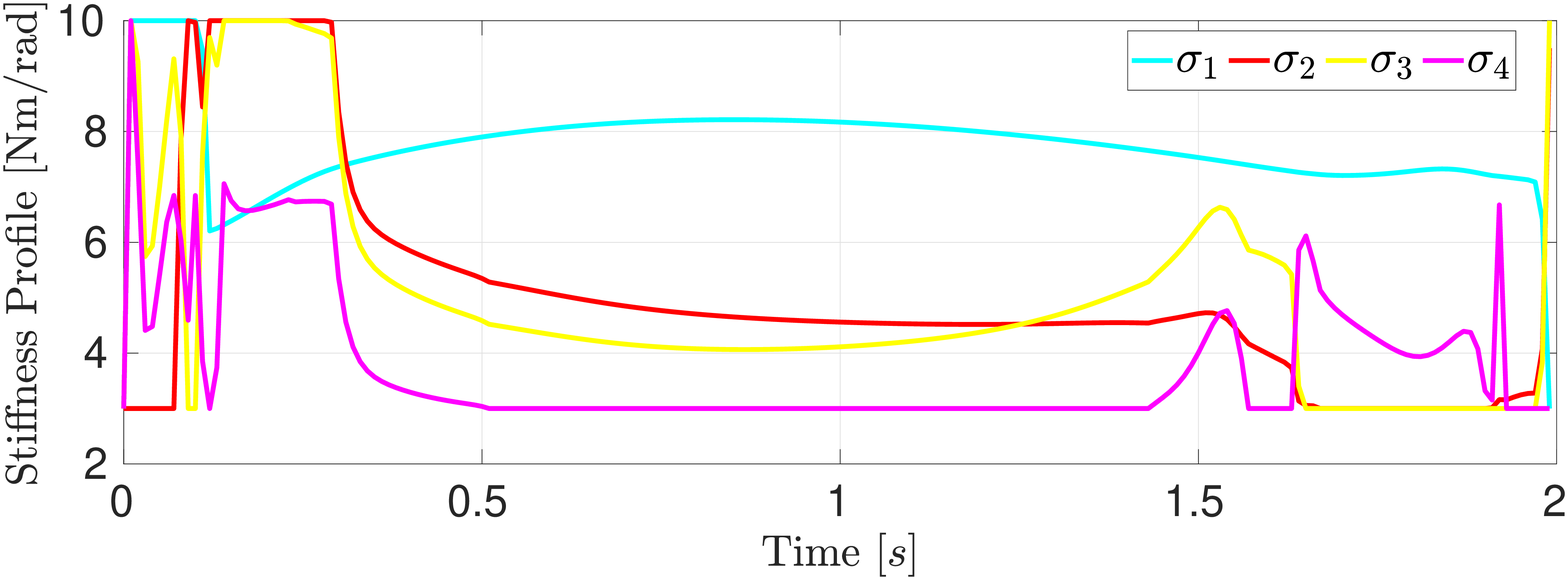}}\vspace{-0.3cm}
\subfigure[Link 1 position]
{\includegraphics[width=0.95\columnwidth]{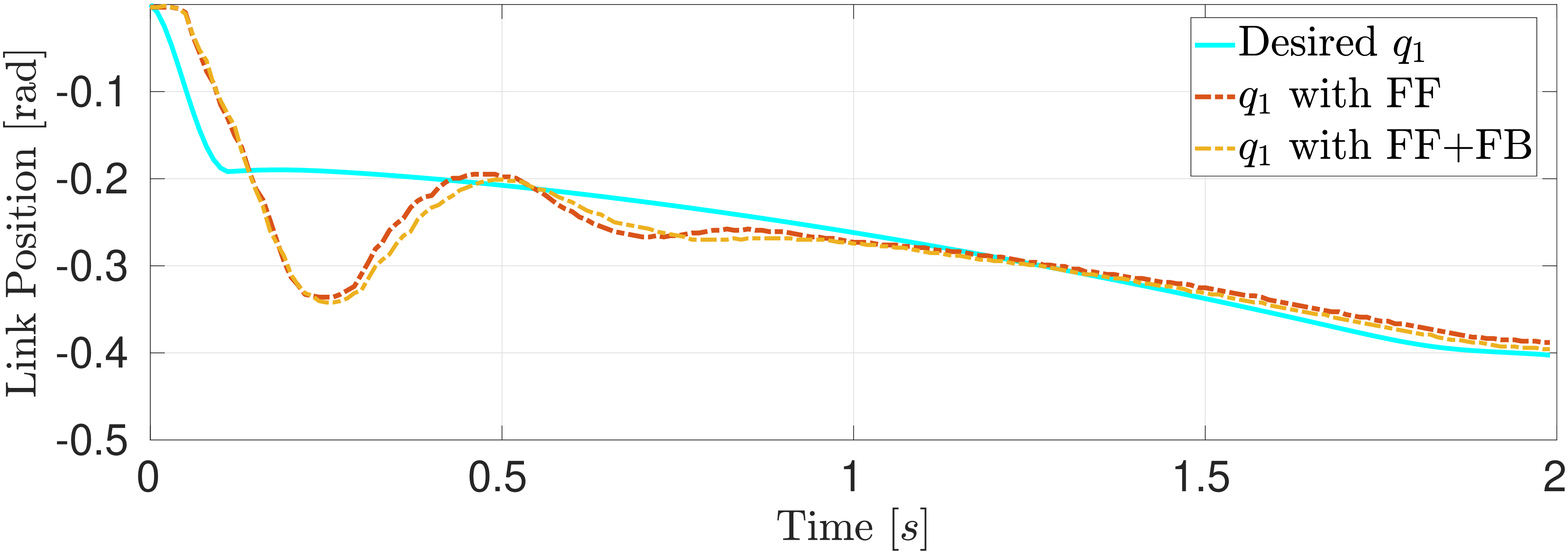}\label{fig:4dof_vsa1}}\vspace{-0.3cm}
\subfigure[Link 2 position]
{\includegraphics[width=0.95\columnwidth]{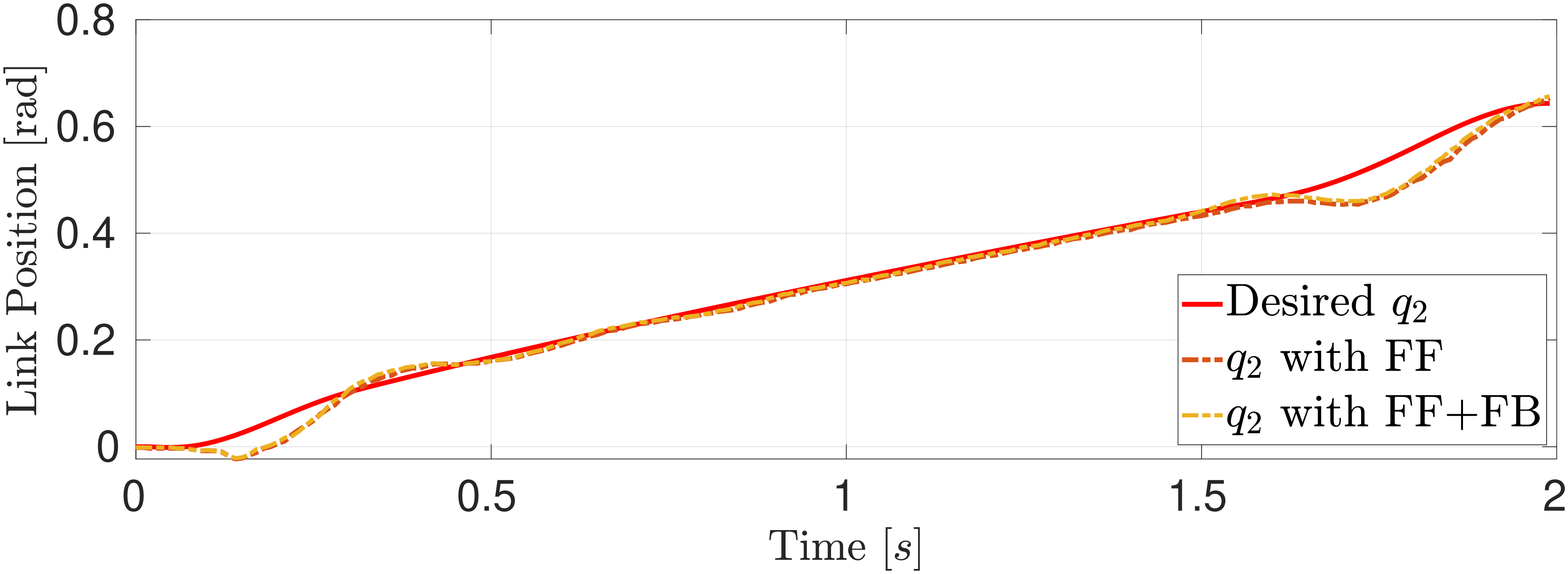}\label{fig:4dof_vsa2}}\vspace{-0.3cm}
\subfigure[Link 3 position ]
{\includegraphics[width=0.95\columnwidth]{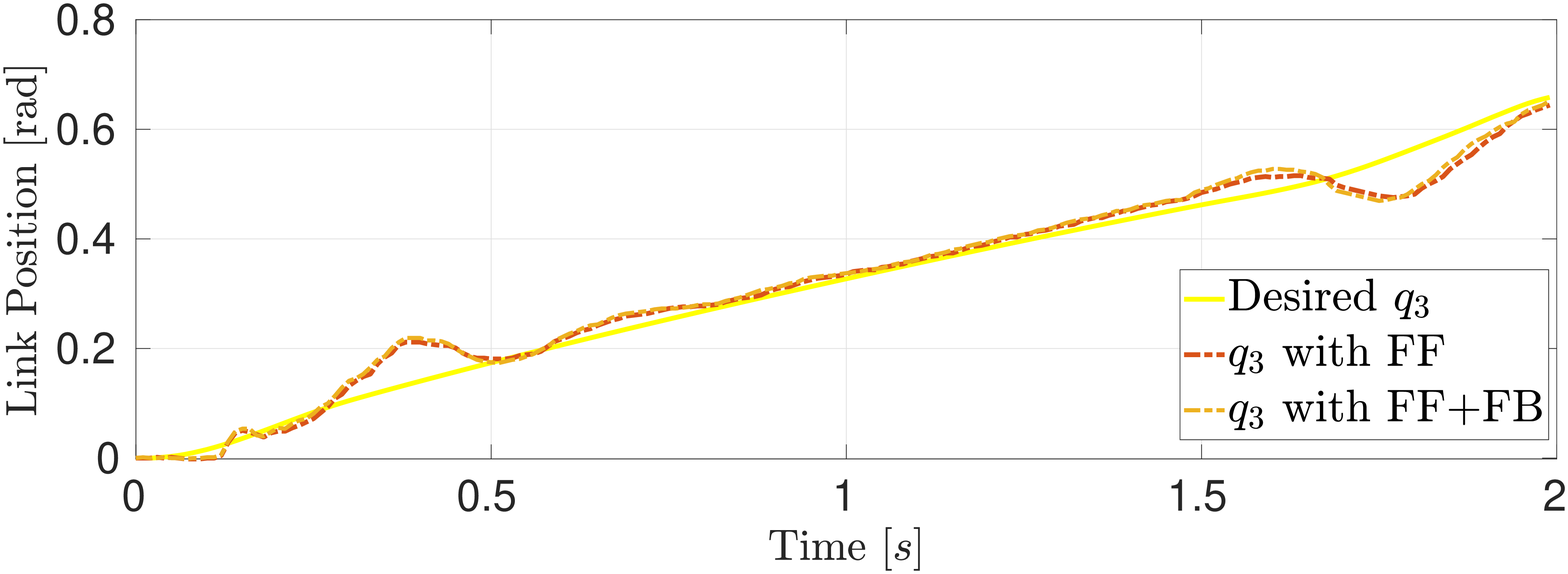}\label{fig:4dof_vsa3}}\vspace{-0.3cm}
\subfigure[Link 4 position ]
{\includegraphics[width=0.95\columnwidth]{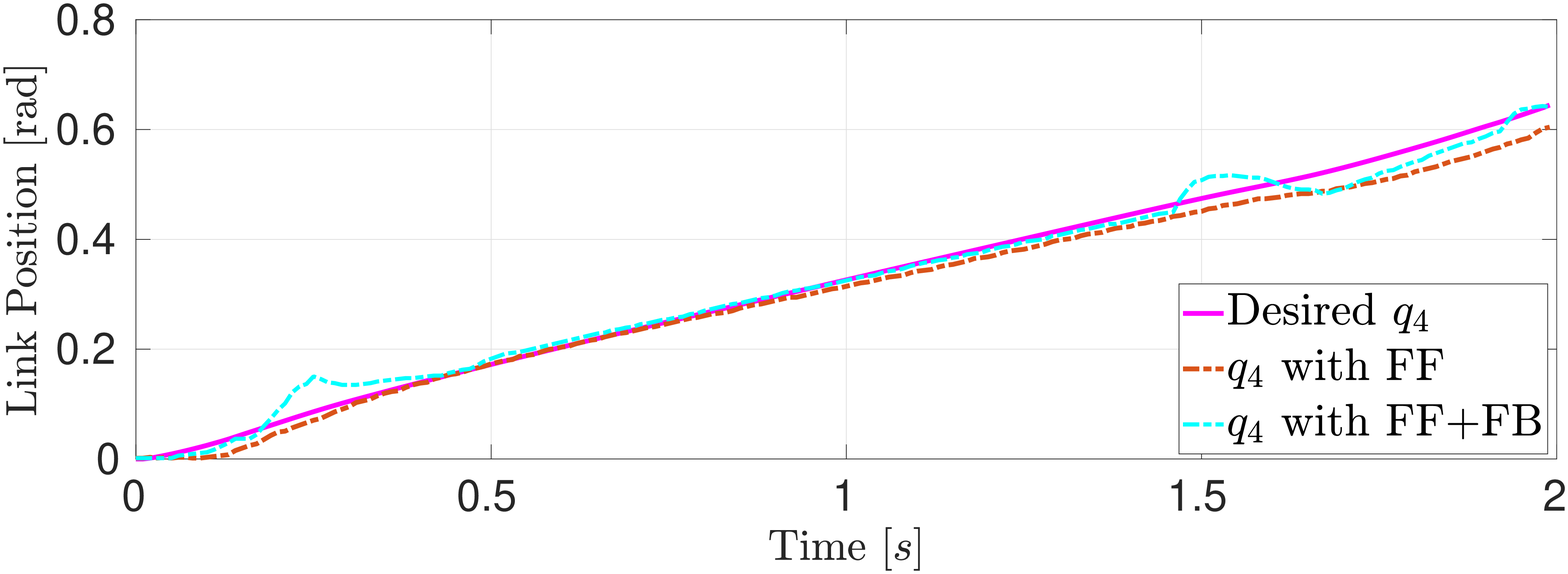}\label{fig:4dof_vsa4}}\vspace{-0.3cm}
\caption{\new{End-effector regulation task of a 4DoF system with VSA in all joints. (a) Link position in simulation. 
	(b) Input torque evolution in simulation. (c) Stiffness profile.
	The desired end-effector position was [.15, .3, .15] m and it reached [0.134, 0.36, 0.13] m.
	(d),(e),(f),(g) Evolution of all the joints in experiments. We compare the desired and the link positions using pure feed-forward (FF) and feed-forward with feedback (FF+FB), which shows better performance.}  \label{fig:4dof_vsa}}
\end{figure}

\rev{
	 Fig. \ref{fig:4dof_exp} shows a configuration of 4DoF system with SEA in each joint. It also illustrates the photo sequence of the experiment for an end-effector regulation.  The desired end-effector position is  [0.1, 0.3, 0.15] m, and the method was able to generate a  trajectory that \new{reaches}  [0.11, 0.33, 0.13] m in simulations.  The link position obtained from the experiment is shown in Fig. \ref{fig:4dof_sea1}-\ref{fig:4dof_sea4}.
In the case of pure feedback control, the RMS error for the first joint was 0.0361 rad, for the  second joint was 0.0545 rad, the third joint was 0.0659 rad and the fourth joint was 0.0388 rad. Whereas using feedback control, the RMS error for the first joint was 0.0344 rad, for the second joint was 0.0550 rad, for third joint was 0.0653 rad and for the fourth joint was  0.0240 rad.

Fig. \ref{fig:4dof_vsa} shows the results for the 4Dof system with VSA in each of the joints. We show the results for an end-effector regulation task with desired end-effector position  as [.15, .3, .15] m and the end-effector position in the simulation was [0.134, 0.36, 0.13] \unit{\meter}. The link position obtained from the experiment is shown in Fig. \ref{fig:4dof_vsa1}-\ref{fig:4dof_vsa4}.
In the case of pure feedback control, the RMS error for the first joint was 0.0428 rad, for the  second joint was 0.0230 rad, the third joint was 0.0294 rad and the fourth joint was 0.0222 rad. Whereas using feedback control, the RMS error for the first joint was 0.0429 rad, for the second joint was 0.0213 rad, for third joint was 0.0294 rad and for the fourth joint was  0.0136 rad.
}

\begin{figure}
\centering
  \subfigure[Link positions ]{\includegraphics[width=0.95\columnwidth]{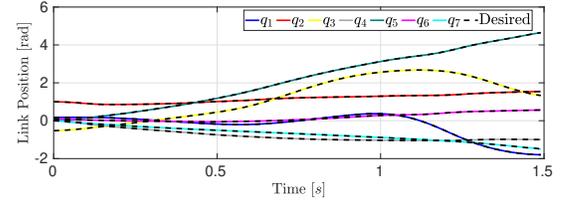}} \vspace{-0.35cm}
  \subfigure[Input torques]{\includegraphics[width=0.95\columnwidth]{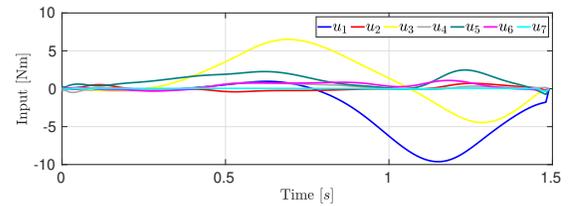}}
\caption{\rev{End-effector regulation task of a 7DoF system with SEA in all joints. (a) Evolution of the link positions in simulation. (b) Input torque evolution in simulation.}\label{fig:7dof}}
\end{figure}

Using the proposed approach, we were also able to produce optimal solutions  for higher dimensional systems. In Fig. \ref{fig:7dof} we \rev{provide} the simulation results, which includes the joint positions and the input sequence, for \rev{a} 7DoF system with SEA at each joint.

\begin{figure}
	\centering
	\includegraphics[width=.312\columnwidth]{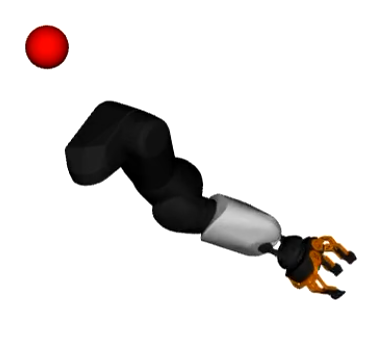}
	\includegraphics[width=.3\columnwidth]{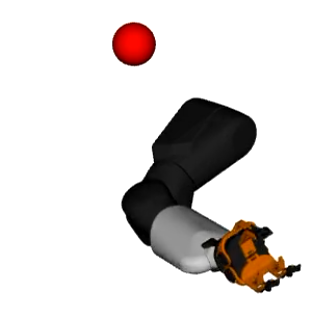}
	\includegraphics[width=.32\columnwidth]{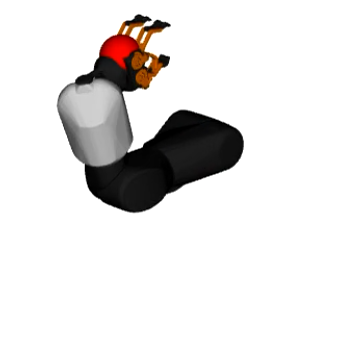}
	\caption{ \rev{Motion of the 7DoF arm with VSA in all the joints performing an end-effector regulation task. The red ball indicate the desired position.} \label{fig:7dof_ps}}
\end{figure}

\begin{figure}
\centering
\subfigure[Link positions ]{\includegraphics[width=0.95\linewidth]{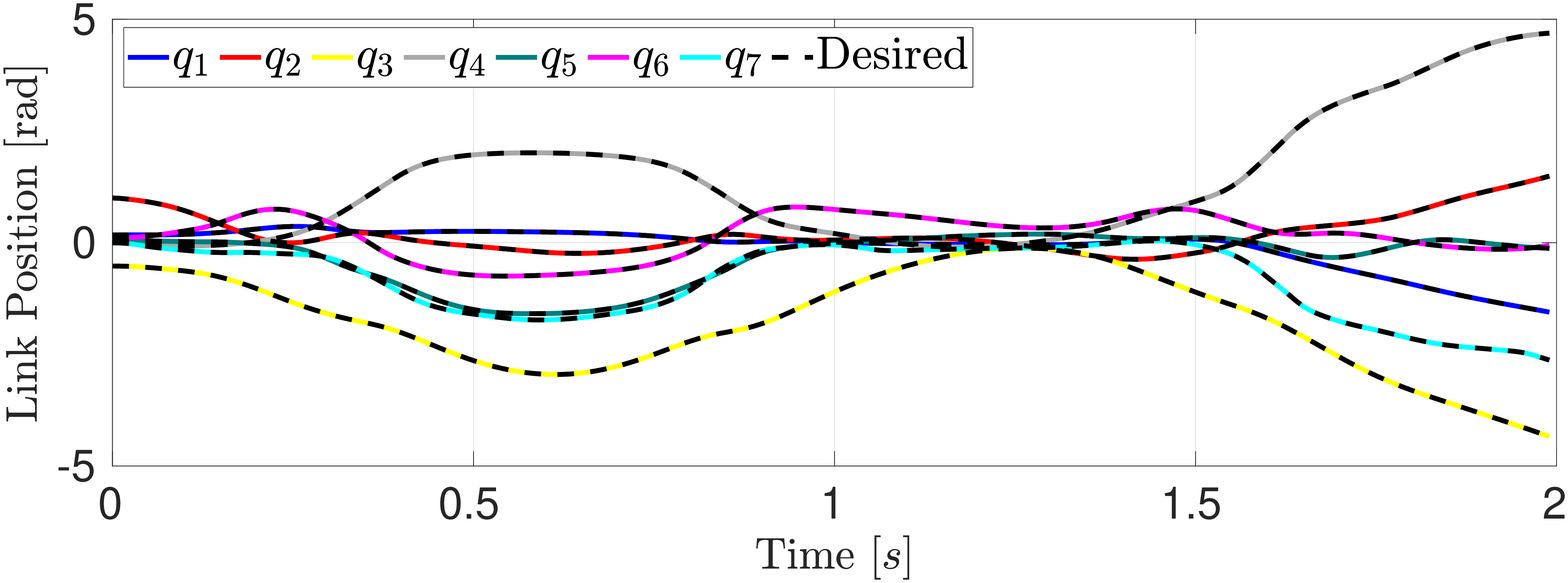}}
\subfigure[Input torques]{\includegraphics[width=0.95\columnwidth]{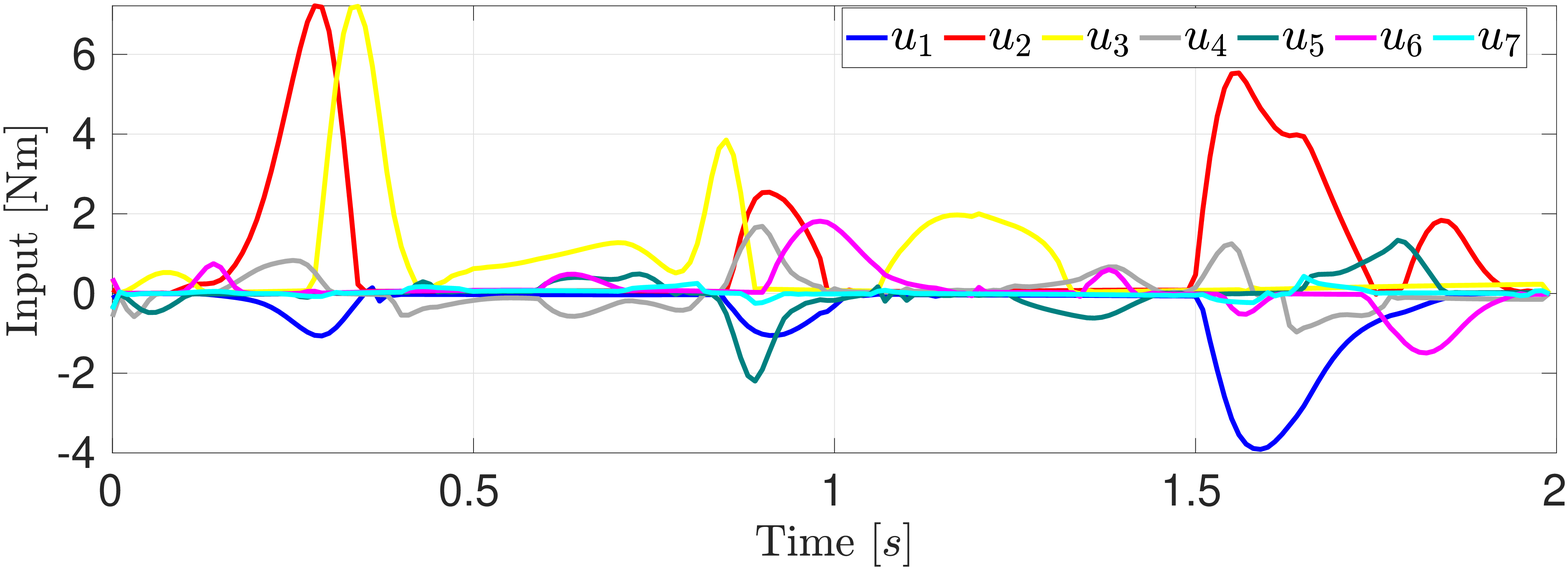}\label{fig:7dof_vsa_1}}
\subfigure[Stiffness profile]{  \includegraphics[width=0.95\columnwidth]{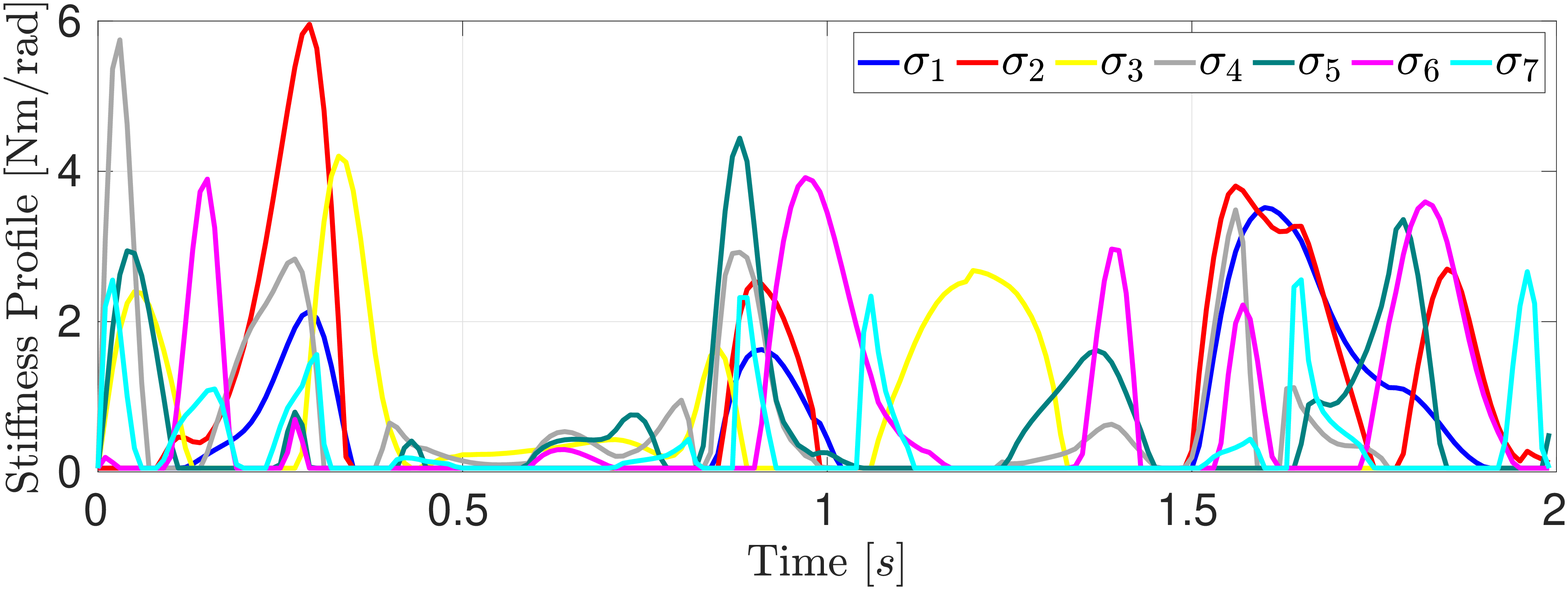}\label{fig:7dof_vsa_2}}
\caption{\rev{End-effector regulation task of a 7DoF system with VSA in all joints. (a)Evolution of the link positions in simulation. (b) Input torque evolution in simulation. (c) Input stiffness evolution in simulation.}
\label{fig:7dof_vsa}}
\end{figure}

In Fig. \ref{fig:7dof_vsa}, we \rev{provide} the simulation results, which includes the input sequence \ref{fig:7dof_vsa_1} and the stiffness profile \ref{fig:7dof_vsa_2}, for 7DoF system with VSAs at each joint.  A photo-sequence of the task is showed in Fig. \ref{fig:7dof_ps}. \change{The error in  the end-effector position for 7DoF SEA systems is $0.0091$ m and for 7DoF VSA system is  $0.005$.}

Thus, the proposed method is \rev{capable} of achieving \rev{successful} results both \rev{in the case } robots actuated by SEA and VSA. It can also be applied to platforms with high degree of freedom. We also show that feedback gain matrix is helpful in reducing the RMS error.

\new{ In comparison to earlier works \cite{mengacci}, \cite{angelini2018decentralized}, the presented method was able to synthesize dynamic motion and control with a smaller time horizon (1-4 seconds). By dynamic motions, we refer to the tasks where the contribution provided by $\mathbf{M}(\mathbf{q})  \ddot{\mathbf{q}}$, $\mathbf{C}(\mathbf{q},\dot{\mathbf{q}})$ is  comparable to the static contribution to the torque and therefore is not negligible,} \new{such that $||\mathbf{M}(\mathbf{q}) \mathbf{\ddot{q}} + \mathbf{C}(\mathbf{q},\dot{\mathbf{q}}) \mathbf{\dot{q}}|| \approx ||\mathbf{G}(\mathbf{q}) + \mathbf{K q}||$}.

Furthermore, it is worth mentioning that the employed actuator present an highly nonlinear dynamics even in the SEA case (see Section \ref{sec:simulation}). Thus, achieving good tracking performance also proves the robust aspect of the proposed method. 

\subsection{Optimal control of \change{underactuated compliant robots} } \label{sec:subsec3}

\begin{figure*}
	\centering
	\includegraphics[width=.32\columnwidth]{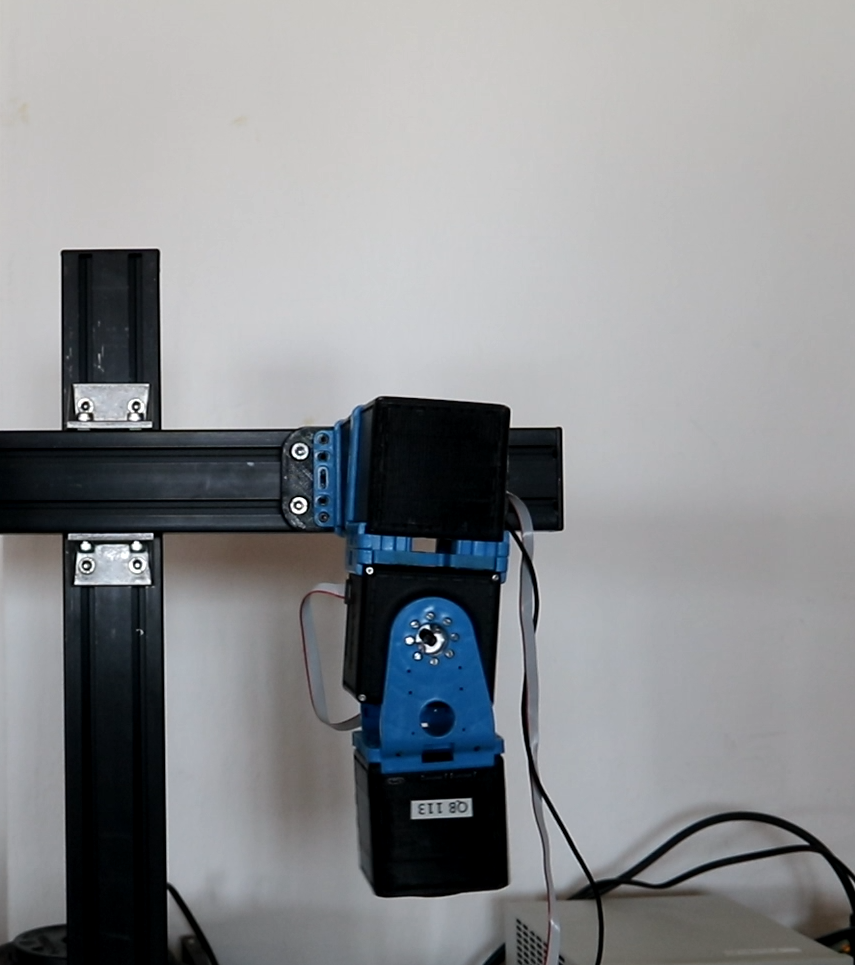}
	\includegraphics[width=.321\columnwidth]{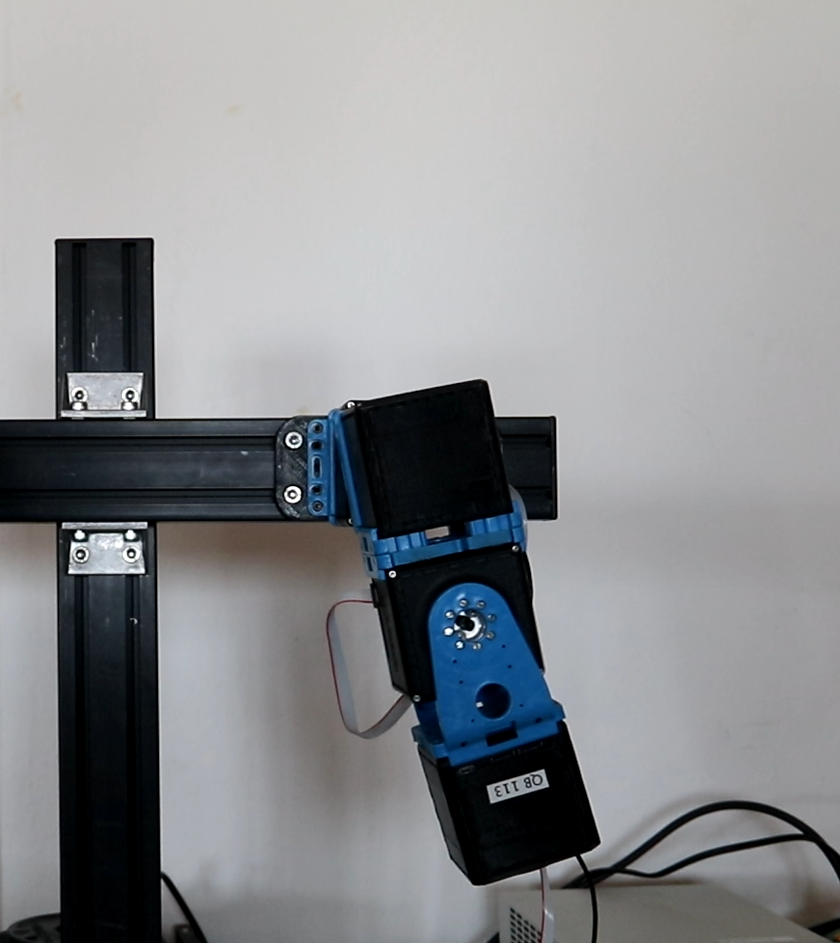}
	\includegraphics[width=.319\columnwidth]{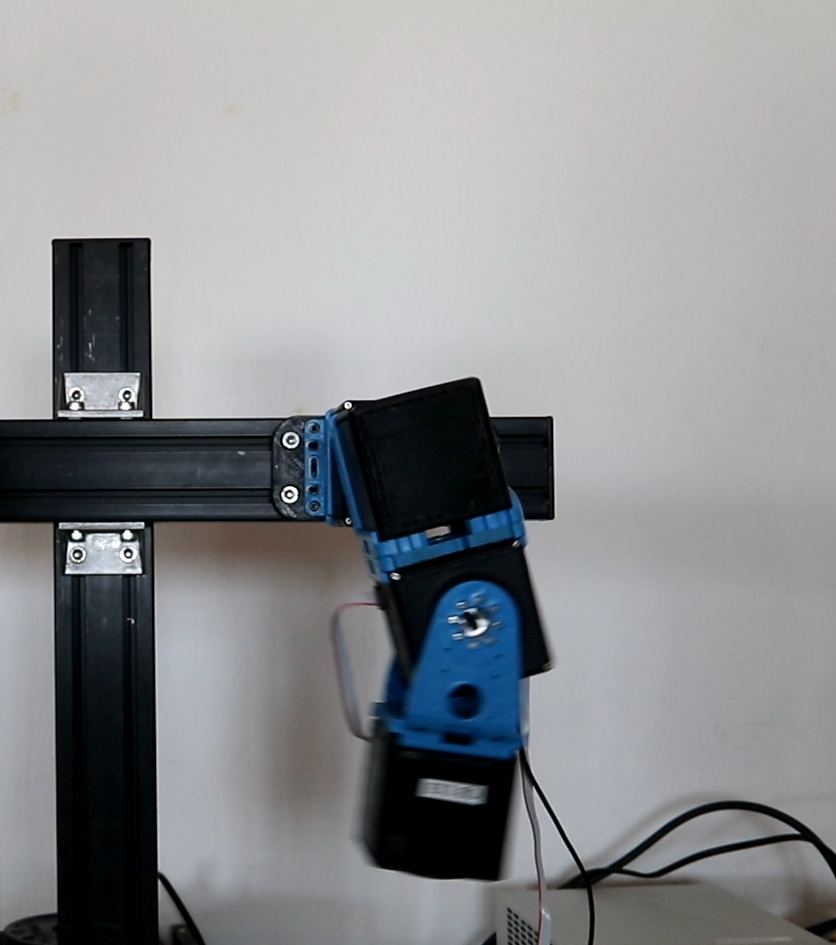}
	\includegraphics[width=.321\columnwidth]{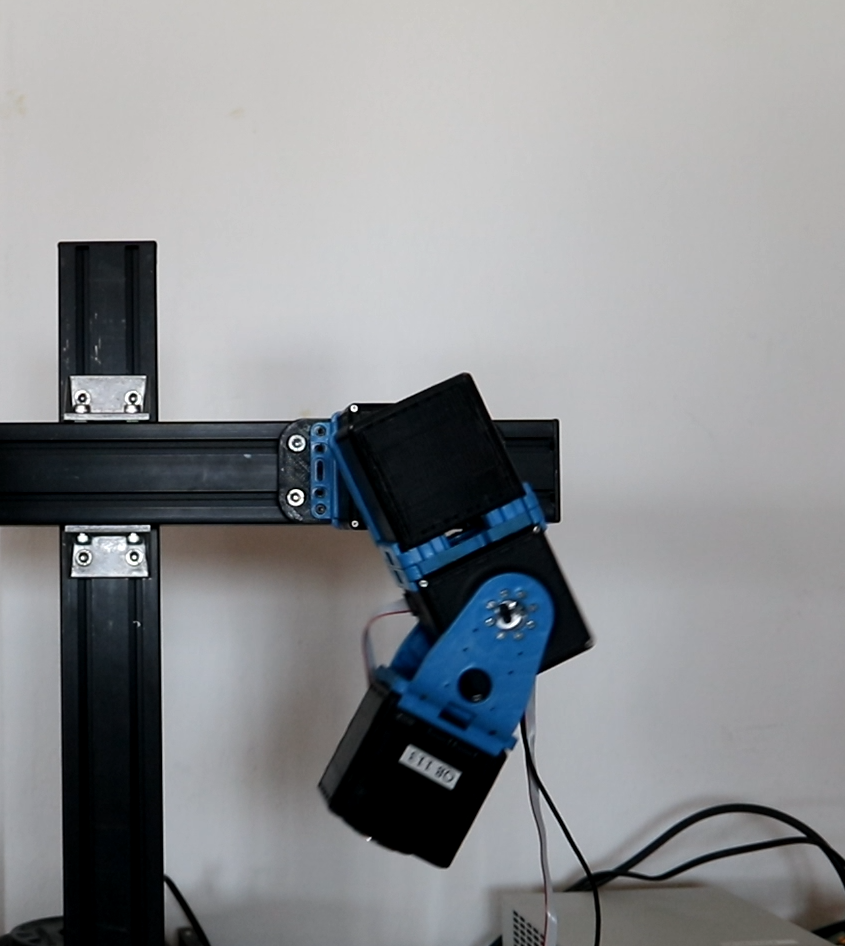}
	\includegraphics[width=.32\columnwidth]{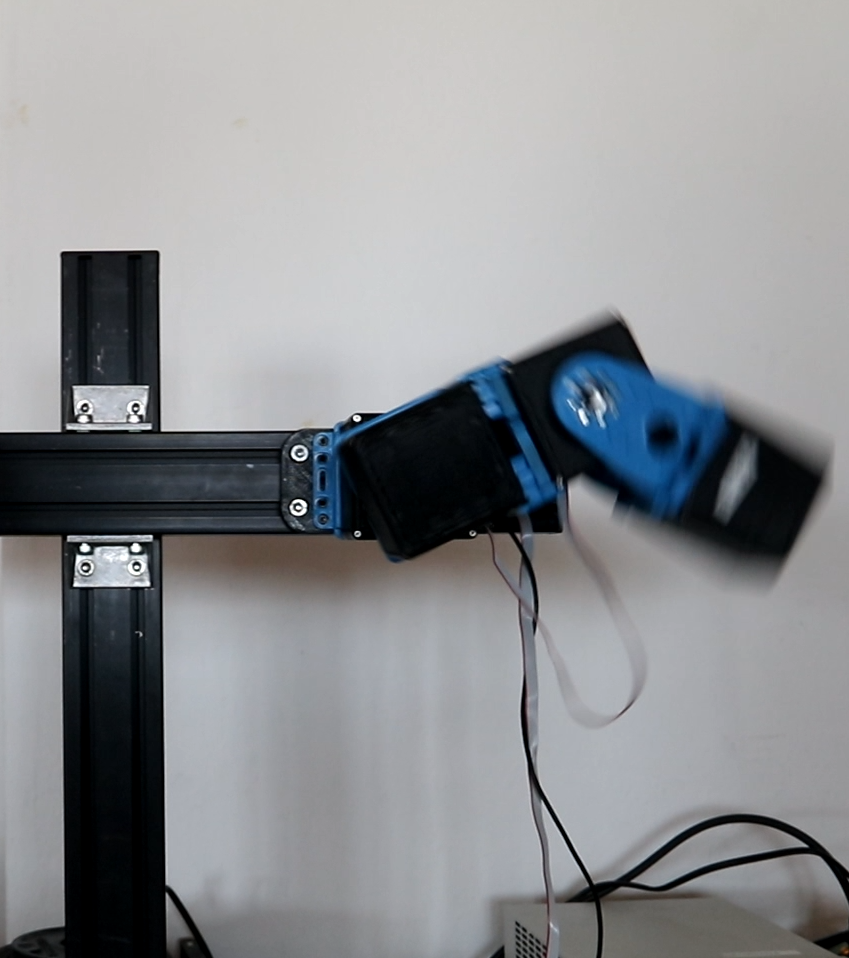}
	\includegraphics[width=.317\columnwidth]{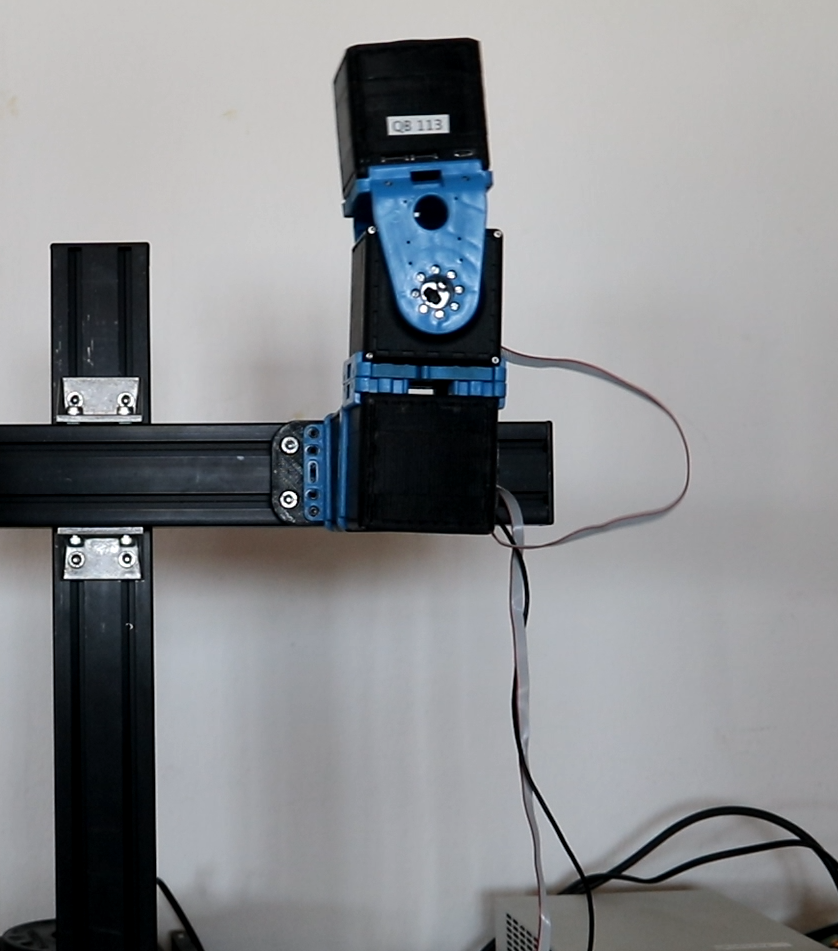}
	\caption{ Photo-sequence of the swing-up task for 2DoF \change{underactuated compliant} arm with a SEA in the first joint. Please refer to the video attachment for more details. \label{fig:sea_flex_ps}}
\end{figure*}

\begin{figure}[!tbh]
	\subfigure[\textcolor{black}{Link positions.}]{\includegraphics[width=0.95\columnwidth ]{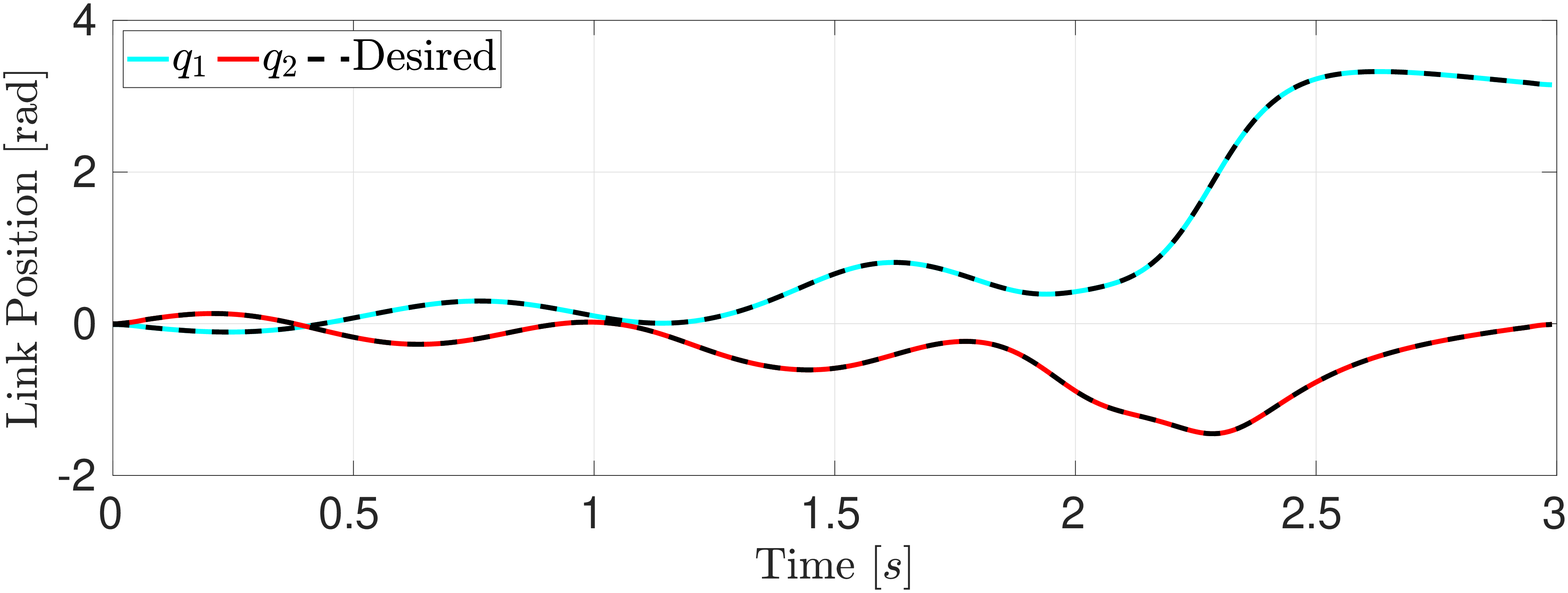}		\label{fig:sea_flex_2dof1}}\vspace{-0.3cm}
	\subfigure[\textcolor{black}{Input torque of first joint.}]{\includegraphics[width=0.95\columnwidth]{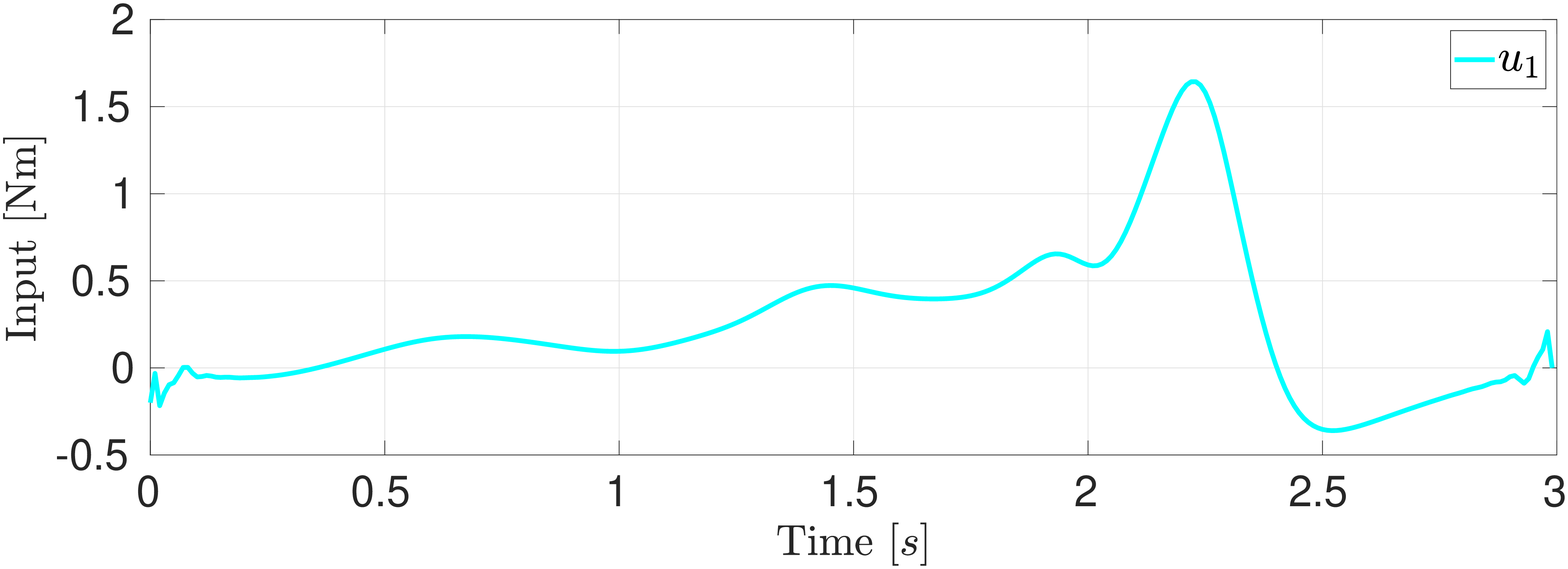}		\label{fig:sea_flex_2dof2}}\vspace{-0.3cm}
	\subfigure[\textcolor{black}{Links position.}]{\includegraphics[width=0.95\columnwidth]{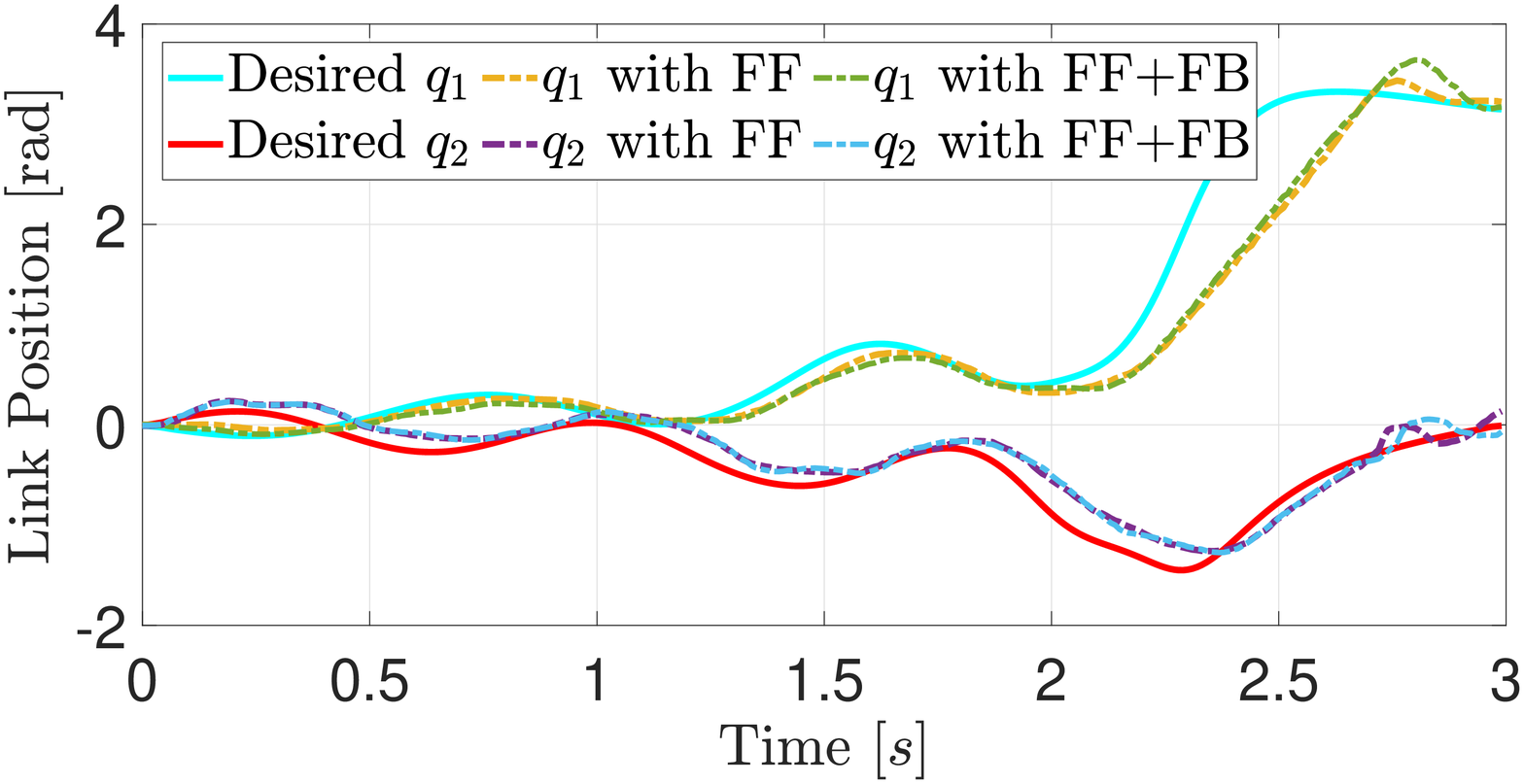}		\label{fig:sea_flex_2dof3}}\vspace{-0.3cm}
	\caption{\new{Swing-up task for the 2DoF underactuated compliant robot with a SEA in the first joint. (a) Joint evolution in simulation. (b) Input torque of the first joint in simulation. (c) Evolution of joint 1 and joint 2 in experiments. We compare the desired and the link positions using both  the pure feed-foward (FF) and the feedback  plus feed-forward (FF+FB) cases,  which shows  better performance in the latter.}
  \label{fig:sea_flex}}
\end{figure}

Fig. \ref{fig:sea_flex} shows the simulation and experimental results of \rev{the} swing-up task performed by \rev{the} 2DoF \change{\change{underactuated compliant}} arm with a SEA  in \rev{the} first joint. This includes the optimal trajectory (Fig. \ref{fig:sea_flex_2dof1}) and the input sequence (Fig. \ref{fig:sea_flex_2dof2}). \new{Fig. \ref{fig:sea_flex_2dof3} illustrates the link positions of both the joints obtained from the experiments.} Snapshots of the experiments \change{are} depicted in Fig. \ref{fig:sea_flex_ps}; please \rev{also} refer to the video attachment.
The RMS error for joint 1 in \rev{the} case of pure feed-forward control was $0.3908$ \unit{\radian} and in \rev{the} case of feedforward plus feedback control was $0.3734$ \unit{\radian}. Similarly for joint 2 we observe that RMS for pure feed-forward case was $0.1607$ \unit{\radian} and for feedforward plus feedback control was $0.1571$ \unit{\radian}.


\begin{figure}
\subfigure[Input torques]{\includegraphics[width=0.95\columnwidth]{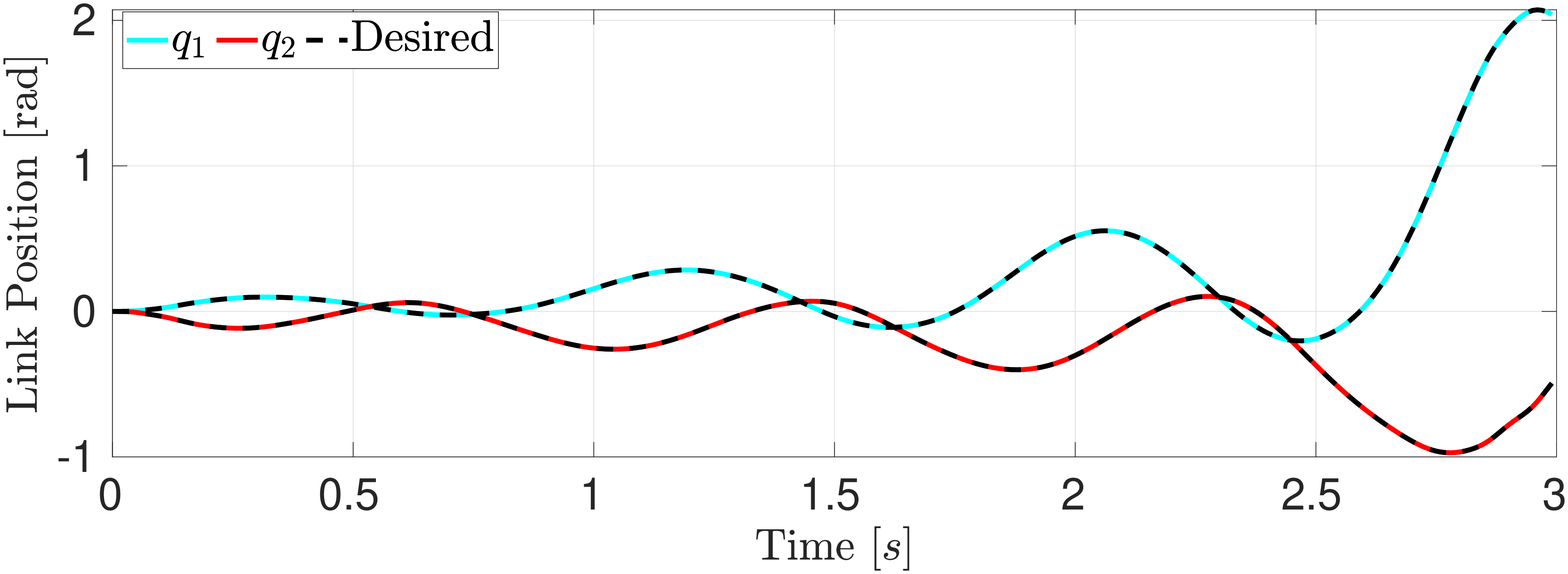}\label{fig:vsa_flex_2dof5}}\vspace{-0.3cm}
  \subfigure[Input torques]{\includegraphics[width=0.95\columnwidth]{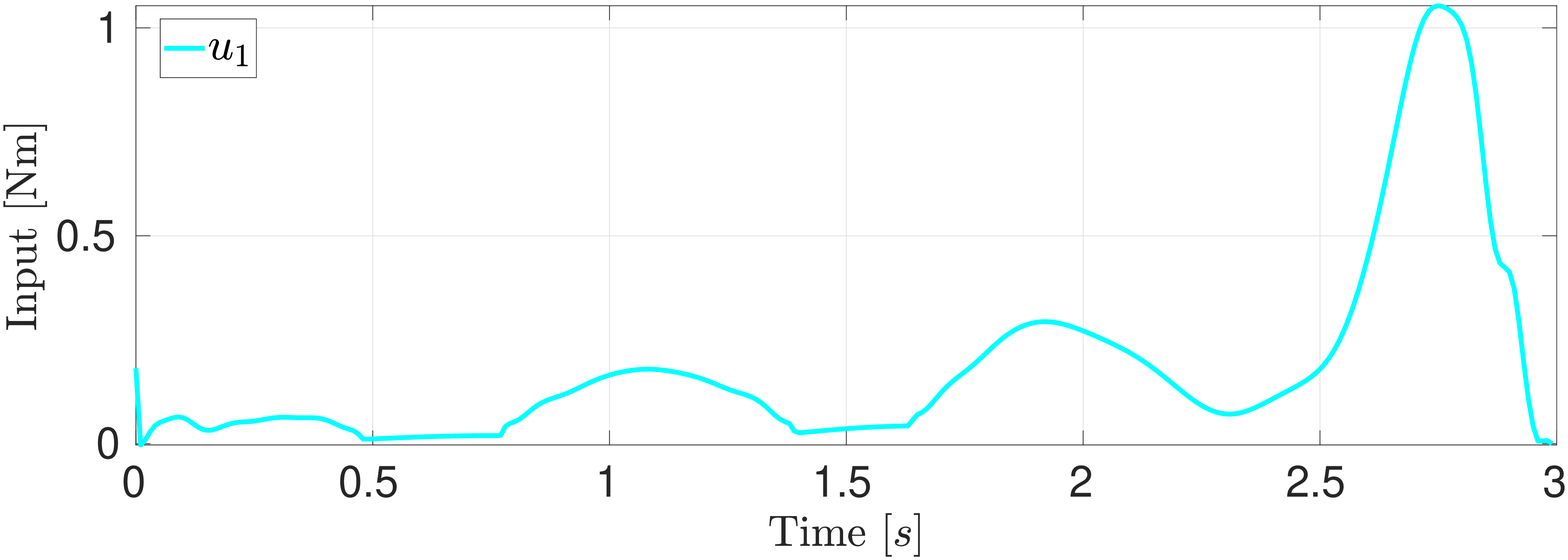}\label{fig:vsa_flex_2dof1}}\vspace{-0.3cm}
    \subfigure[Stiffness profile]{\includegraphics[width=0.95\columnwidth]{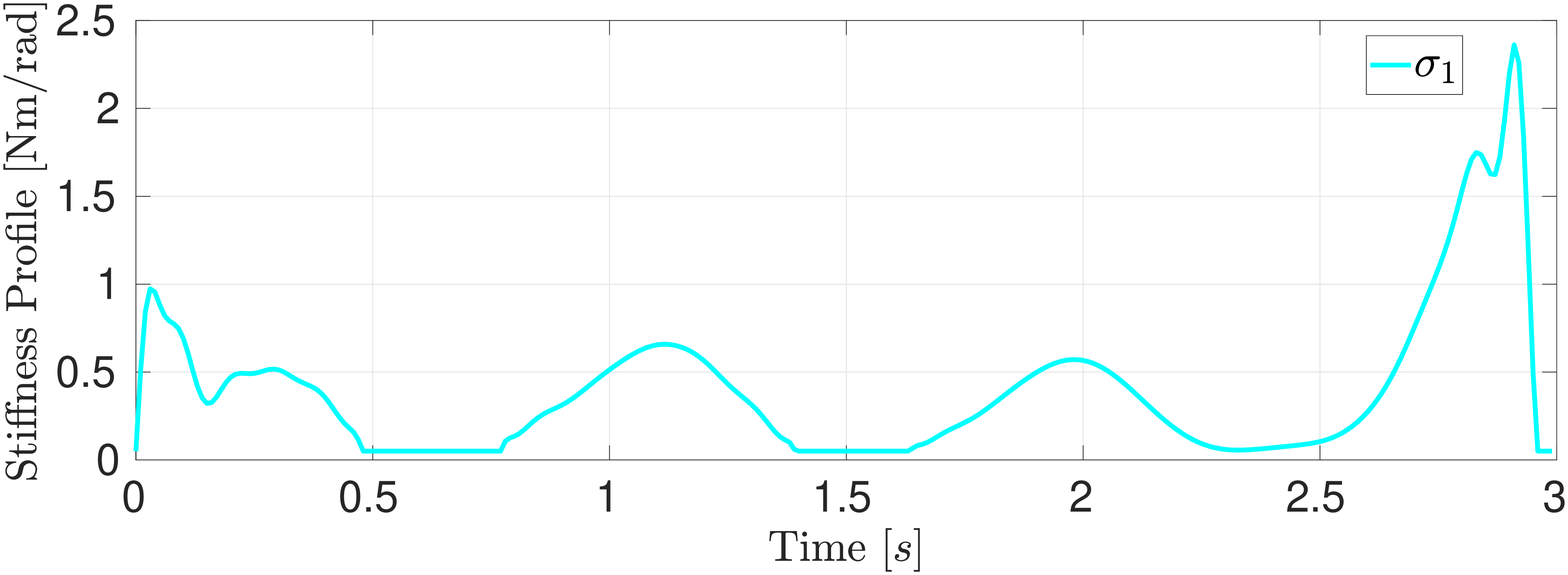}\label{fig:vsa_flex_2dof2}}\vspace{-0.3cm}
    \subfigure[Links positions]{\includegraphics[width=0.95\columnwidth]{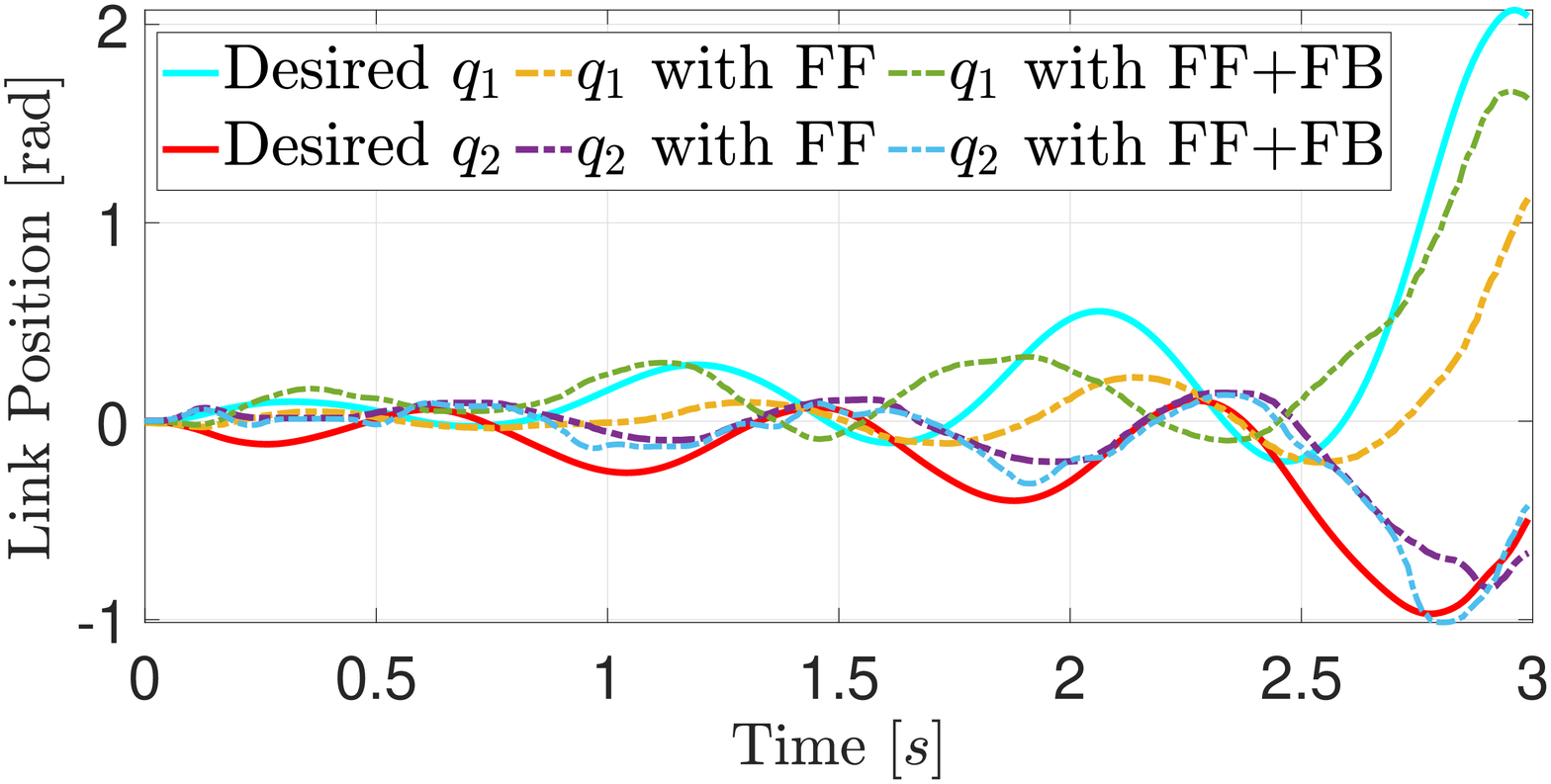}\label{fig:vsa_flex_2dof3}}\vspace{-0.3cm}
  \caption{\new{Swingup task for the 2DoF underactuated compliant arm with a VSA in the first joint. 
  (a) Input torque evolution in simulation. (b) Input stiffness evolution in simulation.
  (c) Evolution of joint 1 and joint 2 in experiments. With pure feed-foward (FF) and the feedback  + feed-forward(FF+FB) cases, we compare the desired and the link positions,  which shows  better performance in the later.}
  \label{fig:vsa_flex}
  }
\label{fig:vsa_flex_2dof_exp}
\end{figure}

The simulation and experimental results for the swing-up task  of \rev{the} 2DoF \change{underactuated compliant} arm with VSA in the first joint \rev{are} shown in Fig. \ref{fig:vsa_flex}.  The RMS error for joint 1 in \rev{the} case of pure feed-forward control was $0.3857$ \unit{\radian} and in \rev{the} case of feedforward plus feedback control is $0.2068$ \unit{\radian}. Similarly, for joint 2 we observe that RMS for the pure feed-forward case was $0.1701$ \unit{\radian} and for feedback control was $0.1341$ \unit{\radian}. 
So, the use of feedback gains reduces the error and helps to stabilize the upward-pointing position in both cases.
\new{Similarly, an underactuated 4DoF system can also be stabilized in the vertical position since the proposed controller uses state-feedback controller (and given the system reachable in the vertical equilibrium).}

\begin{figure}[t]
		\includegraphics[width=.195\linewidth]{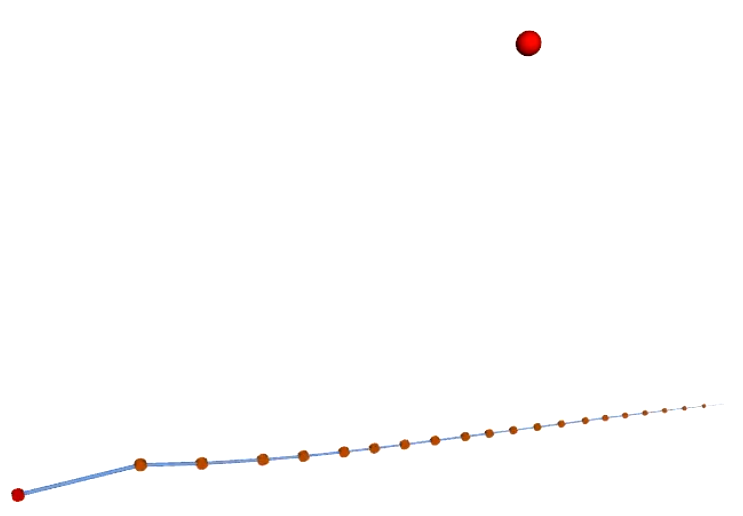}
			\includegraphics[width=.195\linewidth]{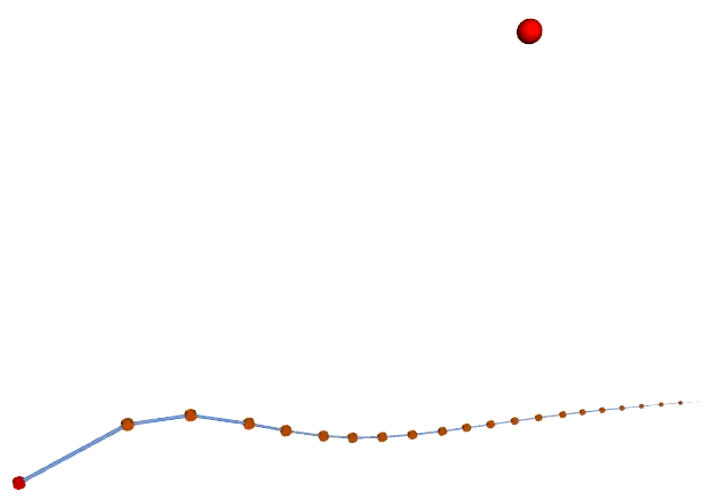}
			\includegraphics[width=.165\linewidth]{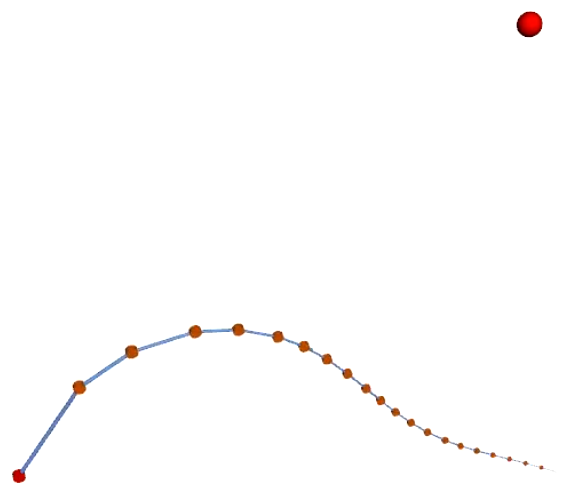}
			\includegraphics[width=.165\linewidth]{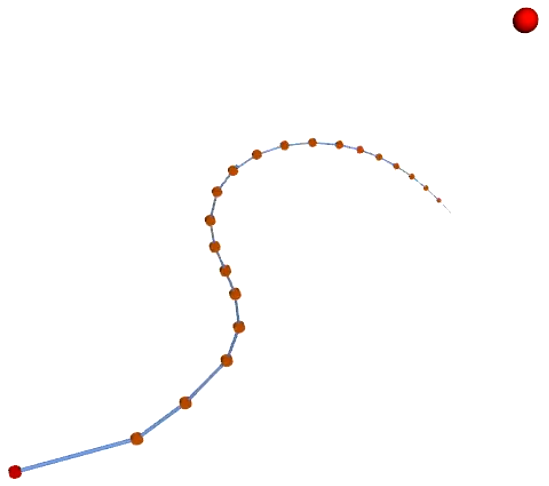}
			\includegraphics[width=.17\linewidth]{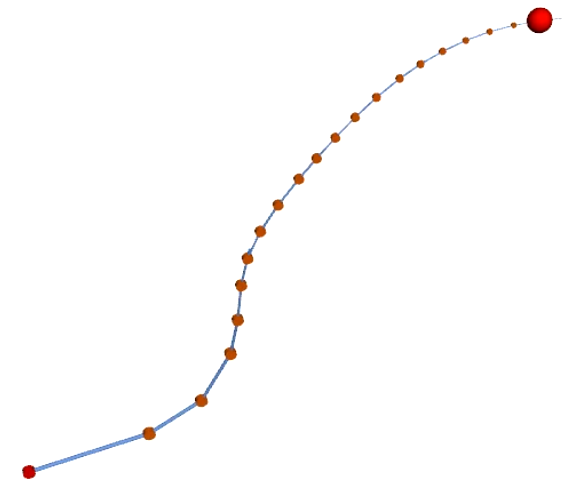}
			\caption{\rev{Photo sequence of motion of end-effector regulation task on under-actuated serial manipulator. Red ball  is the desired end-effector position with coordinates as [2.12, 2.12] m.}  \label{fig:fishing_rod}}
\end{figure}

\begin{figure}
	\centering
	\subfigure[Link positions ]{\includegraphics[width=1\columnwidth]{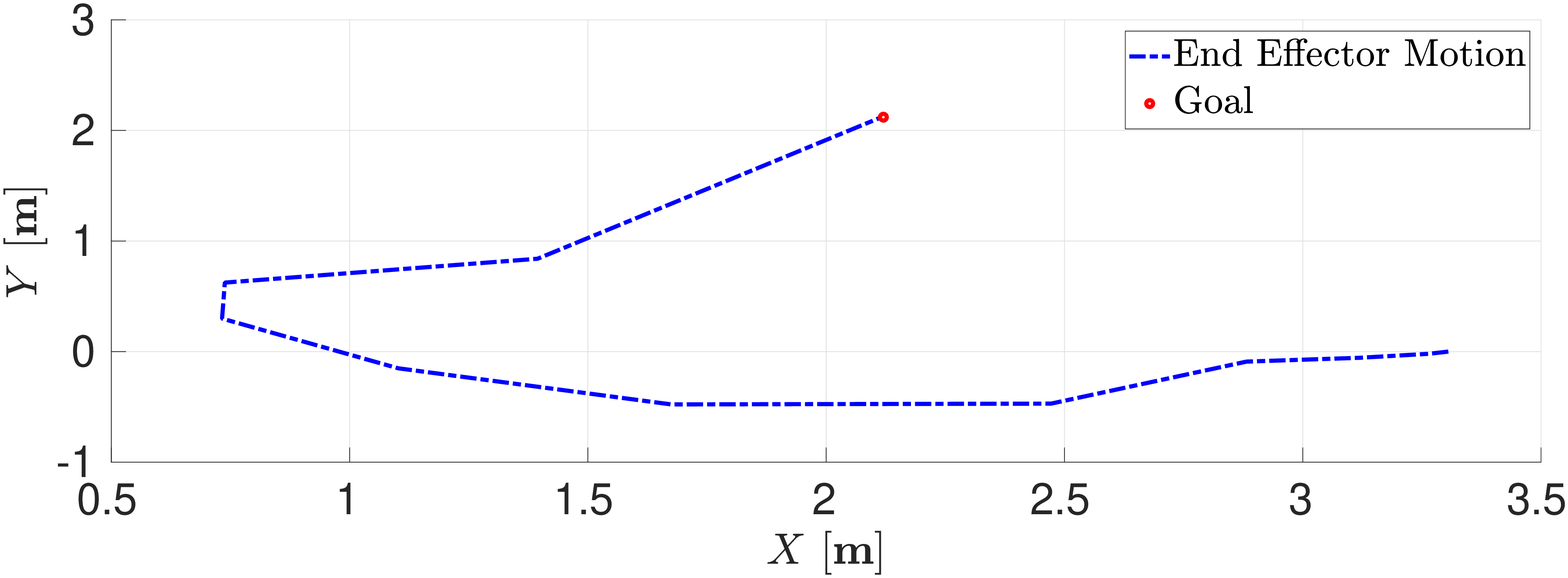}} \vspace{-0.35cm}
	\subfigure[Input torques]{\includegraphics[width=0.97\columnwidth]{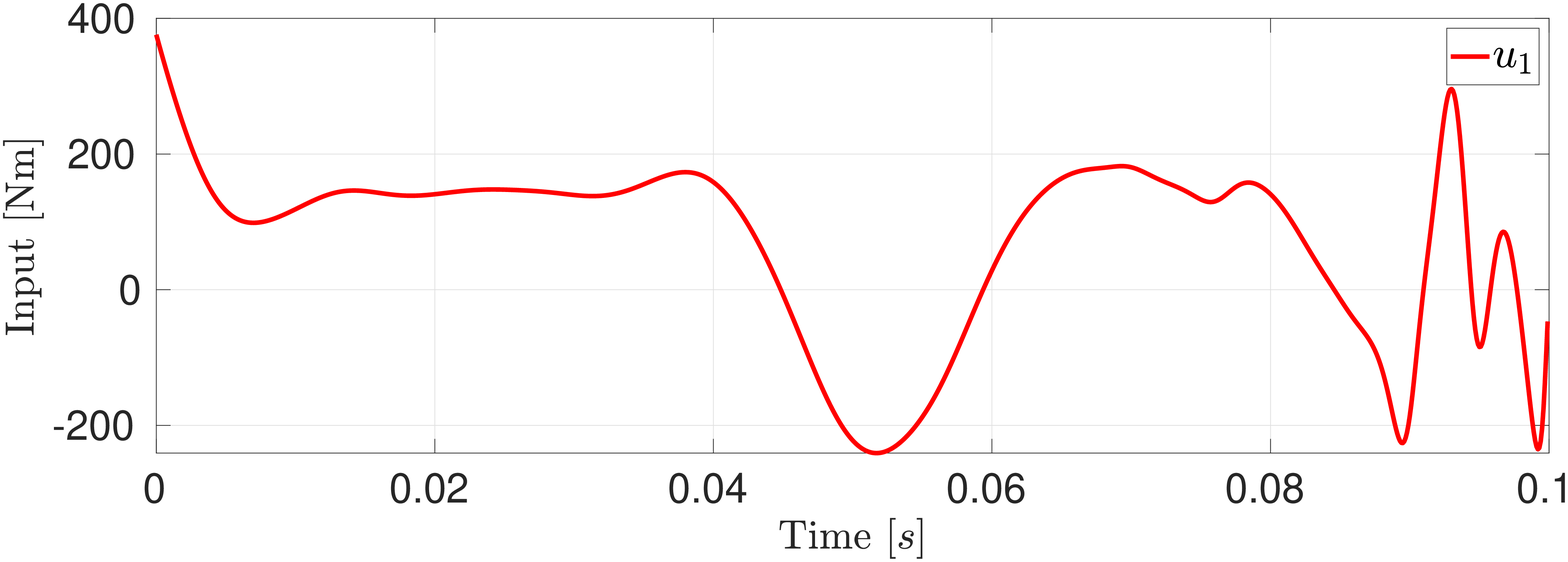}}
	\caption{\rev{End-effector regulation task of a underactuated serial manipulator. (a) Evolution of the link positions in simulation. (b) Input torque evolution in simulation.}\label{fig:fishing_rod_plots}}
\end{figure}

\rev{
Fig \ref{fig:fishing_rod} illustrates the motion synthesized by the proposed algorithm for an underactuated serial manipulator with 21 joints and only one actuated joint. The task presented is an end-effector regulation task with the desired end-effector position being: $[2.12 , 2.12]$ m and the terminal velocity being: $[7.07, 0]$ m/s with the error in simulation being 0.006 m. In Fig. \ref{fig:fishing_rod_plots}, we present the end-effector motion and the input torques.}

Thus, the proposed method is successful in planning optimal trajectories for \change{under-actuated compliant systems} as well.
\rev{ Especially the simulation results with an under-actuated serial manipulator with 20 passive joints are promising as the error in the end-effector position is only $0.006$ m.}

The results presented here \change{neglect} the exact nonlinear actuator dynamics, and thus the tracking performance on the experimental setup illustrates the robustness of the method.

\subsection{Energy consumption}\label{sec:subsec4}
The sum of torque squared over the whole trajectory is assumed to be the \rev{most suitable} candidate to compare different controllers for power consumption\cite{torque_square} i.e., $T = \sum_{k=0}^N \tau_k ^2  $.
\change{Using this metric}, \change{we conduct simulations and } show that elastic actuation reduces  energy consumption. To illustrate this we use end effector regulation task for \rev{a} fully actuated 2DoF system, \rev{a} fully actuated 7DoF and \rev{a} 2DoF \change{underactuated compliant arm}. For each of the systems, the cost weights are kept the same across rigid, SEA and VSA actuation for \rev{a} fair comparison.

Table \ref{tab:Power_comparison_rigid_soft} shows that use of SEA and VSA lowers the energy consumption for all the three systems.  

\begin{table}
\caption{Power consumption comparison between rigid and soft actuators}\vspace{-1em}
\label{tab:Power_comparison_rigid_soft}
\begin{center}
\begin{tabular}{@{} l rcc r rcc @{}}
\toprule

\emph{Problems}       &rigid                       &     &SEA & &VSA              \\
\midrule
\texttt{2DoF}         &$142.07$  & &$138.46$    & &$84.17$      \\
\texttt{2DoF Flexible}      &$101.019$ &  &$87.53$ & &$50.84$\\
\texttt{7DoF}       &$>10000$   &  &$6112.77$ & &$4545.26$\\

\bottomrule
\end{tabular}
\end{center}
\end{table}

\section{Conclusion and Future Work}
In this work, we presented an efficient  optimal control formulation for soft robots based on the Box-FDDP/FDDP algorithms. We proposed an efficient way to compute the dynamics and analytical derivatives of  soft articulated and \change{underactuated compliant} robots.
The state-feedback controller presented in this paper based on local and optimal policies from Box-FDDP/FDDP helped to improve the performance in swing-up and end-effector regulation tasks. 
\rev{Overall, the application of (high authority) feedback to soft robots may be an advantage or a disadvantage\cite{della2020soft}. For instance, in \cite{tomei} it is shown how feedback can stabilize unstable equilibrium points of the system. Furthermore, feedback increases the robustness of model uncertainties and disturbances. However, the negative effect of feedback is the alteration of the mechanical stiffness of the system \cite{della2017controlling, angelini2018decentralized} which defeats the purpose of building soft robots. In fact, compliance is purposefully inserted into robots to confer them the so-called \textit{embodied intelligence} \cite{mengaldo2022concise}. This derives from the interaction between the system body, its sensory-motor control, and the environment, and it has the goal to simplify the robot control thoughtfully inserting complexity, and intelligence, into the robot body \cite{pfeifer2009morphological, hara2003morpho, rus2015design}. For this reason, depending on the specific task, the application of feedback to soft robots should be carefully approached. Future work will focus on preserving the natural behavior of the controlled soft robot including limitations to the feedback authority in the control problem similar to 
\cite{pierallini2022swing}.}

Another future research direction would \change{consist of extending} the formalism to soft articulated legged robots. \change{Accounting for} compliance in legged robots is expected to improve the performance compared to existing approaches. The use of feedback alters  compliance in the system. This behavior is undesirable as it defies the sole purpose of adding the soft elements in the first place. One possible research direction 
would be to explore a DDP-based algorithm, maybe in a bi-level setting, to obtain feedback gains that respects system compliance and still get high performance.
Further, an MPC solution based on the proposed framework is seen as a natural extension of this work.
\vspace{3mm}

\bibliographystyle{unsrt}
\bibliography{references}

\begin{IEEEbiography}[{\includegraphics[width=1in,height=1.25in,clip,keepaspectratio]{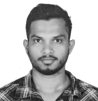}}]
	{Saroj Prasad Chhatoi} received his M.Sc. degree in physics from the Institute of Mathematical Sciences, Chennai, India in 2019.
	
	He is currently pursuing a Ph.D. degree in Automatic Control at LAAS-CNRS, Toulouse, France.
	His research interests include Automatic Control,legged locomotion, control of soft robotic systems, model predictive control.
\end{IEEEbiography}

\begin{IEEEbiography}[{\includegraphics[width=1in,height=1.25in,clip,keepaspectratio]{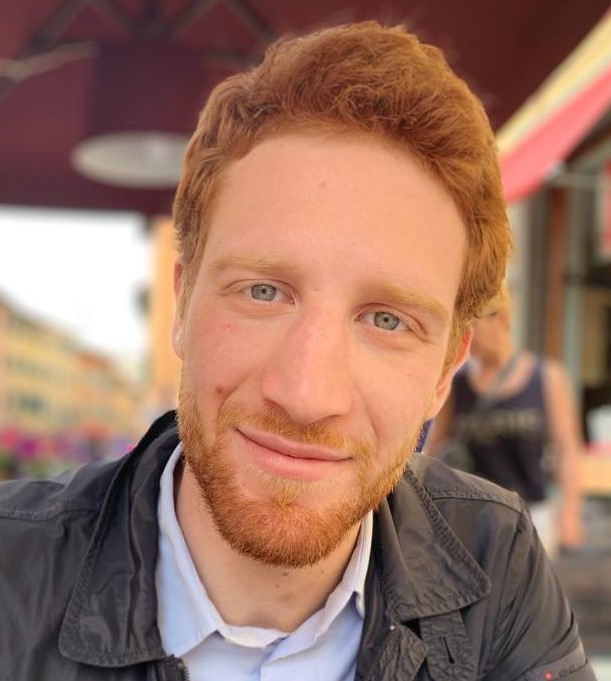}}]{Michele Pierallini}
	received the B.S. degree in biomedical engineering in 2017 and M.S. degree (cum laude) in automation and robotics engineering in 2020 from the University of Pisa, Pisa, Italy, where he is currently working toward the Ph.D. degree in robotics at the Research Center “Enrico Piaggio.” His current research focuses on the control of soft robotic systems and iterative learning control.
\end{IEEEbiography}
\begin{IEEEbiography}[{\includegraphics[width=1in,height=1.25in,clip,keepaspectratio]{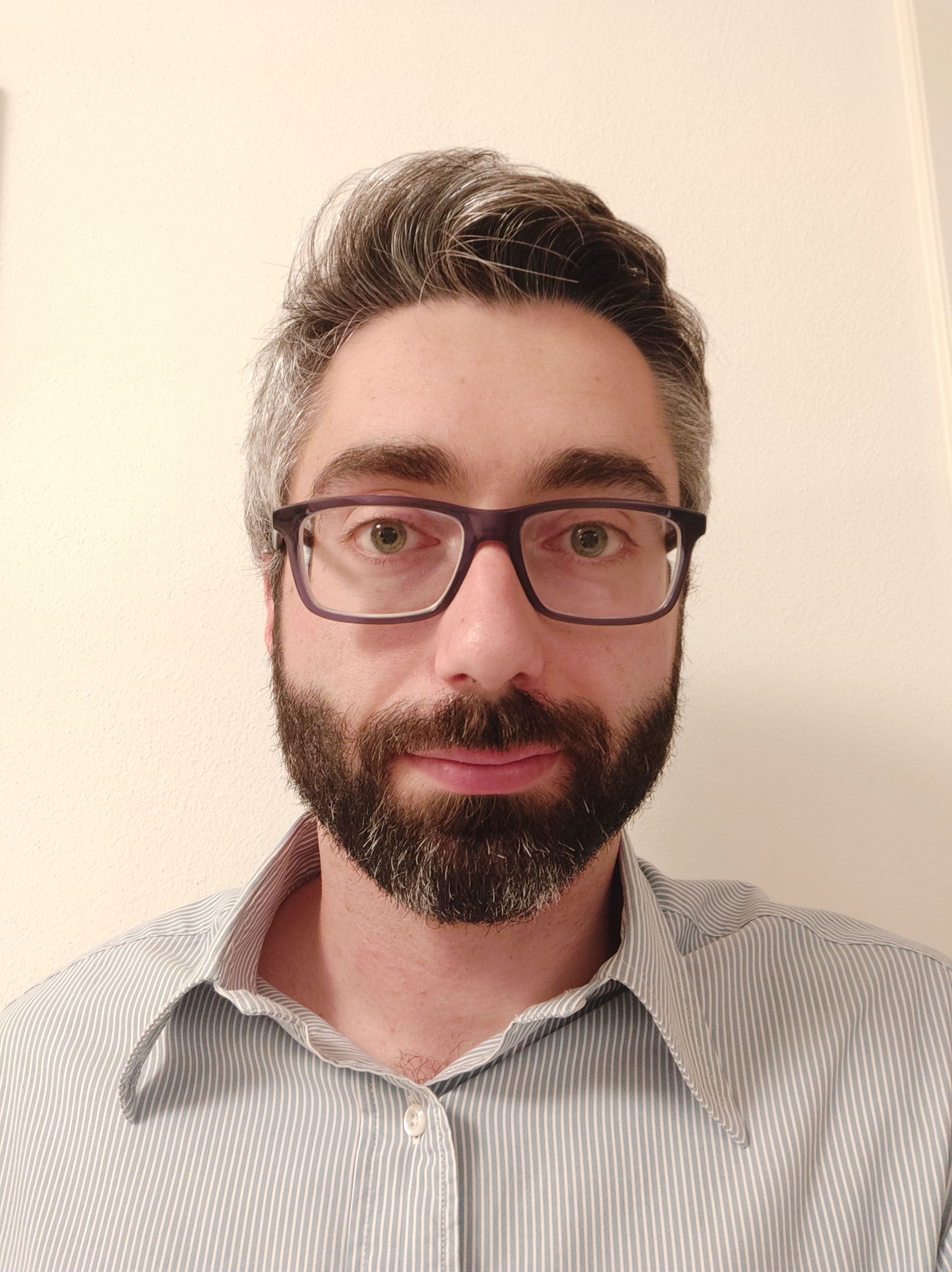}}]{Franco Angelini} received the B.S. degree in computer engineering in 2013 and M.S. degree (cum laude) in automation and robotics engineering in
	2016 from the University of Pisa, Pisa, Italy. University of Pisa granted him also a Ph.D. degree (cum laude) in robotics in 2020. Franco currently has a research fellowship at the Research Center “Enrico Piaggio”, Pisa. His main research interests are control of soft robotic systems, grasping and impedance planning.
\end{IEEEbiography}
\begin{IEEEbiography}[{\includegraphics[width=1in,height=1.25in,clip,keepaspectratio]{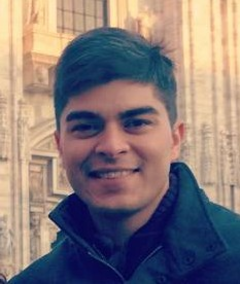}}]
	{Carlos Mastalli} received an M.Sc. degree in mechatronics engineering from the Simón Bolívar University, Caracas, Venezuela, in 2013 and a Ph.D. degree in bio-engineering and robotics from the Istituto Italiano di Tecnologia, Genoa, Italy, in 2017.
	
	He is currently an Assistant Professor at Heriot-Watt University, Edinburgh, U.K.
	He is the Head of the Robot Motor Intelligence (RoMI) Lab affiliated with the National Robotarium and Edinburgh Centre for Robotics.
	He is also appointed as Research Scientist at IHMC, USA.
	Previously, he conducted cutting-edge research in several world-leading labs: Istituto Italiano di Tecnologia (Italy), LAAS-CNRS (France), ETH Zürich (Switzerland), and the University of Edinburgh (UK).
	His research focuses on building athletic intelligence for robots with legs and arms.
	Carlos' research work is at the intersection of model predictive control, numerical optimization, machine learning, and robot co-design.
\end{IEEEbiography}

\begin{IEEEbiography}[{\includegraphics[width=1in,height=1.25in,clip,keepaspectratio]{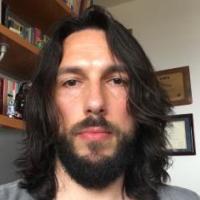}}]{Manolo Garabini}
	graduated in Mechanical Engineering and received the Ph.D. in Robotics from the University of Pisa where he is currently employed as Assistant Professor. His main research interests include the design, planning, and control of soft adaptive robots. He contributed to the realization of modular Variable Stiffness Actuators, to the design of the humanoid robot WALK-MAN and, more recently, to the development of efficient and effective compliance planning algorithms for interaction under uncertainties. Currently, he is the Principal Investigator in the THING H2020 EU Research Project for the University of Pisa, and the coordinator of the Dysturbance H2020 Eurobench sub-project. Finally, he is the coordinator of the H2020 EU Research Project Natural Intelligence.  
\end{IEEEbiography}

\end{document}